\documentclass{article}
\usepackage[accepted]{template/icml2026}
\usepackage{template/macros}
\usepackage{template/packages}
\usepackage[round,sort]{natbib}

\icmltitlerunning{A Mechanistic Study of Transformers Training Dynamics}

\begin{document}
\addtocontents{toc}{\protect\setcounter{tocdepth}{0}}

\twocolumn[
  \icmltitle{A Mechanistic Study of Transformers Training Dynamics}


  \icmlsetsymbol{equal}{*}

  \begin{icmlauthorlist}
    \icmlauthor{Ambroise Odonnat}{a}
    \icmlauthor{Wassim Bouaziz}{b}
    \icmlauthor{Vivien Cabannes}{c}
  \end{icmlauthorlist}

  \icmlaffiliation{a}{Inria, Univ. Rennes 2}
  \icmlaffiliation{b}{Mistral AI; work done while at Meta.}
  \icmlaffiliation{c}{FAIR at Meta}

  \icmlcorrespondingauthor{Ambroise Odonnat}{\href{mailto:ambroise.odonnat@gmail.com}{ambroise.odonnat@gmail.com}}
  
  \icmlkeywords{Machine Learning, ICML}

  \vskip 0.3in
]



\printAffiliationsAndNotice{}  

\begin{abstract}
Large-scale pretraining of transformers has been central to the success of foundation models. However, the scale of those models limits our understanding of the mechanisms at play during optimization. In this work, we study the training dynamics of transformers in a controlled and interpretable setting. On the sparse modular addition task, we demonstrate that specialized attention circuits, called \say{\emph{clustering heads}}, can be implemented during gradient descent to solve the problem. Our experiments show that such pathways naturally emerge during training. By monitoring the evolution of tokens via a visual sandbox, we uncover a two-stage learning and the occurrences of loss spikes due to the high curvature of normalization layers. Our findings provide several insights into patterns observed in more practical settings, such as the pretraining of large language models. 
\begin{center}
\faGithub \quad \href{https://github.com/facebookresearch/pal}{\texttt{facebookresearch/pal}}
\end{center}
\vspace{-1.2em}
\end{abstract}

\section{Introduction}
Transformers~\citep{vaswani2017attention} have become the default backbone of state-of-the-art models in a wide range of domains, including natural language processing~\citep[NLP,][]{touvron2023llama,brown2020gpt3}, computer vision~\citep{caron2021emerging, dosovitskiy2021vit}, time series forecasting~\citep{nie2023patchtst, ilbert2024samformer}, and mathematical reasoning~\citep{comanici2025gemini25pushingfrontier, deepseekai2025deepseekr1}. The models' size and amount of training data have drastically increased as a by-product of the scaling hypothesis~\citep{hoffmann2022trainingcomputeoptimallargelanguage, kaplan2020scalinglawsneurallanguage,charton2024learning}. This makes the analysis and ablation studies on such systems challenging. As a result, our understanding of the \emph{training dynamics} and causes of instability in large transformer models remains limited, with much of the knowledge in this area relying on heuristic tricks and techniques.

\paragraph{Mechanistic viewpoint.} To address these limitations, we draw inspiration from the field of \emph{mechanistic interpretability}, which aims to understand the internal mechanisms at play in large language models by reverse-engineering their computations~\citep{elhage2021mathematical}. A common approach in this framework is to study the pathways, or \emph{circuits}, of a neural network that transform an input into a given output. This viewpoint has been used to study the emerging capabilities of LLMs with in-context learning (ICL)~\citep{olsson2022induction, edelman2024evolution}, chain-of-thought~\citep{cabannes2024iteration}, and memorization in transformers~\citep{bietti2023birth}. In line with these studies, we aim to design a controlled setup to decipher the training of transformers without potential confounding factors. Our focus is on obtaining a finer-grained understanding of the evolution of sequence embeddings during the training of a single transformer block. Training the whole architecture could then be considered as the joint training of several blocks, with the inputs of a given block being the output of its preceding block. 

\paragraph{Methodology.} Motivated by the fact that many NLP and reasoning tasks involve invariances in the data~\citep{robinson2023leveraging, singh2025exposing}, we consider the \emph{sparse modular addition} problem, which requires a correct handling of invariances to be solved and ensures non-trivial optimization dynamics~\citep{odonnat2025circuit}. Similar mathematical tasks have been used in prior works for their clear structure and ease to observe the pathways implemented~\citep{ameisen2025circuit}. We frame the problem as a supervised question-answering task and train small transformers on it. This enables us to fully characterize how transformers learn to solve the task, which is a step towards opening the black box~\citep {charton2024learning, koh2017influence}. To analyze the training dynamics on this task, we view sequences of tokens as a system of interacting particles and design a visual sandbox to observe their evolution during training. It allows us to identify and analyze multiple phenomena observed in more practical settings. This showcases the benefits of our setup in studying, at a small scale, the training behavior of bigger models. 

\paragraph{Our contributions.} The code is open-sourced at \url{https://github.com/facebookresearch/pal} and our main contributions can be summarized as follows:
\begin{enumerate}[leftmargin=*]
    \item \textbf{Theoretical analysis:} We frame the {\em sparse modular addition} problem as a question-answering task and demonstrate that transformers trained with gradient descent implement specific pathways, dubbed \emph{clustering heads}, to solve it by capturing the invariances in the data. 
    \item \textbf{Visual sandbox:} We design a visual sandbox to observe the evolution of tokens, seen as interacting particles, during training and use it to study the implementation of clustering heads. 
    \item \textbf{Training dynamics:} We observe that such circuits emerge after a two-stage learning, akin to the in-context learning settings~\citep{olsson2022induction}. We observe the occurrence of loss spikes during training and suggest potential strategies for mitigation, relating our findings to the pretraining of large language models.
\end{enumerate}
We hope our setup can serve as a viable testbed to study training stability on a small scale to gain intuition for larger-scale settings.

\section{Synthetic setup}
\label{sec:setting}
Let $n,m \in \bN$ be nonnegative integers. Throughout the paper, we use the notation $[n] = \{1, \ldots, n\}$. The Euclidean norm of a vector $x \in \bR^n$ is denoted by $\lVert x \rVert$. By abuse of notation, we denote by $\lVert A \rVert$ the entry-wise $\ell_2$-norm of a matrix $A \in \bR^{n \times m}$ which amounts to computing the Euclidean norm of $A$ flattened. The probability simplex over $\bR^n$ is denoted by $\Delta_n$. The total variation distance between two probability distributions $P_1, P_2$ is denoted by $\TV{P_1}{P_2}$.

\paragraph{Sparse modular addition.} This problem is inspired by the sparse parity problem \citep{barak2023hidden,abbe2023provable} and the modular addition problem \citep{power2022grokking,nanda2023progress} and is characterized by the following parameters:
\begin{itemize}
    \item Input length \( N \in \mathbb{N} \),
    \item Vocabulary size \( p \in \mathbb{N} \),
    \item Sparsity index \( k \in [N] \), and a set of indices \( I \subset [N] \) with cardinality \( |I| = k \).
\end{itemize}
Our default configuration is set to \( N=12 \), \( k=5 \), and \( p \in \{2, 3\} \).
Without loss of generality, we assume \( I = [k] := \{1, 2, \ldots, k\} \).
Inputs are sequences of \( N \) tokens \( x_t \) in the ring \( \mathbb{F}_p = \mathbb{Z} / p\mathbb{Z} \simeq [p] \), and the corresponding targets are the sum of the first \( k \) terms modulo \( p \). Formally, we aim to learn a mapping:
\begin{equation}
\tag{SMA}
\label{eq:sma}
\myfunction{f^*}{(\mathbb{F}_p)^N}{\mathbb{F}_p}{x = (x_1, \ldots, x_N)}{\sum_{t \in [k]} x_t.}
\end{equation}
This mapping defines a deterministic conditional distribution linking input and output data through the formula \( p(y|x) = \mathbf{1}_{\{f^*(x)=y\}} \), and the training samples follow $x,y \sim p(\cdot)$. The~\eqref{eq:sma} problem can be framed as a supervised question-answering task with tokens embedded in a vocabulary space $\mathcal{V} = [p]$ of size $p$. Input questions are represented as sequences of tokens. They are randomly sampled and used to train sequence-to-sequence transformers by minimizing the cross-entropy loss between the model's predictions and the correct answer, that is, the sum of the $k$ first terms of the inputs modulo $p$. While sparse modular addition might seem simple to solve for a human, this is a hard task to solve for neural networks, which have been shown to struggle on similar arithmetic tasks~\citep{charton2024learning, lee2024teaching, dziri2023faith, saxton2018analysing}, involving non-trivial learning dynamics~\citep{odonnat2025circuit}. In particular, the model needs to capture the invariance in the data and correctly focus on the informative, i.e, \emph{non-spurious}, tokens $x_i$ for $i \in [k]$.

\paragraph{Transformer architecture.} To conduct our analysis, we consider a simplified transformer model containing all the key elements of its larger counterparts commonly used in question-answering tasks~\citep{vaswani2017attention, raffel2023exploringlimitstransferlearning}. Notably, it relies on a cross-attention module, a feedforward network, residual connections, normalization layers, and weight tying~\citep{wolf2017tying}. Following modern models~\citep{dubey2024llama3herdmodels}, we use a pre-norm configuration~\citep{xiong2020prenorm} and the RMSNorm~\citep{zhang2019root}, which is similar to LayerNorm~\citep{ba2016layer} but computationally more efficient. To ease notations, we incorporate the RMSNorm trainable parameters into the attention and feedforward weights following~\citet{castin2024smooth}, which does not change the expressivity or optimization of the model. We recall below how tokens are processed through the model. Since the sparse modular addition problem is inherently discrete, we first embed each token in dimension $d$ with both semantic and positional information before normalizing it on the unit sphere $\bS^d$. Formally, given a learnable token embedding $E:\bF_p \to \R^d$, and a learnable position embedding $P:[N] \to \R^d$, a sentence in token space is lifted to a sentence in embedding space $z \in \bR^{d \times N}$ through the following operation that applies to each $x_t$ for $t \in [N]$
\[
\tag{token embedding}
z_t := Z(x_t, t) := \frac{E(x_t) + P(t)}{\norm{E(x_t) + P(t)}}.
\]
Then, a cross-attention mechanism is applied to the sequence of tokens $z$, leading to a single sentence embedding $\xi \in \R^d$. The sentence embedding can be written in matrix form as
\begin{equation}
    \tag{sentence embedding}
    \xi := (V z) \operatorname{softmax}(\frac{z^\top q}{\sqrt{d}}) \in \R^d,
\end{equation}
where we follow~\citet{ruiqi2024linear} and merge the query and value into a vector $q \in \R^d$ to avoid cumbersome notations. We omit the key matrix, which would act as extra parameters that do not increase the expressivity of our model since $z$ can be set to anything thanks to $E$ and $P$. The sentence embeddings are then passed through a feedforward network with hidden dimension $h$ (typically taken as $h=4d$) implemented as an MLP with two layers $U \in \bR^{d \times h}$ and $W \in \bR^{h \times d}$, followed by a residual connection. It implements the following transformation:
\[
\tag{sentence transform}
\psi := \xi + U \sigma\mleft(\frac{W\xi}{\|\xi\|}\mright) \in \mathbb{R}^d
\]
with \( \sigma \colon x \mapsto x \phi(x) \) the GeLU activation~\citep{hendrycks2023gaussianerrorlinearunits}. $W$ receives the sentence embeddings before projecting them into a higher dimension, while $U$ reassembles the embeddings, projecting them back to the original dimension. As such, the feedforward can be equivalently seen as a combination of \( h \in \mathbb{N} \) ``receptors" weights \( w_i \in \mathbb{R}^d \) and \( h \) ``assemblers" vectors \( u_i \in \mathbb{R}^d \) for \( i \in [h] \), with  $(u_i)_{i=1}^h$ the columns of $U$ and $(w_i^\top)_{i=1}^h$ the rows of $W$. The final sentence vector \( \psi \) is decoded back to token space based on how it aligns with the respective token embeddings, which amounts to using weight-tying in the implementation~\citep{press2017using}. More specifically, the model's logits write
\[
\tag{logits}
\zeta = (E(v)^\top \psi)_{v \in \mathbb{F}_p} \in \mathbb{R}^p.
\]
Abstracting all the learnable weights into a single vector \( \theta \), the probability distribution over answers in $[p]$ writes $\hat{p}_\theta(\cdot|x) \in \Delta_n$ and is obtained by applying a softmax layer on top of the logits. For any $j \in [p]$, $\hat{p}_\theta(y=j|x)$ is simply the $j$-th entry of the vector $\hat{p}_\theta(\cdot|x)$. The model is optimized by minimizing the cross-entropy loss between its prediction and the correct answer. The loss writes 
\[
\tag{loss}
\mathcal{L}(\theta) := \mathbb{E}_{x,y}\left[-\log(\hat{p}_\theta(y|x))\right],  
\]
where the expectation is taken over the distribution of training samples $(x, y)$. 

\section{Learning clustering heads by gradient descent}
\label{sec:solving_task}
In this section, we study how transformers learn to solve the~\eqref{eq:sma} problem with gradient descent. We describe a natural circuit, dubbed \emph{clustering head}, that, if correctly implemented, enables the transformer to solve the task, and we demonstrate that such a circuit can be efficiently learned by gradient descent.

\paragraph{Circuit perspective.} In the sparse modular addition problem, the output $y$ is invariant to two sets of transformations of the inputs $x = (x_1, \ldots, x_N)$.
\begin{itemize}
    \item ``Permutation invariance:'' $y$ does not depend on permutation of non-spurious tokens, i.e. for any permutation $\sigma$ of $\{1, \ldots, k\}$, \[f^*(x) = f^*((x_{\sigma(1)}, \ldots, x_{\sigma(k)}, x_{k+1}, \ldots, x_N)).\]
    \item ``Suffix invariance:'' $y$ does not depend on the suffix $(x_{k+1}, \ldots, x_N)$.
\end{itemize}
These two sets of invariants can be easily enforced by the embedding layer.
Any architecture where the position embeddings satisfy $P(t) = P(1)$ for $t\in[k]$, will be permutation invariant, meaning that its output will be invariant to permutations of the non-spurious tokens.
Similarly, suffix invariance can be enforced by ensuring that the query vector primarily aligns with the token embeddings $z_t$ for $t \in [k]$, allowing the sequence embedding $\xi$ to be invariant to the sequence suffix.
Such a construction would yield ${k + p - 1 \choose k}$ clusters of sequence embeddings,\footnote{This number corresponds to the number of ways to split $k$ into $p$ buckets, which is also the number of stars and bars configurations with $k$ stars and $p-1$ bars.} which the feedforward layer could scatter into as many decision regions to map each sequence embedding $\xi$ to the correct output class $y\in[p]$. The pathway that realizes such operations is called a \emph{clustering head}. Clustering heads define one way for the transformer to handle invariances: the model first attends to the non-spurious tokens $(x_i)_{i=1}^k$, then regroups them by value to determine the parity of each subgroup and, finally, of the whole sequence. Although this is not the only possible circuit, it resonates with recent studies~\citep{lindsey2025biology, odonnat2025circuit}, which showed that transformers and notably large language models split hard problems into simpler sub-problems, for instance, to compute a complex arithmetic task.

\paragraph{Learning clustering heads.} A natural question is whether \emph{clustering heads} can be implemented in practice by a transformer trained via gradient descent, which aims at finding local minimizers of the loss $\mathcal{L}$. Such minimizers are stationary points, i.e., solutions of $\nabla \mathcal{L} = 0$. Akin to~\citet{nichani2024factualrecall}, we focus on the attention and feedforward dynamics, assuming fixed embedding weights. In the following lemma, we provide a closed-form expression of the gradients (the full statement is deferred to~\cref{lem:grad_loss}).
\begin{boxlem}[Closed-form gradients]
\label{lem:closed_form_grad_informal}
Let $m = d(1 + d + 2h)$. The trainable parameters of the transformer write $\theta \in \bR^m$, encompassing the attention and feedforward matrices. Denoting by $\hat{p}_{\theta}(\cdot|x)$ and $p(\cdot|x)$ the model's prediction and ground-truth for an input sequence $x = (x_1, \ldots, x_N)$, we have
    \begin{equation*}
    \nabla \mathcal{L}(\theta) = \sum_{j=1}^p \mathbb{E}_x \mleft[ (\hat{p}_{\theta}(y=j|x) - p(y=j|x))\cdot \nabla_\theta \zeta_j \mright].
    \end{equation*}
\end{boxlem}

Computing the loss gradient involves reweighting each logit's gradient by the corresponding discrepancy between the model's prediction $\hat{p}_{\theta}$ and the ground-truth $p$. \cref{lem:closed_form_grad_informal} allows us to characterize the stationary points of the loss and show that the parameters of transformers that implement clustering heads are stationary points (the full statement is deferred to~\cref{prop:stationary_points}).
\begin{boxprop}[Learning clustering heads, informal]
    \label{prop:learning_clustering_heads_informal}
    Let $\theta$ be the parameters of a transformer model that implements a clustering head. Then, $\theta$ is a stationary point of the loss, that is $\nabla \mathcal{L}(\theta) = 0$.
\end{boxprop}
\cref{prop:learning_clustering_heads_informal} implies that clustering heads can be learned by gradient descent. We note that it does not provide guarantees on the convergence to such a circuit (we will see in the next section that, in practice, clustering heads are implemented by transformers). Further theoretical treatment to characterize the stationary points as local, global minima, or saddle points is an interesting direction left for future work. In the following, we focus on the training dynamics during gradient descent.

\section{Visualizing the training dynamics}
\label{sec:visualizing_dynamics}
In this section, we see tokens as interacting particles, and design a visual sandbox to observe their evolution during training. It allows us to study the practical implementation of clustering heads, noticing a two-stage learning and the occurrence of loss spikes during training, and further relate our findings to the pretraining of large language models. The implementation details are given in~\cref{app:add_exp}.

\subsection{Visual sandbox}
Tokens processed by transformers interact through the attention module, which allows us to study their evolution as a dynamic system of interacting particles controlled by an ordinary differential equation~\citep{geshkovski2025measure, lu2019understanding, geshkovski2023perspective}. We adopt this macroscopic viewpoint and follow the mechanistic viewpoint~\citep{olah2022lowdim} by embedding all computations in the plane by taking $d=2$ and tracking several key aspects during training, such as:

\paragraph{Position embeddings.}
We visualize them as a point cloud with the ``spurious" embeddings \(P(t)\) for \(t \notin [k]\) represented by squares, and the ``non-spurious" ones, the \(P(t)\) for \(t \in [k]\), by circles. Ideally, the transformer would collapse all spurious (resp. non-spurious) position embeddings into a single point, learning invariance of \(y\) to sentence suffixes (resp. to permutation of non-spurious tokens positions).

\paragraph{Token embeddings.}
We visualize the token embeddings $Z(x, t) \propto E(x) + P(t)\) for \((x, t) \in \mathbb{F}_p \times [N]\). We maintain the same circle and square distinction and use the same color for both \(E(x) + P(t)\) and \(E(x') + P(t)\), for \(x' \in \mathbb{F}_p \setminus \{x\}\). On the normalized plot, we also plot the query \(q\) as an arrow in \(\mathbb{R}^2\), helping us understand where the attention module learns to focus.

\paragraph{Attention map.} We visualize the concatenation of attention vectors for different sentences. It enables us to follow the change in activation patterns, even though it is a pure function of the normed embedding visualization.

\paragraph{Value transform.}
We visualize \(VZ(x, t)\) as a point cloud. This allows us to understand how sequence embeddings are built and how the value matrix may overcome faulty attention patterns.
\begin{figure*}[!h]
    \centering
    \includegraphics[width=\linewidth]{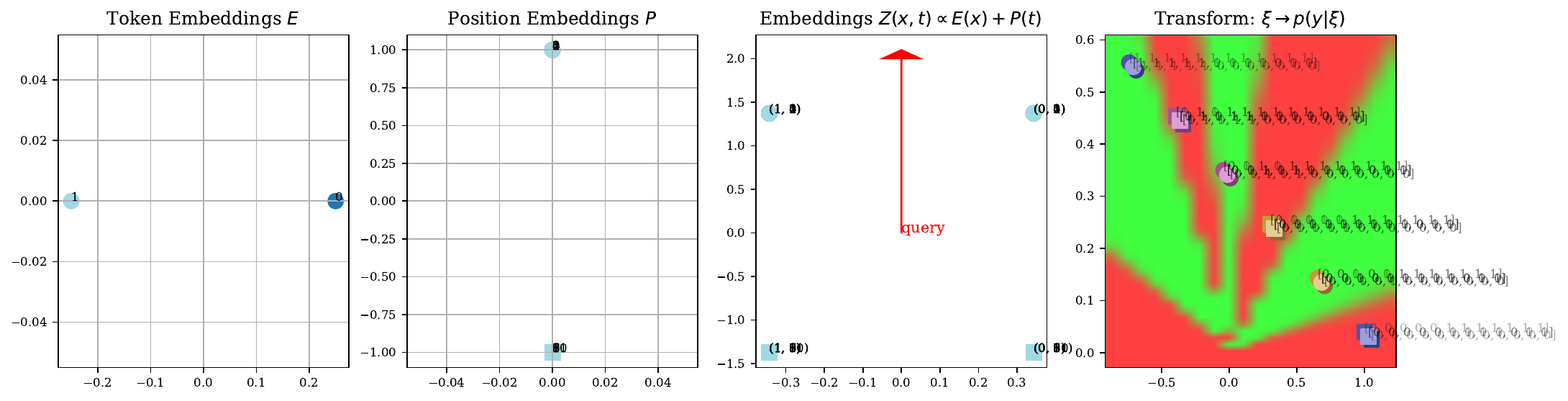}
    \caption{\textbf{Clustering Head}. Implementation of an idealized circuit that captures the invariants of the problem. \textit{From left to right}: we first plot the two token embeddings $E(x)$ for $x\in[p]$ with $p=2$. We then plot the twelve position embeddings $P(t)$. The spurious (resp. non-spurious) positions are represented with squares (resp. circles), and all collapse to a single point. The normalized embeddings $Z(x, t)$ are plotted in the third frame, annotated with $(x, t)$, with the query vector $q$ represented as a red arrow. This arrow points in the direction of the embedding $Z(x, t)$ for $t\in[k]$, allowing the attention mechanism to focus exclusively on non-spurious tokens. Finally, we plot the sequence embeddings $\xi$ for some $(x_t)$, and the output of the feedforward transform. The embeddings are clustered and respect the invariance, and the feedforward layer can then map each cluster to a probability vector $\hat{p}_\theta$. Each output class is represented by a color.}
    \label{fig:circuit_ideal}
\end{figure*}
\paragraph{Sequence embeddings and transforms.}
We visualize the sequence embeddings \(\xi\) (or their transforms \(\zeta\)) for a set of predefined sentences. These sentences are built by iterating over prefixes \((x_t)_{t \in [k]}\) and suffixes \((x_t)_{t \notin [k]}\). Sentences that share the same prefix have the same color. Sentences whose prefixes are equivalent up to a token position permutation have similar colors. Squares, circles, and triangles are used to distinguish between the classes of the sentences. 
The sequence embeddings visualization is a direct function of the value transformation and the normalized embeddings.

\paragraph{Transform level lines.}
We visualize the mapping from sentence embedding \(\xi\) to their associated learned probabilities \(\hat{p}_\theta(y|x)\). We also plot the sentence embedding on the same plot to better understand the level line changes.

\paragraph{MLP receptors and assemblers.}
We visualize the \(w_i \in \mathbb{R}^2\) (and \(u_i \in \mathbb{R}^2\)) that define the MLP transform as a point cloud in \(\mathbb{R}^2\). A consistent color scheme is used to link receptors with the corresponding assemblers.

\paragraph{Loss and accuracy.}
We visualize the current train loss, test loss, and accuracy. It is interesting to put those classical quantities in relation to the other visualizations to better understand the loss spikes, loss plateaus, and phase transitions.
\subsection{Clustering heads implementation}
\label{sec:clustering_heads_implementation}
\begin{figure}[!h]
    \centering
    \includegraphics[width=\linewidth]{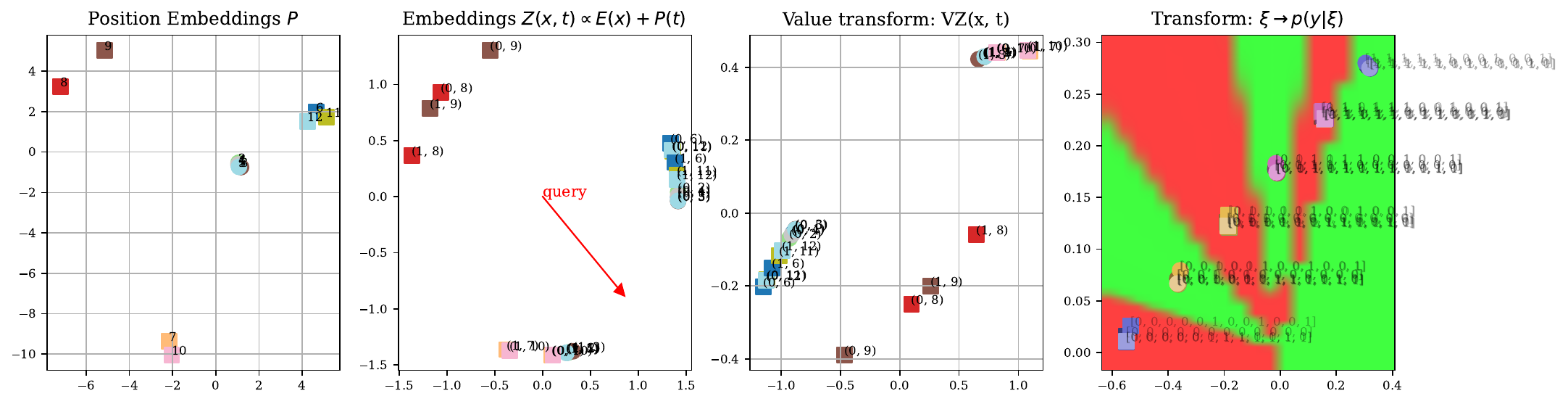}
    \caption{Faulty attention corrected by value collapse.}
    \label{fig:circuits_corrected}
\end{figure}

\begin{figure}[!h]
    \centering
    \includegraphics[width=\linewidth]{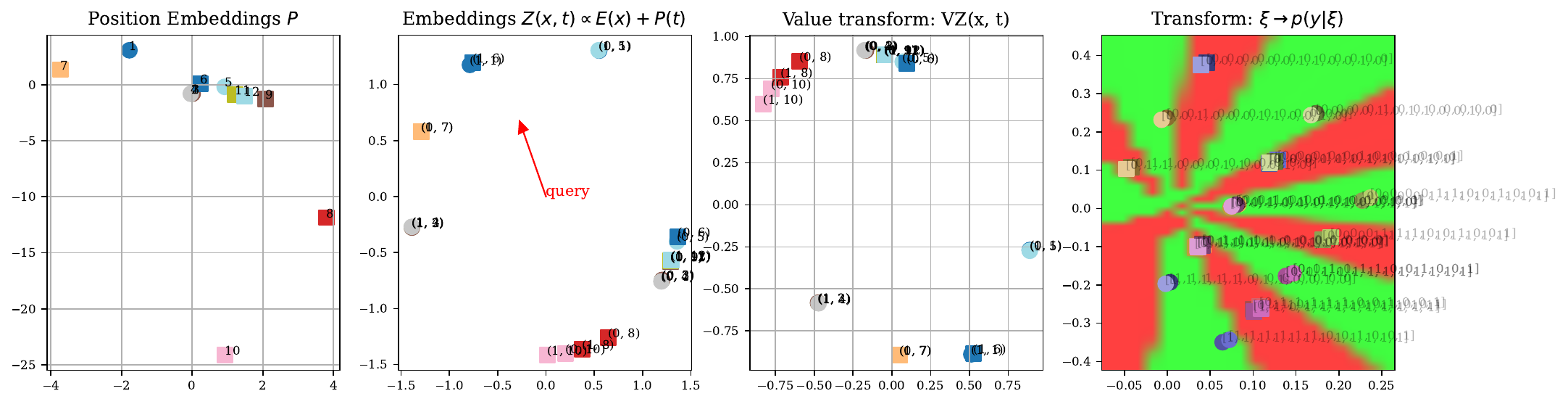}
    \caption{Embeddings having only learned some invariants.}
    \label{fig:circuits_invariant}
\end{figure}

\begin{figure}[!h]
    \centering
    \includegraphics[width=\linewidth]{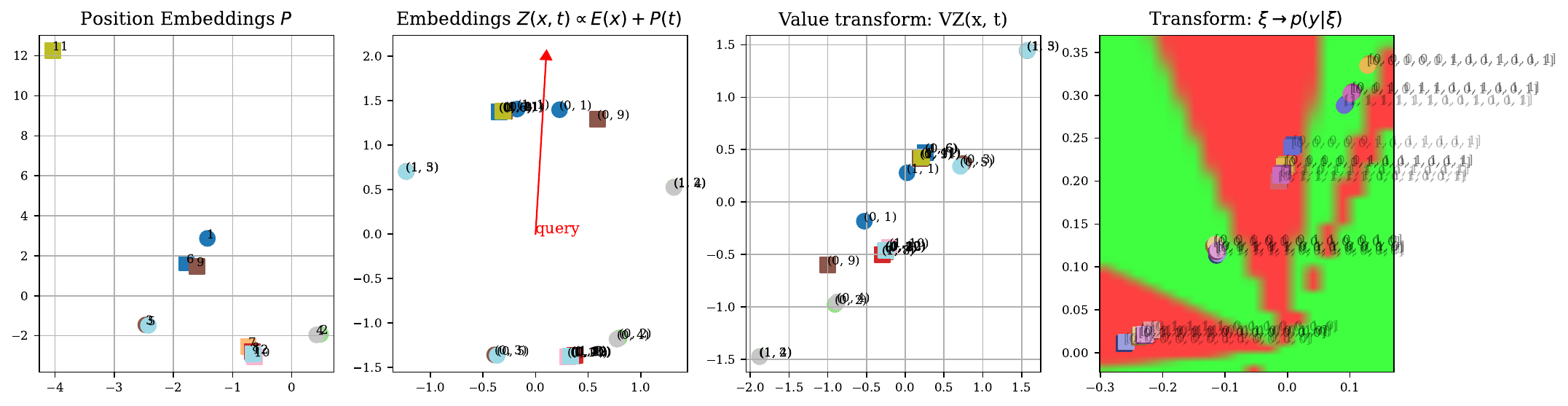}
    \caption{Fuzzy construction.}
    \label{fig:circuits_fuzzy}
\end{figure}

\begin{figure}[!h]
    \centering
    \includegraphics[width=0.45\linewidth]{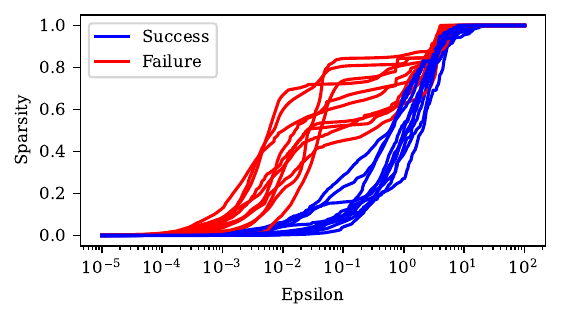}
    \hfill
    \includegraphics[width=0.45\linewidth]{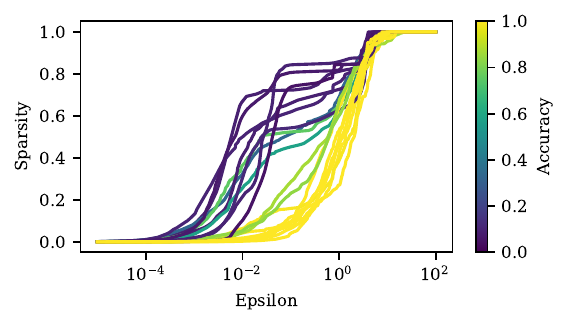}
    \caption{\textit{Connection between performance and sparsity}. Since the activation is the GELU that does not map all the negative entries to $0$ (contrary to the ReLU), we define the activation sparsity as the percentage of entries with values lower than $\epsilon>0$. We display the evolution of the activation sparsity of $20$ trained models with $\varepsilon \in [10^{-5}, 10^2]$. \textbf{Left:} Successful models (i.e., with test accuracy above $0.9$) in \textcolor{blue}{blue} have less sparse activation than failed models in \textcolor{red}{red}. \textbf{Right:} The color indicates the models' test accuracy (the lighter, the better). The performance increases as the activation sparsity decreases.
    }
    \label{fig:sparsity}
\end{figure}

The \emph{clustering head} described in~\cref{sec:solving_task} can be visualized in~\cref{fig:circuit_ideal}, with the model focusing on the informative tokens to regroup them by values with the attention module before computing the parity of the sequence with the feedforward. In the previous section, we showed that these pathways could be implemented by gradient descent. Since transformers can perform similar operations in many different ways, we observe in practice variations from the idealized circuit, categorized below. 
\paragraph{Faulty attention corrected by value.} In many instances, we observe that the attention scores are not fully concentrated on the first five tokens but are compensated by the value matrix, which effectively collapses the embeddings of the spurious tokens that are attended to. 
This allows the sequence embeddings to remain invariant to the suffix of the sequence, as can be seen in \cref{fig:circuits_corrected}. 
\paragraph{Partially learned invariants.} Frequently, the sequence embeddings have not fully learned all the suffix and prefix invariants, resulting in more than six clusters of sequence embeddings\footnote{This corresponds to ${k + p - 1 \choose k}$ for $p=2, k=5$.}. 
Specifically, \cref{fig:circuits_invariant} illustrates a scenario where the sequence embeddings are not invariant to the value \(x_6\), as evidenced by the positions of the blue squares \((0, 6)\) and \((1, 6)\) on the plot.
They also lack invariance to permutations of the token in the first or fifth positions with another of the non-spurious tokens. 
This results in a sequence embedding that presents more clusters than the idealized model, leading to a greater number of connected decision regions in the feedforward layer.
\paragraph{Fuzzy constructions.}
Occasionally, we encounter fuzzy constructions where the sequence embeddings are clustered according to unconventional patterns that nonetheless generalize to unseen data. 
Such a construction is presented in \cref{fig:circuits_fuzzy}. 
\begin{figure*}[!h]
\centering
\begin{minipage}{.45\textwidth}
  \centering
    \includegraphics[width=\linewidth]{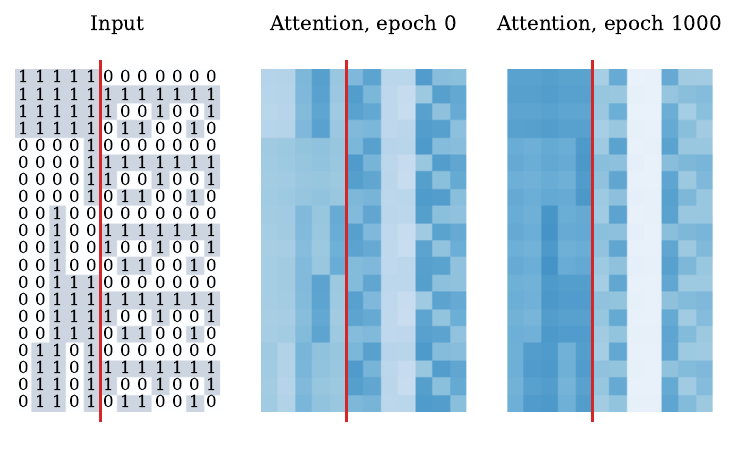}
    \caption{\textbf{Impact of the initialization.} For each input sequence (\textbf{left}), we plot the attention weights at the start of training (\textbf{middle}) and at the end of training (\textbf{right}); the darker, the higher. Final attention maps correlate with the original one, illustrating the impact of the original configuration on the implemented circuit.}
    \label{fig:initialization}
\end{minipage}%
\quad
\begin{minipage}{.45\textwidth}
  \centering
    \includegraphics[width=\linewidth]{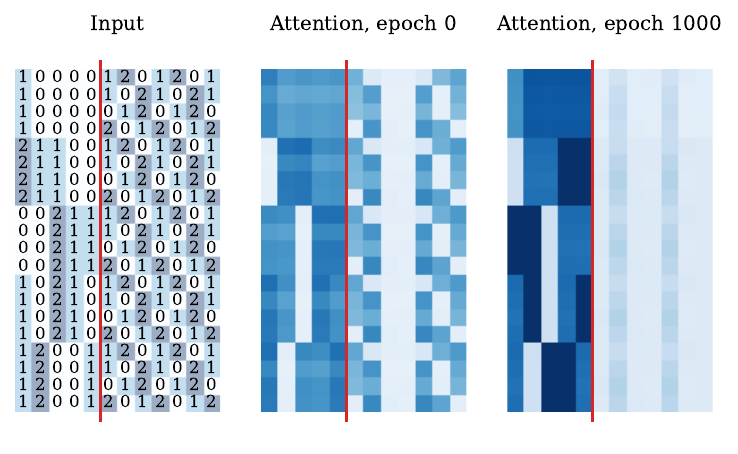}
    \caption{\textbf{Transferability of circuits}. This study is akin to \cref{fig:initialization} when the model is first pre-trained with $p=2$ and then finetuned with $p=3$. This illustrates that the final variations of the original circuit depend on the original weight configuration. We notice that the model learns better to focus on the non-spurious first $5$ tokens.}
    \label{fig:transferability}
\end{minipage}
\end{figure*}
\paragraph{From initialization to transferability of circuits.} We observe in \cref{fig:initialization} the strong similarity between the attention patterns at the start and at the end of the training, indicating that initialization plays a crucial role in the concrete realizations of clustering heads. We notice that the transformer learns to focus on the $1$ among the non-spurious tokens, before counting them and deducing the number of $0$ to make its final prediction. 
Among the lessons of training very large models is that some sources of data may facilitate the learning of certain skills~\citep{touvron2023llamaopenefficientfoundation,dubey2024llama3herdmodels,cabannes2024iteration}, suggesting the transferability of circuits. We illustrate this phenomenon in~\cref{fig:transferability}, which shows that training first a model on a simpler task (e.g., $p=2$) helps it better solve a more challenging one (e.g., $p=3$), as can be seen in along with the impact of initialization (more details in \cref{app:transfer}).
\paragraph{Connectivity of neurons.} Motivated by recent works studying the sparsity of transformers' activations~\citep{li2023lazy, mirzadeh2024relu}, we investigate how neurons connect to implement \emph{clustering heads}. We show in \cref{fig:sparsity} that the models that successfully solve the task have denser activation, suggesting that many neurons are necessary to correctly implement a clustering head and thus capturing the invariance of the task (more details in \cref{app:sparsity}). 
\begin{figure*}[!t]
    \centering
    \includegraphics[width=\linewidth]{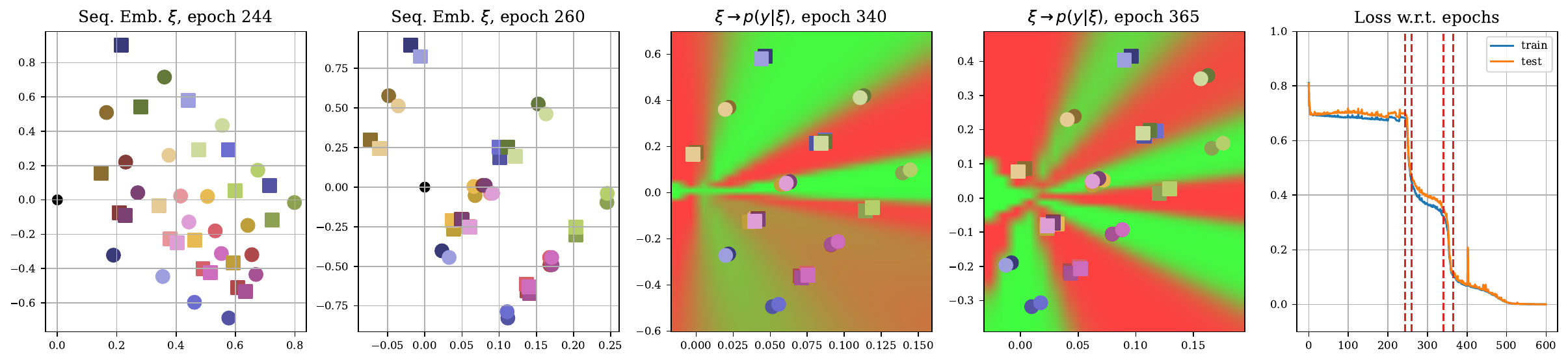}
    \caption{\textbf{Two-phase learning}. 
    \textit{From left to right:} 
    \textbf{(1)} During the first snapshot, the sequence embeddings lack any clear structure.
    \textbf{(2)} They suddenly become clustered after the first loss drops, as seen in the second snapshot. 
    \textbf{(3)} At this point, the MLP already classifies some clusters correctly (third snapshot).
    \textbf{(4)} A second loss drop occurs as the MLP gets fitted (last snapshot). \textit{Right:} Loss profile featuring two significant drops in loss, marked by four red dashed lines at key snapshots.
    }
    \label{fig:dynamics}
\vspace{-1em}
\end{figure*}
\subsection{Two-stage learning: escaping saddle points}
\label{sec:training_dyn}
To better understand how clustering heads emerge, we observe the evolution of the tokens along with the loss landscape during training. We notice a two-stage learning process with the model first clustering the sequence embeddings, before fitting a classifier on top. As displayed in~\cref{fig:dynamics} (right), the loss curves present two drops, corresponding to the learning of different parts of the network. Our findings resonate with theoretical arguments in the deep learning theory literature on saddle point \citep{dauphin2014identifyingattackingsaddlepoint,du2017gradientdescentexponentialtime,Chi_2019} and stage-wise transitions of in-context learning of transformers~\citep{olsson2022induction, hoffmann2024eurekamoments, reddy2024abrupt, edelman2024evolution, minegishi2025beyond, varre2025learning}.
\begin{figure}[!b]
    \centering
    \includegraphics[width=\linewidth]{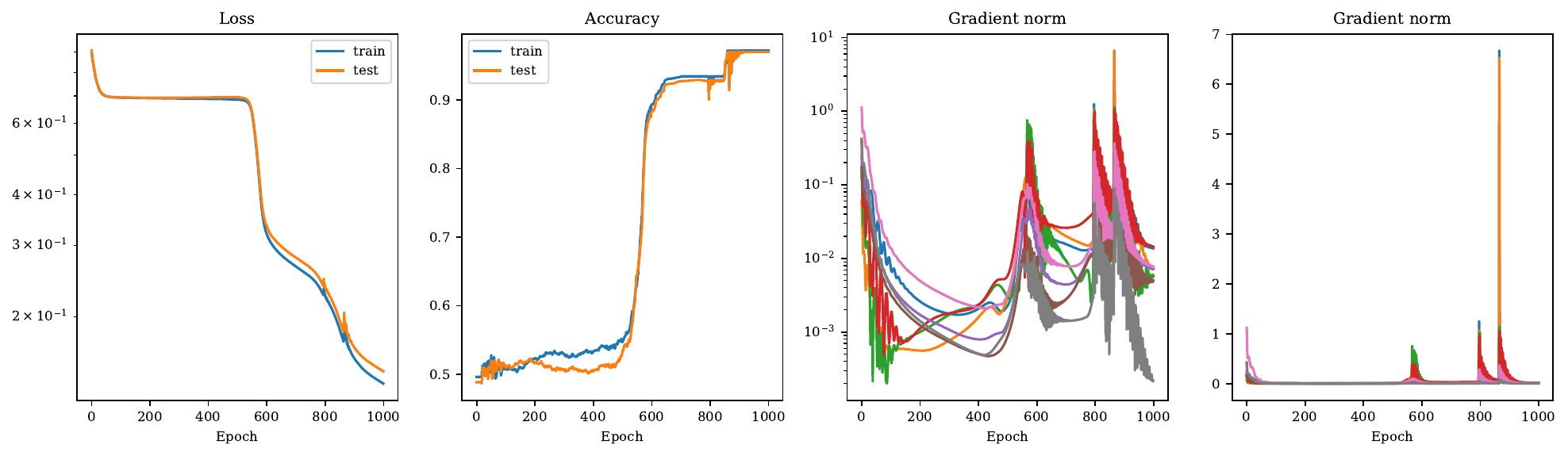}
    \caption{\textbf{Escaping saddle points}. \textit{From left to right:} Evolution of train and test losses, the corresponding accuracies, the evolution of gradient norms for each layer in log-scale, and the similar evolution in linear scale in full-batch. We see that the learning phases of \cref{sec:training_dyn} appear in tandem with high gradient norms. This can be seen in the last subfigure, where the three peaks correspond to the loss drops and their corresponding plateaus.}
    \label{fig:grad-norms-full-batch}
\end{figure}
\paragraph{First loss drop: learning of the sequence embeddings.}
The first loss drop coincides with the learning of the sequence embeddings.
It corresponds to a phase change in the dynamics of the weights.
Before the first phase change, the weights seem to wander as if trapped in a saddle point, waiting for a clear signal to escape it.
At one point, they all move quite rapidly to create a relatively definitive structure for the sequence embeddings.
Interestingly, we also notice that the time it takes for this phase change to occur when changing the training hyperparameters can vary quite a lot, reflecting the highly unpredictable time needed to escape from the saddle point.
This is illustrated in \cref{app:hyper_params} where \cref{fig:ablation-seed} shows the high variability of the test accuracy after 1000 epochs when slightly varying one hyperparameter (\cref{fig:ablation} shows the same phenomenon with more regularity when averaging over the runs).
The exit from a saddle point can also be understood via the gradient norms, as discussed at the end of this section.
\paragraph{Second loss drop: fitting of the MLP.}
The second loss drop is due to the learning of the feedforward network.
This change is about fitting the MLP weights to assign the correct classes to the different clusters created during the learning of the sequence embeddings.
Interestingly, this second loss drop does not correspond to a clear phase change in the dynamics of the weights.
The MLP weights seem to evolve at a continuous speed, although the corresponding decision frontiers change relatively strongly.
We notice that this second loss drop appears soon after the first one, if not simultaneously. When the model's training stagnates, the loss plateaus for several epochs before decreasing again.
This behavior may be understood by the optimizer reaching a saddle point in the loss function w.r.t. the model's parameter, that is, the gradient of the loss is (almost) zero, but the optimization has not yet reached a local minimum. The connection between gradient norms and learning phases is salient in \cref{fig:grad-norms-full-batch}, where we can see that loss drops occur in tandem with high gradient norms for each layer. 

\paragraph{Theoretical insights.} \cref{prop:informal_gradient_bounds} characterize how the gradient magnitude is controlled along the training (the full statement is deferred to \cref{prop:gradients_bound}).
\begin{boxprop}[Gradient upper-bound, informal]
\label{prop:informal_gradient_bounds}
Let $\theta$ be the transformer's parameters. Assuming bounded token embeddings, the gradient of the loss verifies 
 \[
 \lVert \nabla \mathcal{L}(\theta)\rVert = \mathcal{O}\mleft( \tilde{B} \cdot \sqrt{\mathscr{E}}\mright),
 \]  
 where $\mathscr{E} = \mathbb{E}_x\mleft[\TV{\hat{p}_{\theta}(\cdot|x)}{p(\cdot|x)}\mright]$ captures the ability of the transformer to solve the task and $\tilde{B}$ depends solely on the spectral norm of the model's parameters.
\end{boxprop}
We see that the term $\mathscr{E}$ captures the ability of the model to correctly solve the task. In the early iterations, the model behaves poorly, which imposes weak control on the gradient norm. This allows for large gradient norms and hence for bigger updates to the weights. The loss can vary rapidly, as can be seen by the steep descent of \cref{fig:dynamics}. During training, the model improves as can be seen by the increase in accuracy (both train and test), leading to a smaller $\mathscr{E}$ and strengthening the control on the norm. In parallel, the gradient updates are of smaller magnitude as well as the loss variations. This implies a less steep descent in the loss as observed in~\cref{fig:dynamics}.
\subsection{Occurrence of loss spikes}
\label{sec:loss_spikes}
\begin{figure}[!h]
    \centering
    \includegraphics[width=.9\linewidth]{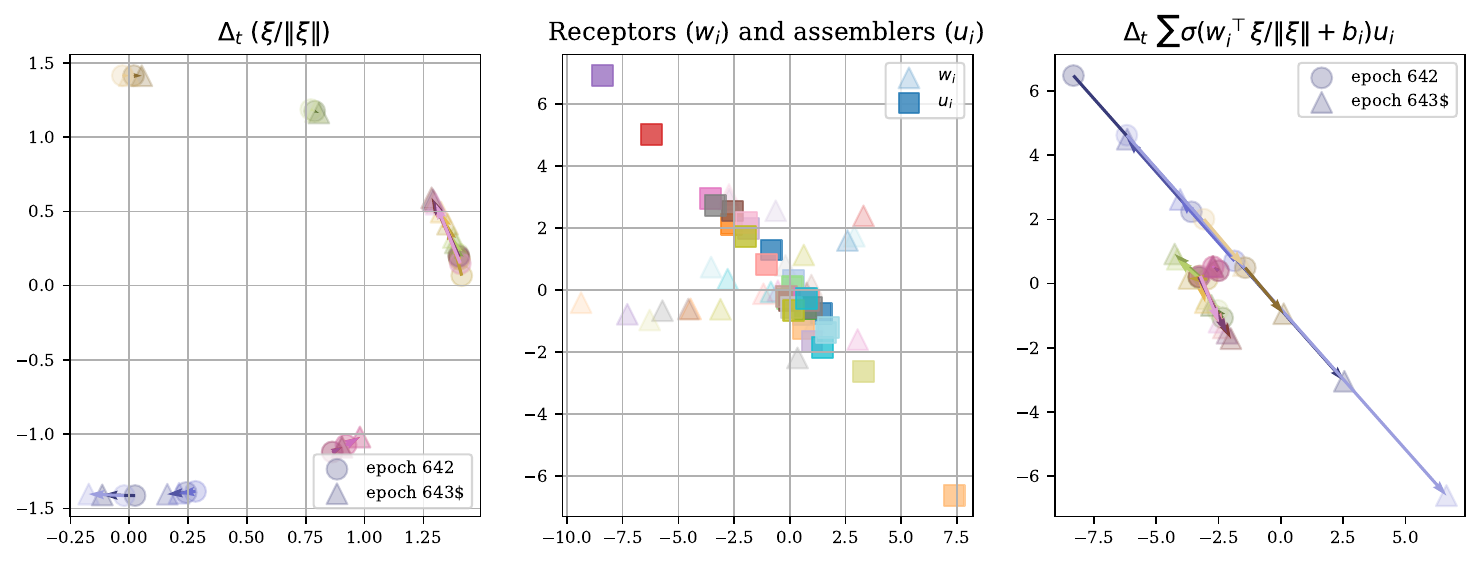}
    \caption{
    A small change in sequence embeddings (left) can lead to a big change in MLP response (right).
    This is due to heavy or small but heavily correlated assemblers (middle).
    }
    \label{fig:spike-details}
\end{figure}

Better understanding why loss spikes occur is of great interest as they can be detrimental to the training stability, particularly with very large-scale transformers such as large language models~\citep{chowdhery2023palm}. While some know-how exists for mitigating the impact of peak losses, it is often experimentally motivated and requires a lot of computational resources to achieve operational success~\citep{grattafiori2024llama3herdmodels}. Our visual inspection showcases two aspects leading to loss spikes: the high curvature of the RMS normalization layer near the origin, as well as the high curvature of multi-layer perceptrons with heavy weights, or with numerous small correlated weights.
\begin{figure}[!h]
    \centering
    \includegraphics[width=.8\linewidth]{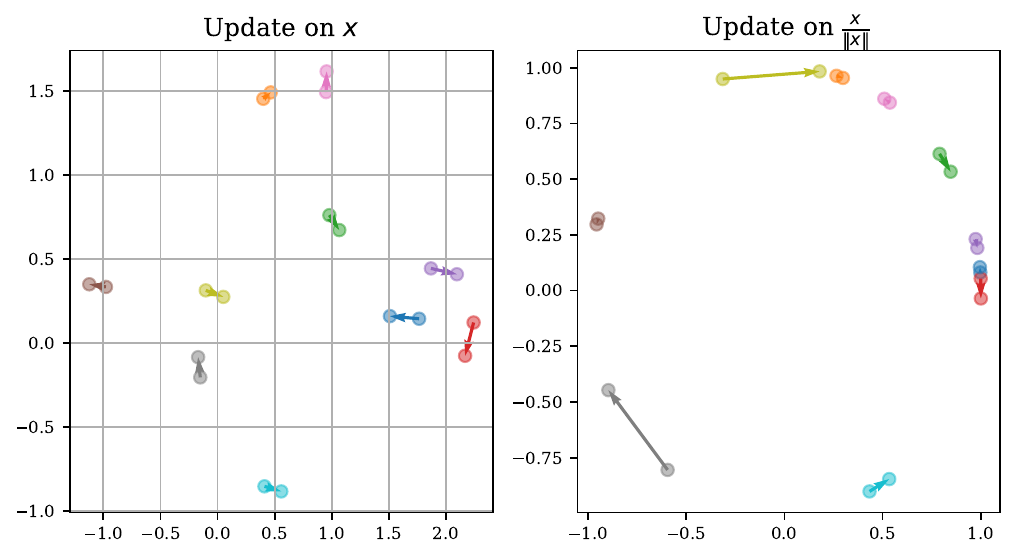}
    \caption{
    \textbf{High-curvature normalization}. Loss spikes are linked to the high curvature of internal network functions. A small update to an element $x$ can result in a substantial change to its normalized version $x/\norm{x}$, significantly altering the network's subsequent behavior.
    }
    \label{fig:spike-norm}
\end{figure}
\paragraph{High-curvature normalization.} \cref{fig:spike-norm} illustrates that a small modification to an element can result in a disproportionately large change in its normalized version.
In theory, the gradient points towards directions that would reduce the training loss. 
However, considering a large step size in these directions could be counterproductive. 
This is especially true for functions with high curvature, such as the normalization layer $f(x) = x / \norm{x}$ near the origin.
At any point \( x_t \), the gradient descent update rule suggests that one can update \( x_{t+1} \) as \( x_t - \eta_t u_t \) without changing \( f(x_t) \), where \( u_t = x_t / \|x_t\| \) and \( \eta_t \) is the learning rate.
This holds only if the learning is small enough, \( \eta_t < \|x_t\| \).
When \( x_t \) is close to zero, ensuring \( \eta_t < \|x_t\| \) becomes challenging, particularly if the step size \( \eta_t \) was predetermined by some scheduler. 
This behavior is related to the "edge-of-stability" phenomenon highlighted by \citet{cohen2022gradientdescentneuralnetworks}, further theoretical insights being provided by \citet{cabannes2024learning}.
Interestingly, this analysis suggests removing some loss spikes by smoothing out the normalization layer.
For example, consider using $f(x) = \sigma(\norm{x}) x / \norm{x}$ where $\sigma$ is a smooth function with $\sigma(0) = 0$ and $\sigma([1, \infty)) = \{1\}$.
This demonstrates the usefulness of our visual sandbox in gaining insights and building intuition, which can then be validated on a larger scale in subsequent works.
\begin{figure}[!h]
    \centering
    \includegraphics[width=0.49\linewidth]{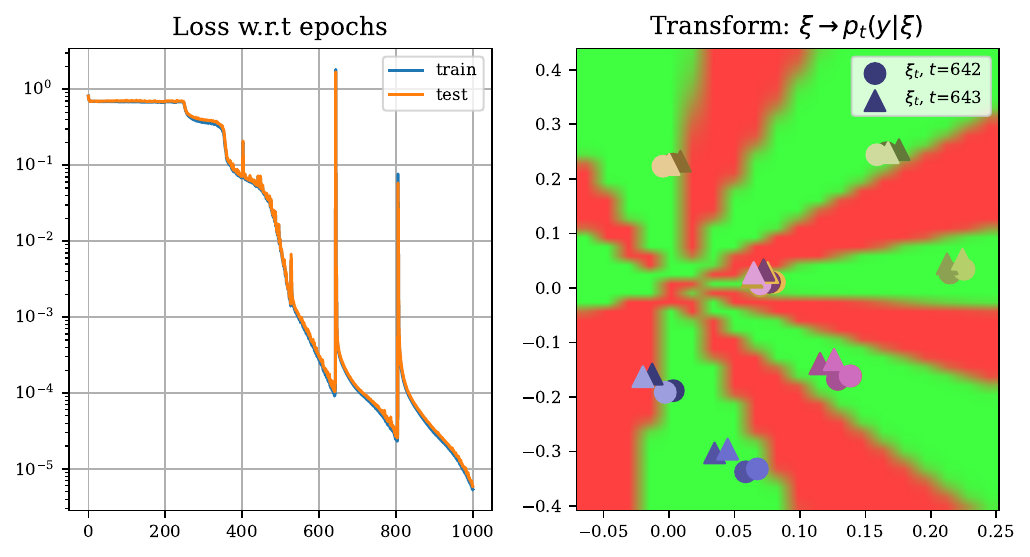}
    \hfill
    \includegraphics[width=0.49\linewidth]{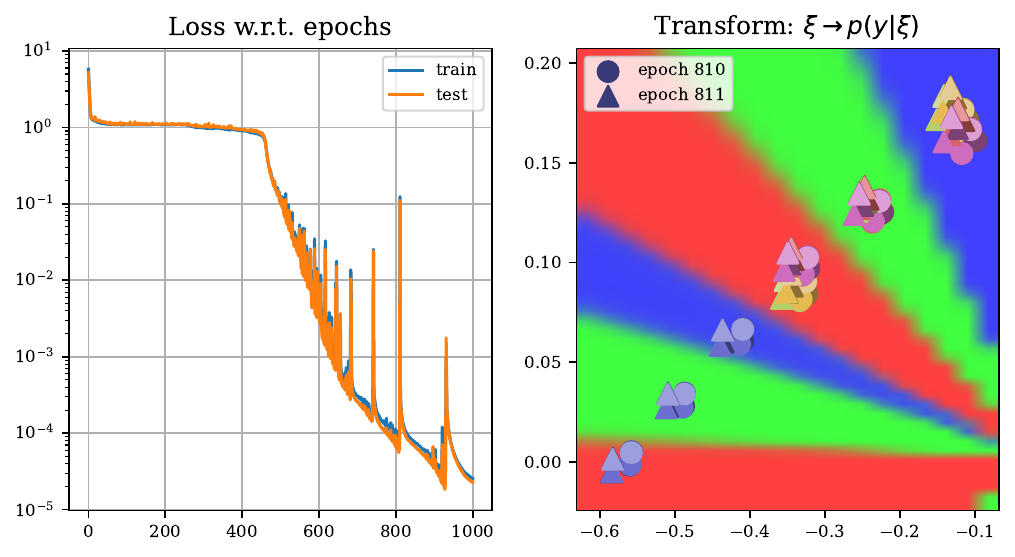}
    \caption{
    \textbf{High-curvature feed-forward}. Loss spikes (both left) resulting from a small change from one iteration to another in sequence embeddings that are close to the decision boundaries of the subsequent feedforward layer (both right).}
    \label{fig:loss-spikes}
\end{figure}
\paragraph{High-curvature feed-forward.} Another source of loss spikes is illustrated in \cref{fig:loss-spikes}.
They are due to the decision boundaries of the feedforward layer being quite close to the sequence embeddings, meaning that a small change in the sequence embedding can lead to a great change in their classification.
This is again due to the high curvature of the MLP layer, as illustrated in \cref{fig:spike-details}.
In particular, the heavy, or the small but heavily correlated, weights in the MLP cause the response $\zeta$ to vary highly as a function of $\zeta$.
Once again, one can imagine different ways to regularize these types of loss spikes, with various regularization measures, or by ensuring that the capacity of the MLP is large enough for the MLP to avoid creating these heavy or highly correlated weights.

\begin{figure}[!h]
    \centering    
    \includegraphics[width=\linewidth]{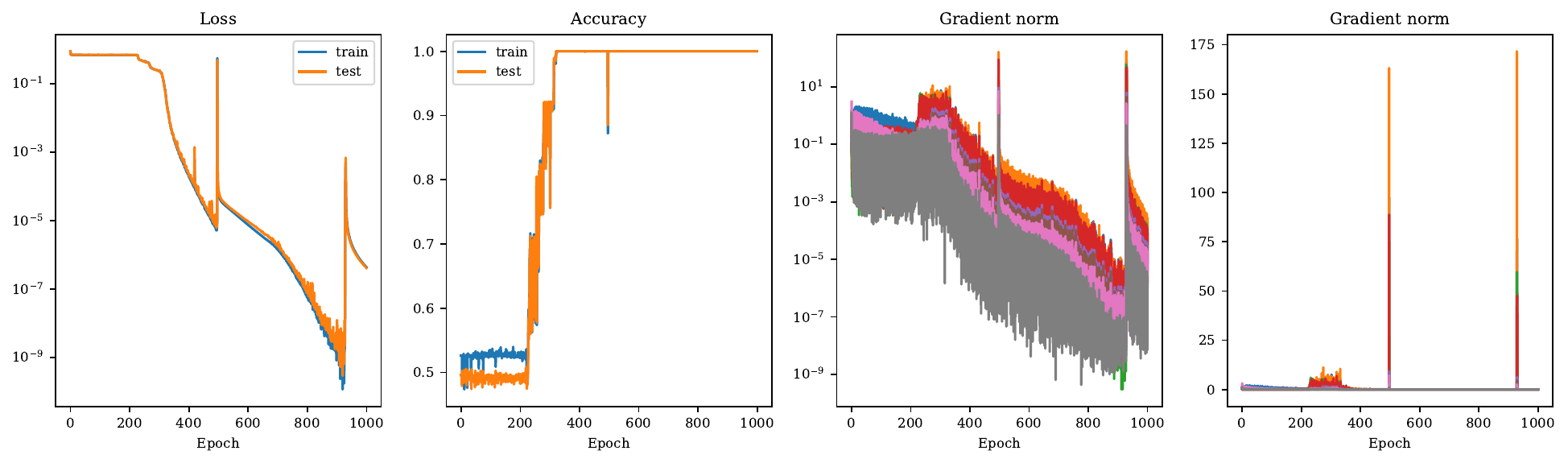}
    \caption{\textbf{Loss spikes and gradient norms.} Akin to \cref{fig:grad-norms-full-batch} in mini-batch, this experiment makes the connection between gradient norm and loss spikes salient.}
    \label{fig:grad-norms}
\end{figure}

\paragraph{Connection to gradient norms.} Loss spikes and high gradient norms appear in tandem as can be seen in~\cref{fig:grad-norms}, indicating that when a too-large step size deviates the model from its current small loss region, it is taken back to where it was with large updates. It motivates further study of the Hessian of the loss, which is promising in the literature~\citep{ilbert2024samformer, foret2021sharpnessaware, zhang2024transformersneedadamhessian, gomes2025adafisher}. 

\paragraph{Experiments in larger dimensions.} When the embedding dimension $d$ is bigger, we cannot represent the tokens in the plane. However, we can plot the evolution of other metrics, such as the loss, the gradient norms, or the activation sparsity. We replicate the experiments of ~\cref{sec:training_dyn} with $d > 2$ in~\cref{app:emb_dim} and obtain similar findings, which are also coherent with the existing literature in more practical settings~\citep{li2023thelazy,mirzadeh2024relu,zhang2024transformersneedadamhessian, ilbert2024samformer}.

\paragraph{Implication for large language models.} Ensuring the stability of training is key to reducing budget and computational costs. A common issue during pretraining is the occurrence of loss spikes~\citep[see]{chowdhery2024palm, marin,li2025trainingdynamics17bllama, takase2025spike}, which can lead to loss divergence. Better understanding their root cause is crucial to reducing the cost of training large models, e.g., through the design of novel optimizers. Our findings suggest that reducing gradient norms and avoiding embeddings with small norms can help stabilize the training. This resonates with practical heuristics such as QK-Norm~\citep{dehghani2023scaling, wortsman2024smallscale} or MuonClip~\citep{team2025kimi} that can be seen as a way to enforce small gradient norms by avoiding exploding attention logits, while imposing a unit norm to embedding vectors, thus mitigating the high-curvature region of normalization layers, is beneficial for both the training stability and the sample efficiency~\citep{loshchilov2025ngpt}. Another promising avenue is the use of second-order algorithms such as Shampoo~\citep{gupta2018shampoo}, Muon~\citep{jordan2024muon}, and AdaFisher~\citep{gomes2025adafisher}, which allow for better control over the gradient variations.

\section{Related work}
\label{app:related_work}
\paragraph{Mechanistic interpretability.} 
Neural networks are often seen as \say{black boxes} whose internal computations are hard to decipher. Researchers have developed various methods, such as extracting meaningful features from neural network activations \citep{fel2023lens}, and assessing the impact of perturbations on model inputs \citep{koh2017influence,fel2021variance}, among others. Methods have also been proposed to discover circuits in neural networks automatically~\citep{conmy2023towards}.
Recently, the field of mechanistic interpretability has advocated for exposing the internal mechanisms of transformers to provide novel insights into their capabilities \citep{elhage2021mathematical,olsson2022induction}, by either relying on interventions on large models~\citep{geva2023dissecting, wang2023interpretability, meng2022locating}, or by conducting precise ablation studies in a controlled setup~\citep{charton2022mathtransformerdoing,liu2022towards,nanda2023progress,cabannes2024iteration, bietti2023birth}. Our work aligns with the latter, aiming to make the internal behavior of transformers more explicit through carefully selected visualizations in controlled settings, with a focus on their training dynamics. 

\paragraph{Training dynamics in neural networks.} With non-convex losses and models of increasing sizes, understanding how models learn is more and more difficult. One approach to address this issue is to employ mathematical abstractions, as seen with neural tangent kernels (NTK) \cite{jacot2018ntk,chizat2020lazytrainingdifferentiableprogramming} and mean-field analysis \citep{Mei_2018,chen2024generalization}, among others \citep{abbe2022mergedstaircase,ahn2024linear,cabannes2024learning}. Obtaining formal results often requires deviating from the practical implementation~\citep{bietti2023birth, cabannes2024learning, boix_adsera_2023_gradual, tian2023scan, ahn2023transformers, geshkovski2023clusters, jelassi2022vision, mahankali2024one}. In our work, we keep the main transformer block components intact to ensure that our findings have practical implications.

\subsection{Discussion}
This paper aims to advance our understanding of the transformer's training dynamics through a detailed study in a controlled and interpretable setting. We show how transformers trained with gradient descent can solve the sparse modular addition by capturing invariants, as is often the case in NLP and reasoning. Our visual sandbox allows us to monitor the evolution of tokens during training. We use it to uncover the stage-wise learning and the occurrence of loss spikes. By connecting our findings with more practical settings, we show the benefits of our sandbox to study complex behavior at a small scale. We hope it can be used to build intuition towards improving current training pipelines.

\paragraph{Limitations.} Our work is a controlled study of a single transformer block, which allows both theory and experiments towards a better understanding of transformer training. In future work, we will aim to extend our visualization tools to larger models.

\section*{Impact Statement}
This paper presents work whose goal is to advance the field of Machine
Learning. There are many potential societal consequences of our work, none
which we feel must be specifically highlighted here.

\bibliography{references}

@inproceedings{abbe2023provable,
title={Provable Advantage of Curriculum Learning on Parity Targets with Mixed Inputs},
author={Emmanuel Abbe and Elisabetta Cornacchia and Aryo Lotfi},
booktitle={Thirty-seventh Conference on Neural Information Processing Systems},
year={2023},
url={https://openreview.net/forum?id=9Ihu0VBOTq}
}

@misc{barak2023hidden,
      title={Hidden Progress in Deep Learning: {SGD} Learns Parities Near the Computational Limit}, 
      author={Boaz Barak and Benjamin L. Edelman and Surbhi Goel and Sham Kakade and Eran Malach and Cyril Zhang},
      year={2023},
      eprint={2207.08799},
      archivePrefix={arXiv},
      primaryClass={cs.LG}
}

@misc{kingma2017adam,
      title={Adam: A Method for Stochastic Optimization}, 
      author={Diederik P. Kingma and Jimmy Ba},
      year={2017},
      eprint={1412.6980},
      archivePrefix={arXiv},
      primaryClass={cs.LG}
}

@misc{paszke2019pytorch,
      title={{PyTorch}: An Imperative Style, High-Performance Deep Learning Library}, 
      author={Adam Paszke and Sam Gross and Francisco Massa and Adam Lerer and James Bradbury and Gregory Chanan and Trevor Killeen and Zeming Lin and Natalia Gimelshein and Luca Antiga and Alban Desmaison and Andreas Köpf and Edward Yang and Zach DeVito and Martin Raison and Alykhan Tejani and Sasank Chilamkurthy and Benoit Steiner and Lu Fang and Junjie Bai and Soumith Chintala},
      year={2019},
      eprint={1912.01703},
      archivePrefix={arXiv},
      primaryClass={cs.LG}
}

@misc{power2022grokking,
      title={Grokking: Generalization Beyond Overfitting on Small Algorithmic Datasets}, 
      author={Alethea Power and Yuri Burda and Harri Edwards and Igor Babuschkin and Vedant Misra},
      year={2022},
      eprint={2201.02177},
      archivePrefix={arXiv},
      primaryClass={cs.LG},
}

@inproceedings{zhang2019root,
 author = {Zhang, Biao and Sennrich, Rico},
 booktitle = {Advances in Neural Information Processing Systems},
 editor = {H. Wallach and H. Larochelle and A. Beygelzimer and F. d\textquotesingle Alch\'{e}-Buc and E. Fox and R. Garnett},
 pages = {},
 publisher = {Curran Associates, Inc.},
 title = {Root Mean Square Layer Normalization},
 url = {https://proceedings.neurips.cc/paper_files/paper/2019/file/1e8a19426224ca89e83cef47f1e7f53b-Paper.pdf},
 volume = {32},
 year = {2019}
}

@misc{abbe2022mergedstaircase,
      title={The merged-staircase property: a necessary and nearly sufficient condition for {SGD} learning of sparse functions on two-layer neural networks}, 
      author={Emmanuel Abbe and Enric Boix-Adsera and Theodor Misiakiewicz},
      year={2022},
      eprint={2202.08658},
      archivePrefix={arXiv},
      primaryClass={cs.LG}
}

@inproceedings{bietti2023birth,
title={Birth of a Transformer: A Memory Viewpoint},
author={Alberto Bietti and Vivien Cabannes and Diane Bouchacourt and Herve Jegou and Leon Bottou},
booktitle={Thirty-seventh Conference on Neural Information Processing Systems},
year={2023},
url={https://openreview.net/forum?id=3X2EbBLNsk}
}

@misc{cabannes2024scaling,
      title={Scaling Laws for Associative Memories}, 
      author={Vivien Cabannes and Elvis Dohmatob and Alberto Bietti},
      year={2024},
      eprint={2310.02984},
      archivePrefix={arXiv},
      primaryClass={stat.ML}
}

@misc{cabannes2024learning,
      title={Learning Associative Memories with Gradient Descent}, 
      author={Vivien Cabannes and Berfin Simsek and Alberto Bietti},
      year={2024},
      eprint={2402.18724},
      archivePrefix={arXiv},
      primaryClass={cs.LG}
}

@inproceedings{edelman2024evolution,
title={The Evolution of Statistical Induction Heads: In-Context Learning Markov Chains},
author={Ezra Edelman and Nikolaos Tsilivis and Benjamin L. Edelman and eran malach and Surbhi Goel},
booktitle={The Thirty-eighth Annual Conference on Neural Information Processing Systems},
year={2024},
url={https://openreview.net/forum?id=qaRT6QTIqJ}
}

@article{elhage2021mathematical,
   title={A Mathematical Framework for Transformer Circuits},
   author={Elhage, Nelson and Nanda, Neel and Olsson, Catherine and Henighan, Tom and Joseph, Nicholas and Mann, Ben and Askell, Amanda and Bai, Yuntao and Chen, Anna and Conerly, Tom and DasSarma, Nova and Drain, Dawn and Ganguli, Deep and Hatfield-Dodds, Zac and Hernandez, Danny and Jones, Andy and Kernion, Jackson and Lovitt, Liane and Ndousse, Kamal and Amodei, Dario and Brown, Tom and Clark, Jack and Kaplan, Jared and McCandlish, Sam and Olah, Chris},
   year={2021},
   journal={Transformer Circuits Thread}
}

@misc{geva2023dissecting,
      title={Dissecting Recall of Factual Associations in Auto-Regressive Language Models}, 
      author={Mor Geva and Jasmijn Bastings and Katja Filippova and Amir Globerson},
      year={2023},
      eprint={2304.14767},
      archivePrefix={arXiv},
      primaryClass={cs.CL}
}

@inproceedings{vaswani2017attention,
 author = {Vaswani, Ashish and Shazeer, Noam and Parmar, Niki and Uszkoreit, Jakob and Jones, Llion and Gomez, Aidan N and Kaiser, \L ukasz and Polosukhin, Illia},
 booktitle = {Advances in Neural Information Processing Systems},
 editor = {I. Guyon and U. Von Luxburg and S. Bengio and H. Wallach and R. Fergus and S. Vishwanathan and R. Garnett},
 publisher = {Curran Associates, Inc.},
 title = {Attention is All you Need},
 url = {https://proceedings.neurips.cc/paper_files/paper/2017/file/3f5ee243547dee91fbd053c1c4a845aa-Paper.pdf},
 volume = {30},
 year = {2017}
}

@article{olsson2022induction,
   title={In-context Learning and Induction Heads},
   author={Olsson, Catherine and Elhage, Nelson and Nanda, Neel and Joseph, Nicholas and DasSarma, Nova and Henighan, Tom and Mann, Ben and Askell, Amanda and Bai, Yuntao and Chen, Anna and Conerly, Tom and Drain, Dawn and Ganguli, Deep and Hatfield-Dodds, Zac and Hernandez, Danny and Johnston, Scott and Jones, Andy and Kernion, Jackson and Lovitt, Liane and Ndousse, Kamal and Amodei, Dario and Brown, Tom and Clark, Jack and Kaplan, Jared and McCandlish, Sam and Olah, Chris},
   year={2022},
   journal={Transformer Circuits Thread},
   note={https://transformer-circuits.pub/2022/in-context-learning-and-induction-heads/index.html}
}

@misc{hoffmann2022trainingcomputeoptimallargelanguage,
      title={Training Compute-Optimal Large Language Models}, 
      author={Jordan Hoffmann and Sebastian Borgeaud and Arthur Mensch and Elena Buchatskaya and Trevor Cai and Eliza Rutherford and Diego de Las Casas and Lisa Anne Hendricks and Johannes Welbl and Aidan Clark and Tom Hennigan and Eric Noland and Katie Millican and George van den Driessche and Bogdan Damoc and Aurelia Guy and Simon Osindero and Karen Simonyan and Erich Elsen and Jack W. Rae and Oriol Vinyals and Laurent Sifre},
      year={2022},
      eprint={2203.15556},
      archivePrefix={arXiv},
      primaryClass={cs.CL},
      url={https://arxiv.org/abs/2203.15556}, 
}

@misc{kaplan2020scalinglawsneurallanguage,
      title={Scaling Laws for Neural Language Models}, 
      author={Jared Kaplan and Sam McCandlish and Tom Henighan and Tom B. Brown and Benjamin Chess and Rewon Child and Scott Gray and Alec Radford and Jeffrey Wu and Dario Amodei},
      year={2020},
      eprint={2001.08361},
      archivePrefix={arXiv},
      primaryClass={cs.LG},
      url={https://arxiv.org/abs/2001.08361}, 
}

@inproceedings{wortsman2024smallscale,
title={Small-scale proxies for large-scale Transformer training instabilities},
author={Mitchell Wortsman and Peter J Liu and Lechao Xiao and Katie E Everett and Alexander A Alemi and Ben Adlam and John D Co-Reyes and Izzeddin Gur and Abhishek Kumar and Roman Novak and Jeffrey Pennington and Jascha Sohl-Dickstein and Kelvin Xu and Jaehoon Lee and Justin Gilmer and Simon Kornblith},
booktitle={The Twelfth International Conference on Learning Representations},
year={2024},
url={https://openreview.net/forum?id=d8w0pmvXbZ}
}

@article{chowdhery2024palm,
author = {Chowdhery, Aakanksha and Narang, Sharan and Devlin, Jacob and Bosma, Maarten and Mishra, Gaurav and Roberts, Adam and Barham, Paul and Chung, Hyung Won and Sutton, Charles and Gehrmann et al.},
title = {PaLM: scaling language modeling with pathways},
year = {2024},
issue_date = {January 2023},
publisher = {JMLR.org},
volume = {24},
number = {1},
issn = {1532-4435},
journal = {J. Mach. Learn. Res.},
month = {mar},
articleno = {240},
numpages = {113}
}

@InProceedings{dehghani2023scaling,
  title = 	 {Scaling Vision Transformers to 22 Billion Parameters},
  author =       {Dehghani, Mostafa and Djolonga, Josip and Mustafa, Basil and Padlewski, Piotr and Heek, Jonathan and Gilmer, Justin and Steiner, Andreas Peter and Caron, Mathilde and Geirhos, Robert and Alabdulmohsin, Ibrahim and Jenatton, Rodolphe and Beyer, Lucas and Tschannen, Michael and Arnab, Anurag and Wang, Xiao and Riquelme Ruiz et al.},
  booktitle = 	 {Proceedings of the 40th International Conference on Machine Learning},
  pages = 	 {7480--7512},
  year = 	 {2023},
  editor = 	 {Krause, Andreas and Brunskill, Emma and Cho, Kyunghyun and Engelhardt, Barbara and Sabato, Sivan and Scarlett, Jonathan},
  volume = 	 {202},
  series = 	 {Proceedings of Machine Learning Research},
  month = 	 {23--29 Jul},
  publisher =    {PMLR},
  url = 	 {https://proceedings.mlr.press/v202/dehghani23a.html},
}

@inproceedings{ahn2024linear,
title={Linear attention is (maybe) all you need (to understand Transformer optimization)},
author={Kwangjun Ahn and Xiang Cheng and Minhak Song and Chulhee Yun and Ali Jadbabaie and Suvrit Sra},
booktitle={The Twelfth International Conference on Learning Representations},
year={2024},
url={https://openreview.net/forum?id=0uI5415ry7}
}

@inproceedings{chen2024generalization,
title={Generalization of Scaled Deep ResNets in the Mean-Field Regime},
author={Yihang Chen and Fanghui Liu and Yiping Lu and Grigorios Chrysos and Volkan Cevher},
booktitle={The Twelfth International Conference on Learning Representations},
year={2024},
url={https://openreview.net/forum?id=tMzPZTvz2H}
}

@inproceedings{jacot2018ntk,
 author = {Jacot, Arthur and Gabriel, Franck and Hongler, Clement},
 booktitle = {Advances in Neural Information Processing Systems},
 editor = {S. Bengio and H. Wallach and H. Larochelle and K. Grauman and N. Cesa-Bianchi and R. Garnett},
 pages = {},
 publisher = {Curran Associates, Inc.},
 title = {Neural Tangent Kernel: Convergence and Generalization in Neural Networks},
 url = {https://proceedings.neurips.cc/paper_files/paper/2018/file/5a4be1fa34e62bb8a6ec6b91d2462f5a-Paper.pdf},
 volume = {31},
 year = {2018}
}

@inproceedings{nanda2023progress,
title={Progress measures for grokking via mechanistic interpretability},
author={Neel Nanda and Lawrence Chan and Tom Lieberum and Jess Smith and Jacob Steinhardt},
booktitle={The Eleventh International Conference on Learning Representations },
year={2023},
url={https://openreview.net/forum?id=9XFSbDPmdW}
}

@inproceedings{cabannes2024iteration,
title={Iteration Head: A Mechanistic Study of Chain-of-Thought},
author={Vivien Cabannes and Charles Arnal and Wassim Bouaziz and Xingyu Alice Yang and Francois Charton and Julia Kempe},
booktitle={ICML 2024 Workshop on Mechanistic Interpretability},
year={2024},
url={https://openreview.net/forum?id=D66dtunCnP}
}

@article{olah2022lowdim,
   title={Mechanistic Interpretability, Variables, and the Importance of Interpretable Bases},
   author={Olah, Chris},
   year={2023},
   journal={Transformer Circuits Thread},
   url={https://transformer-circuits.pub/2022/mech-interp-essay/index.html}
}

@inproceedings{fel2023lens,
title={A Holistic Approach to Unifying Automatic Concept Extraction and Concept Importance Estimation},
author={Thomas Fel and Victor Boutin and Louis B{\'e}thune and Remi Cadene and Mazda Moayeri and L{\'e}o And{\'e}ol and Mathieu Chalvidal and Thomas Serre},
booktitle={Thirty-seventh Conference on Neural Information Processing Systems},
year={2023},
url={https://openreview.net/forum?id=MziFFGjpkb}
}

@inproceedings{fel2021variance,
 author = {Fel, Thomas and Cadene, Remi and Chalvidal, Mathieu and Cord, Matthieu and Vigouroux, David and Serre, Thomas},
 booktitle = {Advances in Neural Information Processing Systems},
 editor = {M. Ranzato and A. Beygelzimer and Y. Dauphin and P.S. Liang and J. Wortman Vaughan},
 pages = {26005--26014},
 publisher = {Curran Associates, Inc.},
 title = {Look at the Variance! Efficient Black-box Explanations with Sobol-based Sensitivity Analysis},
 url = {https://proceedings.neurips.cc/paper_files/paper/2021/file/da94cbeff56cfda50785df477941308b-Paper.pdf},
 volume = {34},
 year = {2021}
}

@InProceedings{koh2017influence,
  title = 	 {Understanding Black-box Predictions via Influence Functions},
  author =       {Pang Wei Koh and Percy Liang},
  booktitle = 	 {Proceedings of the 34th International Conference on Machine Learning},
  pages = 	 {1885--1894},
  year = 	 {2017},
  editor = 	 {Precup, Doina and Teh, Yee Whye},
  volume = 	 {70},
  series = 	 {Proceedings of Machine Learning Research},
  month = 	 {06--11 Aug},
  publisher =    {PMLR},
  url = 	 {https://proceedings.mlr.press/v70/koh17a.html},
}

@misc{charton2022mathtransformerdoing,
      title={What is my math transformer doing? -- Three results on interpretability and generalization}, 
      author={François Charton},
      year={2022},
      eprint={2211.00170},
      archivePrefix={arXiv},
      primaryClass={cs.LG},
      url={https://arxiv.org/abs/2211.00170}, 
}

@inproceedings{liu2022towards,
title={Towards Understanding Grokking: An Effective Theory of Representation Learning},
author={Ziming Liu and Ouail Kitouni and Niklas Nolte and Eric J Michaud and Max Tegmark and Mike Williams},
booktitle={Advances in Neural Information Processing Systems},
editor={Alice H. Oh and Alekh Agarwal and Danielle Belgrave and Kyunghyun Cho},
year={2022},
url={https://openreview.net/forum?id=6at6rB3IZm}
}

@inproceedings{meng2022locating,
title={Locating and Editing Factual Associations in {GPT}},
author={Kevin Meng and David Bau and Alex J Andonian and Yonatan Belinkov},
booktitle={Advances in Neural Information Processing Systems},
editor={Alice H. Oh and Alekh Agarwal and Danielle Belgrave and Kyunghyun Cho},
year={2022},
url={https://openreview.net/forum?id=-h6WAS6eE4}
}

@inproceedings{wang2023interpretability,
title={Interpretability in the Wild: a Circuit for Indirect Object Identification in {GPT}-2 Small},
author={Kevin Ro Wang and Alexandre Variengien and Arthur Conmy and Buck Shlegeris and Jacob Steinhardt},
booktitle={The Eleventh International Conference on Learning Representations },
year={2023},
url={https://openreview.net/forum?id=NpsVSN6o4ul}
}

@misc{chizat2020lazytrainingdifferentiableprogramming,
      title={On Lazy Training in Differentiable Programming}, 
      author={Lenaic Chizat and Edouard Oyallon and Francis Bach},
      year={2020},
      eprint={1812.07956},
      archivePrefix={arXiv},
      primaryClass={math.OC},
      url={https://arxiv.org/abs/1812.07956}, 
}

@article{Mei_2018,
   title={A mean field view of the landscape of two-layer neural networks},
   volume={115},
   ISSN={1091-6490},
   url={http://dx.doi.org/10.1073/pnas.1806579115},
   DOI={10.1073/pnas.1806579115},
   number={33},
   journal={Proceedings of the National Academy of Sciences},
   publisher={Proceedings of the National Academy of Sciences},
   author={Mei, Song and Montanari, Andrea and Nguyen, Phan-Minh},
   year={2018},
   month=jul }

@misc{dauphin2014identifyingattackingsaddlepoint,
      title={Identifying and attacking the saddle point problem in high-dimensional non-convex optimization}, 
      author={Yann Dauphin and Razvan Pascanu and Caglar Gulcehre and Kyunghyun Cho and Surya Ganguli and Yoshua Bengio},
      year={2014},
      eprint={1406.2572},
      archivePrefix={arXiv},
      primaryClass={cs.LG},
      url={https://arxiv.org/abs/1406.2572}, 
}

@misc{touvron2023llamaopenefficientfoundation,
      title={LLaMA: Open and Efficient Foundation Language Models}, 
      author={Hugo Touvron and Thibaut Lavril and Gautier Izacard and Xavier Martinet and Marie-Anne Lachaux and Timothée Lacroix and Baptiste Rozière and Naman Goyal and Eric Hambro and Faisal Azhar and Aurelien Rodriguez and Armand Joulin and Edouard Grave and Guillaume Lample},
      year={2023},
      eprint={2302.13971},
      archivePrefix={arXiv},
      primaryClass={cs.CL},
      url={https://arxiv.org/abs/2302.13971}, 
}

@misc{dubey2024llama3herdmodels,
      title={The Llama 3 Herd of Models}, 
      author={{Llama Team}},
      year={2024},
      eprint={2407.21783},
      archivePrefix={arXiv},
      primaryClass={cs.AI},
      url={https://arxiv.org/abs/2407.21783}, 
}

@article{Chi_2019,
   title={Nonconvex Optimization Meets Low-Rank Matrix Factorization: An Overview},
   volume={67},
   ISSN={1941-0476},
   url={http://dx.doi.org/10.1109/TSP.2019.2937282},
   DOI={10.1109/tsp.2019.2937282},
   number={20},
   journal={IEEE Transactions on Signal Processing},
   publisher={Institute of Electrical and Electronics Engineers (IEEE)},
   author={Chi, Yuejie and Lu, Yue M. and Chen, Yuxin},
   year={2019},
   month=oct, pages={5239–5269} 
}

@misc{du2017gradientdescentexponentialtime,
      title={Gradient Descent Can Take Exponential Time to Escape Saddle Points}, 
      author={Simon S. Du and Chi Jin and Jason D. Lee and Michael I. Jordan and Barnabas Poczos and Aarti Singh},
      year={2017},
      eprint={1705.10412},
      archivePrefix={arXiv},
      primaryClass={math.OC},
      url={https://arxiv.org/abs/1705.10412}, 
}

@misc{cohen2022gradientdescentneuralnetworks,
      title={Gradient Descent on Neural Networks Typically Occurs at the Edge of Stability}, 
      author={Jeremy M. Cohen and Simran Kaur and Yuanzhi Li and J. Zico Kolter and Ameet Talwalkar},
      year={2022},
      eprint={2103.00065},
      archivePrefix={arXiv},
      primaryClass={cs.LG},
      url={https://arxiv.org/abs/2103.00065}, 
}

@InProceedings{ilbert2024samformer,
  title = 	 {{SAM}former: Unlocking the Potential of Transformers in Time Series Forecasting with Sharpness-Aware Minimization and Channel-Wise Attention},
  author =       {Ilbert, Romain and Odonnat, Ambroise and Feofanov, Vasilii and Virmaux, Aladin and Paolo, Giuseppe and Palpanas, Themis and Redko, Ievgen},
  booktitle = 	 {Proceedings of the 41st International Conference on Machine Learning},
  pages = 	 {20924--20954},
  year = 	 {2024},
  editor = 	 {Salakhutdinov, Ruslan and Kolter, Zico and Heller, Katherine and Weller, Adrian and Oliver, Nuria and Scarlett, Jonathan and Berkenkamp, Felix},
  volume = 	 {235},
  series = 	 {Proceedings of Machine Learning Research},
  month = 	 {21--27 Jul},
  publisher =    {PMLR},
  url = 	 {https://proceedings.mlr.press/v235/ilbert24a.html},
}

@inproceedings{zhang2024transformersneedadamhessian,
title={Why Transformers Need Adam: A Hessian Perspective},
author={Yushun Zhang and Congliang Chen and Tian Ding and Ziniu Li and Ruoyu Sun and Zhi-Quan Luo},
booktitle={The Thirty-eighth Annual Conference on Neural Information Processing Systems},
year={2024},
url={https://openreview.net/forum?id=X6rqEpbnj3}
}

@inproceedings{foret2021sharpnessaware,
title={Sharpness-aware Minimization for Efficiently Improving Generalization},
author={Pierre Foret and Ariel Kleiner and Hossein Mobahi and Behnam Neyshabur},
booktitle={International Conference on Learning Representations},
year={2021},
url={https://openreview.net/forum?id=6Tm1mposlrM}
}

@inproceedings{mairal2009sparse,
author = {Mairal, Julien and Bach, Francis and Ponce, Jean and Sapiro, Guillermo},
title = {Online dictionary learning for sparse coding},
year = {2009},
isbn = {9781605585161},
publisher = {Association for Computing Machinery},
address = {New York, NY, USA},
url = {https://doi.org/10.1145/1553374.1553463},
doi = {10.1145/1553374.1553463},
booktitle = {Proceedings of the 26th Annual International Conference on Machine Learning},
pages = {689–696},
numpages = {8},
location = {Montreal, Quebec, Canada},
series = {ICML '09}
}

@article{chen1998atomic,
author = {Chen, Scott Shaobing and Donoho, David L. and Saunders, Michael A.},
title = {Atomic Decomposition by Basis Pursuit},
journal = {SIAM Journal on Scientific Computing},
volume = {20},
number = {1},
pages = {33-61},
year = {1998},
doi = {10.1137/S1064827596304010},
URL = {https://doi.org/10.1137/S1064827596304010},
eprint = {https://doi.org/10.1137/S1064827596304010},
}

@article{barth2012neocortex,
title = {Experimental evidence for sparse firing in the neocortex},
journal = {Trends in Neurosciences},
volume = {35},
number = {6},
pages = {345-355},
year = {2012},
issn = {0166-2236},
doi = {https://doi.org/10.1016/j.tins.2012.03.008},
url = {https://www.sciencedirect.com/science/article/pii/S0166223612000513},
author = {Alison L. Barth and James F.A. Poulet},
}

@ARTICLE{fukushima1969relu,
  author={Fukushima, Kunihiko},
  journal={IEEE Transactions on Systems Science and Cybernetics}, 
  title={Visual Feature Extraction by a Multilayered Network of Analog Threshold Elements}, 
  year={1969},
  volume={5},
  number={4},
  pages={322-333},
  doi={10.1109/TSSC.1969.300225}}

@article{elwing2018silu,
title = {Sigmoid-weighted linear units for neural network function approximation in reinforcement learning},
journal = {Neural Networks},
volume = {107},
pages = {3-11},
year = {2018},
note = {Special issue on deep reinforcement learning},
issn = {0893-6080},
doi = {https://doi.org/10.1016/j.neunet.2017.12.012},
url = {https://www.sciencedirect.com/science/article/pii/S0893608017302976},
author = {Stefan Elfwing and Eiji Uchibe and Kenji Doya}
}

@inproceedings{mirzadeh2024relu,
title={Re{LU} Strikes Back: Exploiting Activation Sparsity in Large Language Models},
author={Seyed Iman Mirzadeh and Keivan Alizadeh-Vahid and Sachin Mehta and Carlo C del Mundo and Oncel Tuzel and Golnoosh Samei and Mohammad Rastegari and Mehrdad Farajtabar},
booktitle={The Twelfth International Conference on Learning Representations},
year={2024},
url={https://openreview.net/forum?id=osoWxY8q2E}
}

@inproceedings{li2023lazy,
title={The Lazy Neuron Phenomenon: On Emergence of Activation Sparsity in Transformers},
author={Zonglin Li and Chong You and Srinadh Bhojanapalli and Daliang Li and Ankit Singh Rawat and Sashank J. Reddi and Ke Ye and Felix Chern and Felix Yu and Ruiqi Guo and Sanjiv Kumar},
booktitle={The Eleventh International Conference on Learning Representations },
year={2023},
url={https://openreview.net/forum?id=TJ2nxciYCk-}
}

@misc{touvron2023llama,
    title={LLaMA: Open and Efficient Foundation Language Models},
    author={Hugo Touvron and Thibaut Lavril and Gautier Izacard and Xavier Martinet and Marie-Anne Lachaux and Timothée Lacroix and Baptiste Rozière and Naman Goyal and Eric Hambro and Faisal Azhar and Aurelien Rodriguez and Armand Joulin and Edouard Grave and Guillaume Lample},
    year={2023},
    eprint={2302.13971},
    archivePrefix={arXiv},
    primaryClass={cs.CL}
}

@misc{almazrouei2023falcon,
      title={The Falcon Series of Open Language Models}, 
      author={Ebtesam Almazrouei and Hamza Alobeidli and Abdulaziz Alshamsi and Alessandro Cappelli and Ruxandra Cojocaru and Mérouane Debbah and Étienne Goffinet and Daniel Hesslow and Julien Launay and Quentin Malartic and Daniele Mazzotta and Badreddine Noune and Baptiste Pannier and Guilherme Penedo},
      year={2023},
      eprint={2311.16867},
      archivePrefix={arXiv},
      primaryClass={cs.CL},
      url={https://arxiv.org/abs/2311.16867}, 
}

@misc{hendrycks2023gaussianerrorlinearunits,
      title={Gaussian Error Linear Units (GELUs)}, 
      author={Dan Hendrycks and Kevin Gimpel},
      year={2023},
      eprint={1606.08415},
      archivePrefix={arXiv},
      primaryClass={cs.LG},
      url={https://arxiv.org/abs/1606.08415}, 
}

@ARTICLE{gribonval2003sparse,
  author={Gribonval, R. and Nielsen, M.},
  journal={IEEE Transactions on Information Theory}, 
  title={Sparse representations in unions of bases}, 
  year={2003},
  volume={49},
  number={12},
  pages={3320-3325},
  doi={10.1109/TIT.2003.820031}}

@InProceedings{press2017using,
  author    = {Press, Ofir  and  Wolf, Lior},
  title     = {Using the Output Embedding to Improve Language Models},
  booktitle = {Proceedings of the 15th Conference of the European Chapter of the Association for Computational Linguistics: Volume 2, Short Papers},
  month     = {April},
  year      = {2017},
  address   = {Valencia, Spain},
  publisher = {Association for Computational Linguistics},
  pages     = {157--163},
}

@inproceedings{charton2024learning,
title={Learning the greatest common divisor: explaining transformer predictions},
author={Francois Charton},
booktitle={The Twelfth International Conference on Learning Representations},
year={2024},
url={https://openreview.net/forum?id=cmcD05NPKa}
}

@inproceedings{li2023thelazy,
title={The Lazy Neuron Phenomenon: On Emergence of Activation Sparsity in Transformers},
author={Zonglin Li and Chong You and Srinadh Bhojanapalli et al.},
booktitle={ICLR},
year={2023},
url={https://openreview.net/forum?id=TJ2nxciYCk-}
}

@misc{nichani2024factualrecall,
      title={Understanding Factual Recall in Transformers via Associative Memories}, 
      author={Eshaan Nichani and Jason D. Lee and Alberto Bietti},
      year={2024},
      eprint={2412.06538},
      archivePrefix={arXiv},
      primaryClass={cs.LG},
      url={https://arxiv.org/abs/2412.06538}, 
}

@inproceedings{odonnat2025circuit,
  author={Odonnat, Ambroise and Bouaziz, Wassim and Cabannes, Vivien},
  booktitle={ICASSP 2025 - 2025 IEEE International Conference on Acoustics, Speech and Signal Processing (ICASSP)}, 
  title={Easing Optimization Paths: a Circuit Perspective}, 
  year={2025},
  volume={},
  number={},
  pages={1-5},
  doi={10.1109/ICASSP49660.2025.10888894}
}

@article{chowdhery2023palm,
  author  = {Aakanksha Chowdhery and Sharan Narang and Jacob Devlin and Maarten Bosma and Gaurav Mishra and Adam Roberts and Paul Barham and Hyung Won Chung and Charles Sutton and Sebastian Gehrmann and Parker Schuh et al.},
  title   = {PaLM: Scaling Language Modeling with Pathways},
  journal = {Journal of Machine Learning Research},
  year    = {2023},
  volume  = {24},
  number  = {240},
  pages   = {1--113},
  url     = {http://jmlr.org/papers/v24/22-1144.html}
}

@misc{grattafiori2024llama3herdmodels,
      title={The Llama 3 Herd of Models}, 
      author={Aaron Grattafiori and Abhimanyu Dubey and Abhinav Jauhri and Abhinav Pandey and Abhishek Kadian and Ahmad Al-Dahle and Aiesha Letman and Akhil Mathur and Alan Schelten and Alex Vaughan et al.},
      year={2024},
      eprint={2407.21783},
      archivePrefix={arXiv},
      primaryClass={cs.AI},
      url={https://arxiv.org/abs/2407.21783}, 
}

@inproceedings{boix_adsera_2023_gradual,
 author = {Boix-Adsera, Enric and Littwin, Etai and Abbe, Emmanuel and Bengio, Samy and Susskind, Joshua},
 booktitle = {Advances in Neural Information Processing Systems},
 editor = {A. Oh and T. Naumann and A. Globerson and K. Saenko and M. Hardt and S. Levine},
 pages = {24519--24551},
 publisher = {Curran Associates, Inc.},
 title = {Transformers learn through gradual rank increase},
 url = {https://proceedings.neurips.cc/paper_files/paper/2023/file/4d69c1c057a8bd570ba4a7b71aae8331-Paper-Conference.pdf},
 volume = {36},
 year = {2023}
}

@inproceedings{geshkovski2023clusters,
title={The emergence of clusters in self-attention dynamics},
author={Borjan Geshkovski and Cyril Letrouit and Yury Polyanskiy and Philippe Rigollet},
booktitle={Thirty-seventh Conference on Neural Information Processing Systems},
year={2023},
url={https://openreview.net/forum?id=aMjaEkkXJx}
}

@article{zhang2024linear,
  author  = {Ruiqi Zhang and Spencer Frei and Peter L. Bartlett},
  title   = {Trained Transformers Learn Linear Models In-Context},
  journal = {Journal of Machine Learning Research},
  year    = {2024},
  volume  = {25},
  number  = {49},
  pages   = {1--55},
  url     = {http://jmlr.org/papers/v25/23-1042.html}
}

@inproceedings{mahankali2024one,
title={One Step of Gradient Descent is Provably the Optimal In-Context Learner with One Layer of Linear Self-Attention},
author={Arvind V. Mahankali and Tatsunori Hashimoto and Tengyu Ma},
booktitle={The Twelfth International Conference on Learning Representations},
year={2024},
url={https://openreview.net/forum?id=8p3fu56lKc}
}

@inproceedings{ahn2023transformers,
title={Transformers learn to implement preconditioned gradient descent for in-context learning},
author={Kwangjun Ahn and Xiang Cheng and Hadi Daneshmand and Suvrit Sra},
booktitle={Thirty-seventh Conference on Neural Information Processing Systems},
year={2023},
url={https://openreview.net/forum?id=LziniAXEI9}
}

@inproceedings{tian2023scan,
title={Scan and Snap: Understanding Training Dynamics and Token Composition in 1-layer Transformer},
author={Yuandong Tian and Yiping Wang and Beidi Chen and Simon Shaolei Du},
booktitle={Thirty-seventh Conference on Neural Information Processing Systems},
year={2023},
url={https://openreview.net/forum?id=l3HUgVHqGQ}
}

@inproceedings{jelassi2022vision,
title={Vision Transformers provably learn spatial structure},
author={Samy Jelassi and Michael Eli Sander and Yuanzhi Li},
booktitle={Advances in Neural Information Processing Systems},
editor={Alice H. Oh and Alekh Agarwal and Danielle Belgrave and Kyunghyun Cho},
year={2022},
url={https://openreview.net/forum?id=eMW9AkXaREI}
}

@inproceedings{chistikov2023learning,
title={Learning a Neuron by a Shallow Re{LU} Network: Dynamics and Implicit Bias for Correlated Inputs},
author={Dmitry Chistikov and Matthias Englert and Ranko Lazic},
booktitle={Thirty-seventh Conference on Neural Information Processing Systems},
year={2023},
url={https://openreview.net/forum?id=xgY4QcOiEZ}
}

@inproceedings{pesme2023saddletosaddle,
title={Saddle-to-Saddle Dynamics in Diagonal Linear Networks},
author={Scott Pesme and Nicolas Flammarion},
booktitle={Thirty-seventh Conference on Neural Information Processing Systems},
year={2023},
url={https://openreview.net/forum?id=iuqCXg1Gng}
}

@article{chi2019nonconvex,
author = {Chi, Yuejie and Lu, Yue M. and Chen, Yuxin},
title = {Nonconvex Optimization Meets Low-Rank Matrix Factorization: An Overview},
year = {2019},
issue_date = {Oct.15, 2019},
publisher = {IEEE Press},
volume = {67},
number = {20},
issn = {1053-587X},
url = {https://doi.org/10.1109/TSP.2019.2937282},
doi = {10.1109/TSP.2019.2937282},
journal = {Trans. Sig. Proc.},
month = oct,
pages = {5239–5269},
numpages = {31}
}

@book{nesterov2014convex,
author = {Nesterov, Yurii},
title = {Introductory Lectures on Convex Optimization: A Basic Course},
year = {2014},
isbn = {1461346916},
publisher = {Springer Publishing Company, Incorporated},
edition = {1},
}

@inproceedings{agarwal2017approximate,
author = {Agarwal, Naman and Allen-Zhu, Zeyuan and Bullins, Brian and Hazan, Elad and Ma, Tengyu},
title = {Finding approximate local minima faster than gradient descent},
year = {2017},
isbn = {9781450345286},
publisher = {Association for Computing Machinery},
address = {New York, NY, USA},
url = {https://doi.org/10.1145/3055399.3055464},
doi = {10.1145/3055399.3055464},
booktitle = {Proceedings of the 49th Annual ACM SIGACT Symposium on Theory of Computing},
pages = {1195–1199},
numpages = {5},
location = {Montreal, Canada},
series = {STOC 2017}
}

@InProceedings{odonnat2024diversity,
  title = 	 {Leveraging Ensemble Diversity for Robust Self-Training in the Presence of Sample Selection Bias},
  author =       {Odonnat, Ambroise and Feofanov, Vasilii and Redko, Ievgen},
  booktitle = 	 {Proceedings of The 27th International Conference on Artificial Intelligence and Statistics},
  pages = 	 {595--603},
  year = 	 {2024},
  editor = 	 {Dasgupta, Sanjoy and Mandt, Stephan and Li, Yingzhen},
  volume = 	 {238},
  series = 	 {Proceedings of Machine Learning Research},
  month = 	 {02--04 May},
  publisher =    {PMLR},
  url = 	 {https://proceedings.mlr.press/v238/odonnat24a.html},
}

@inproceedings{cascante2021curriculum,
  title={Curriculum labeling: Revisiting pseudo-labeling for semi-supervised learning},
  author={Cascante-Bonilla, Paola and Tan, Fuwen and Qi, Yanjun and Ordonez, Vicente},
  booktitle={Proceedings of the AAAI conference on artificial intelligence},
  volume={35},
  pages={6912--6920},
  year={2021}
}

@misc{comanici2025gemini25pushingfrontier,
      title={Gemini 2.5: Pushing the Frontier with Advanced Reasoning, Multimodality, Long Context, and Next Generation Agentic Capabilities}, 
      author={Gheorghe Comanici and Eric Bieber and Mike Schaekermann and Ice Pasupat and Noveen Sachdeva and Inderjit Dhillon and Marcel Blistein and Ori Ram et al.},
      year={2025},
      eprint={2507.06261},
      archivePrefix={arXiv},
      primaryClass={cs.CL},
      url={https://arxiv.org/abs/2507.06261}, 
}

@misc{brown2020gpt3,
    title={Language Models are Few-Shot Learners},
    author={Tom B. Brown and Benjamin Mann and Nick Ryder and Melanie Subbiah and Jared Kaplan and Prafulla Dhariwal and Arvind Neelakantan and Pranav Shyam and Girish Sastry and Amanda Askell and Sandhini Agarwal and Ariel Herbert-Voss and Gretchen Krueger and Tom Henighan and Rewon Child and Aditya Ramesh and Daniel M. Ziegler and Jeffrey Wu and Clemens Winter and Christopher Hesse and Mark Chen and Eric Sigler and Mateusz Litwin and Scott Gray and Benjamin Chess and Jack Clark and Christopher Berner and Sam McCandlish and Alec Radford and Ilya Sutskever and Dario Amodei},
    year={2020},
    eprint={2005.14165},
    archivePrefix={arXiv},
    primaryClass={cs.CL}
}

@misc{team2025kimi,
    title={Kimi K2: Open Agentic Intelligence},
    author={{Kimi Team} and Yifan Bai and Yiping Bao and Guanduo Chen and Jiahao Chen and Ningxin Chen and Ruijue Chen and Yanru Chen and Yuankun Chen and Yutian Chen and Zhuofu Chen and Jialei Cui and Hao Ding and Mengnan Dong and Angang Du and Chenzhuang Du and Dikang Du and Yulun Du and Yu Fan and Yichen Feng and Kelin Fu and Bofei Gao et al.},
    year={2025},
    eprint={2507.20534},
    archivePrefix={arXiv},
    primaryClass={cs.LG}
}

@misc{li2025trainingdynamics17bllama,
      title={Training Dynamics of a 1.7B LLaMa Model: A Data-Efficient Approach}, 
      author={Miles Q. Li and Benjamin C. M. Fung and Shih-Chia Huang},
      year={2025},
      eprint={2412.13335},
      archivePrefix={arXiv},
      primaryClass={cs.CL},
      url={https://arxiv.org/abs/2412.13335}, 
}

@misc{marin,
  title = {Marin: developing foundation models},
  author={Marin},
  year={2025},
  howpublished = {\url{https://marin.community/}},
  note = {Accessed: 2025-09-30}
}

@inproceedings{robinson2023leveraging,
title={Leveraging Large Language Models for Multiple Choice Question Answering},
author={Joshua Robinson and David Wingate},
booktitle={The Eleventh International Conference on Learning Representations },
year={2023},
url={https://openreview.net/forum?id=yKbprarjc5B}
}

@inproceedings{singh2025exposing,
    title = "Exposing the Achilles' Heel: Evaluating {LLM}s Ability to Handle Mistakes in Mathematical Reasoning",
    author = "Singh, Joykirat  and
      Nambi, Akshay  and
      Vineet, Vibhav",
    editor = "Che, Wanxiang  and
      Nabende, Joyce  and
      Shutova, Ekaterina  and
      Pilehvar, Mohammad Taher",
    booktitle = "Proceedings of the 63rd Annual Meeting of the Association for Computational Linguistics (Volume 1: Long Papers)",
    month = jul,
    year = "2025",
    address = "Vienna, Austria",
    publisher = "Association for Computational Linguistics",
    url = "https://aclanthology.org/2025.acl-long.1313/",
    doi = "10.18653/v1/2025.acl-long.1313",
    pages = "27044--27065",
    ISBN = "979-8-89176-251-0",
}

@inproceedings{loshchilov2025ngpt,
title={n{GPT}: Normalized Transformer with Representation Learning on the Hypersphere},
author={Ilya Loshchilov and Cheng-Ping Hsieh and Simeng Sun and Boris Ginsburg},
booktitle={The Thirteenth International Conference on Learning Representations},
year={2025},
url={https://openreview.net/forum?id=se4vjm7h4E}
}

@inproceedings{saxton2018analysing,
title={Analysing Mathematical Reasoning Abilities of Neural Models},
author={David Saxton and Edward Grefenstette and Felix Hill and Pushmeet Kohli},
booktitle={International Conference on Learning Representations},
year={2019},
url={https://openreview.net/forum?id=H1gR5iR5FX},
}

@inproceedings{dziri2023faith,
title={Faith and Fate: Limits of Transformers on Compositionality},
author={Nouha Dziri and Ximing Lu and Melanie Sclar and Xiang Lorraine Li and Liwei Jiang and Bill Yuchen Lin and Sean Welleck and Peter West and Chandra Bhagavatula and Ronan Le Bras and Jena D. Hwang and Soumya Sanyal and Xiang Ren and Allyson Ettinger and Zaid Harchaoui and Yejin Choi},
booktitle={Thirty-seventh Conference on Neural Information Processing Systems},
year={2023},
url={https://openreview.net/forum?id=Fkckkr3ya8}
}

@inproceedings{lee2024teaching,
title={Teaching Arithmetic to Small Transformers},
author={Nayoung Lee and Kartik Sreenivasan and Jason D. Lee and Kangwook Lee and Dimitris Papailiopoulos},
booktitle={The Twelfth International Conference on Learning Representations},
year={2024},
url={https://openreview.net/forum?id=dsUB4bst9S}
}

@misc{raffel2023exploringlimitstransferlearning,
      title={Exploring the Limits of Transfer Learning with a Unified Text-to-Text Transformer}, 
      author={Colin Raffel and Noam Shazeer and Adam Roberts and Katherine Lee and Sharan Narang and Michael Matena and Yanqi Zhou and Wei Li and Peter J. Liu},
      year={2023},
      eprint={1910.10683},
      archivePrefix={arXiv},
      primaryClass={cs.LG},
      url={https://arxiv.org/abs/1910.10683}, 
}

@article{ruiqi2024linear,
author = {Zhang, Ruiqi and Frei, Spencer and Bartlett, Peter L.},
title = {Trained transformers learn linear models in-context},
year = {2024},
issue_date = {January 2024},
publisher = {JMLR.org},
volume = {25},
number = {1},
issn = {1532-4435},
journal = {J. Mach. Learn. Res.},
month = jan,
articleno = {49},
numpages = {55}
}

@misc{xiong2020prenorm,
title={On Layer Normalization in the Transformer Architecture},
author={Ruibin Xiong and Yunchang Yang and Di He and Kai Zheng and Shuxin Zheng and Huishuai Zhang and Yanyan Lan and Liwei Wang and Tie-Yan Liu},
year={2020},
url={https://openreview.net/forum?id=B1x8anVFPr}
}

@inproceedings{wolf2017tying,
    title = "Using the Output Embedding to Improve Language Models",
    author = "Press, Ofir  and
      Wolf, Lior",
    editor = "Lapata, Mirella  and
      Blunsom, Phil  and
      Koller, Alexander",
    booktitle = "Proceedings of the 15th Conference of the {E}uropean Chapter of the Association for Computational Linguistics: Volume 2, Short Papers",
    month = apr,
    year = "2017",
    address = "Valencia, Spain",
    publisher = "Association for Computational Linguistics",
    url = "https://aclanthology.org/E17-2025/",
    pages = "157--163",
}

@misc{ba2016layer,
    title={Layer Normalization},
    author={Jimmy Lei Ba and Jamie Ryan Kiros and Geoffrey E. Hinton},
    year={2016},
    eprint={1607.06450},
    archivePrefix={arXiv},
    primaryClass={stat.ML}
}

@article{ameisen2025circuit,
  author={Ameisen, Emmanuel and Lindsey, Jack and Pearce, Adam and Gurnee, Wes and Turner, Nicholas L. and Chen, Brian and Citro, Craig and Abrahams, David and Carter, Shan and Hosmer, Basil and Marcus, Jonathan and Sklar, Michael and Templeton, Adly and Bricken, Trenton and McDougall, Callum and Cunningham, Hoagy and Henighan, Thomas and Jermyn, Adam and Jones, Andy and Persic, Andrew and Qi, Zhenyi and Ben Thompson, T. and Zimmerman, Sam and Rivoire, Kelley and Conerly, Thomas and Olah, Chris and Batson, Joshua},
  title={Circuit Tracing: Revealing Computational Graphs in Language Models},
  journal={Transformer Circuits Thread},
  year={2025},
  url={https://transformer-circuits.pub/2025/attribution-graphs/methods.html}
}

@article{lindsey2025biology,
  author={Lindsey, Jack and Gurnee, Wes and Ameisen, Emmanuel and Chen, Brian and Pearce, Adam and Turner, Nicholas L. and Citro, Craig and Abrahams, David and Carter, Shan and Hosmer, Basil and Marcus, Jonathan and Sklar, Michael and Templeton, Adly and Bricken, Trenton and McDougall, Callum and Cunningham, Hoagy and Henighan, Thomas and Jermyn, Adam and Jones, Andy and Persic, Andrew and Qi, Zhenyi and Thompson, T. Ben and Zimmerman, Sam and Rivoire, Kelley and Conerly, Thomas and Olah, Chris and Batson, Joshua},
  title={On the Biology of a Large Language Model},
  journal={Transformer Circuits Thread},
  year={2025},
  url={https://transformer-circuits.pub/2025/attribution-graphs/biology.html}
}

@inproceedings{lu2019understanding,
title={Understanding and Improving Transformer From a Multi-Particle Dynamic System Point of View.},
author={Yiping Lu and Zhuohan Li and Di He and Zhiqing Sun and Bin Dong and Tao Qin and Liwei Wang and Tie-yan Liu},
booktitle={ICLR 2020 Workshop on Integration of Deep Neural Models and Differential Equations},
year={2019},
url={https://openreview.net/forum?id=pxlqJa21C}
}

@misc{geshkovski2023perspective,
      title={A mathematical perspective on Transformers}, 
      author={Borjan Geshkovski and Cyril Letrouit and Yury Polyanskiy and Philippe Rigollet},
      year={2023},
      eprint={2312.10794},
      archivePrefix={arXiv},
      primaryClass={cs.LG}
}

@misc{geshkovski2025measure,
      title={Measure-to-measure interpolation using Transformers}, 
      author={Borjan Geshkovski and Philippe Rigollet and Domènec Ruiz-Balet},
      year={2025},
      eprint={2411.04551},
      archivePrefix={arXiv},
      primaryClass={math.OC},
      url={https://arxiv.org/abs/2411.04551}, 
}

@inproceedings{reddy2024abrupt,
title={The mechanistic basis of data dependence and abrupt learning in an in-context classification task},
author={Gautam Reddy},
booktitle={The Twelfth International Conference on Learning Representations},
year={2024},
url={https://openreview.net/forum?id=aN4Jf6Cx69}
}

@inproceedings{gomes2025adafisher,
title={AdaFisher: Adaptive Second Order Optimization via Fisher Information},
author={Damien Martins Gomes and Yanlei Zhang and Eugene Belilovsky and Guy Wolf and Mahdi S. Hosseini},
booktitle={The Thirteenth International Conference on Learning Representations},
year={2025},
url={https://openreview.net/forum?id=puTxuiK2qO}
}

@inproceedings{hoffmann2024eurekamoments,
title={Eureka-Moments in Transformers: Multi-Step Tasks Reveal Softmax Induced Optimization Problems},
author={David T Hoffmann and Simon Schrodi and Jelena Bratuli{\'c} and Nadine Behrmann and Volker Fischer and Thomas Brox},
booktitle={Forty-first International Conference on Machine Learning},
year={2024},
url={https://openreview.net/forum?id=HssOwuZiaB}
}

@inproceedings{minegishi2025beyond,
title={Beyond Induction Heads: In-Context Meta Learning Induces Multi-Phase Circuit Emergence},
author={Gouki Minegishi and Hiroki Furuta and Shohei Taniguchi and Yusuke Iwasawa and Yutaka Matsuo},
booktitle={Forty-second International Conference on Machine Learning},
year={2025},
url={https://openreview.net/forum?id=Xw01vF13aV}
}

@inproceedings{varre2025learning,
title={Learning In-context \$n\$-grams with Transformers: Sub-\$n\$-grams Are Near-Stationary Points},
author={Aditya Varre and Gizem Y{\"u}ce and Nicolas Flammarion},
booktitle={Forty-second International Conference on Machine Learning},
year={2025},
url={https://openreview.net/forum?id=OMwdvGDeHL}
}

@misc{jordan2024muon,
  author       = {Keller Jordan and Yuchen Jin and Vlado Boza and Jiacheng You and
                  Franz Cesista and Laker Newhouse and Jeremy Bernstein},
  title        = {Muon: An optimizer for hidden layers in neural networks},
  year         = {2024},
  url          = {https://kellerjordan.github.io/posts/muon/}
}

@InProceedings{gupta2018shampoo,
  title = 	 {Shampoo: Preconditioned Stochastic Tensor Optimization},
  author =       {Gupta, Vineet and Koren, Tomer and Singer, Yoram},
  booktitle = 	 {Proceedings of the 35th International Conference on Machine Learning},
  pages = 	 {1842--1850},
  year = 	 {2018},
  editor = 	 {Dy, Jennifer and Krause, Andreas},
  volume = 	 {80},
  series = 	 {Proceedings of Machine Learning Research},
  month = 	 {10--15 Jul},
  publisher =    {PMLR},
  url = 	 {https://proceedings.mlr.press/v80/gupta18a.html},
}

@InProceedings{castin2024smooth,
  title = 	 {How Smooth Is Attention?},
  author =       {Castin, Val\'{e}rie and Ablin, Pierre and Peyr\'{e}, Gabriel},
  booktitle = 	 {Proceedings of the 41st International Conference on Machine Learning},
  pages = 	 {5817--5840},
  year = 	 {2024},
  editor = 	 {Salakhutdinov, Ruslan and Kolter, Zico and Heller, Katherine and Weller, Adrian and Oliver, Nuria and Scarlett, Jonathan and Berkenkamp, Felix},
  volume = 	 {235},
  series = 	 {Proceedings of Machine Learning Research},
  month = 	 {21--27 Jul},
  publisher =    {PMLR},
  url = 	 {https://proceedings.mlr.press/v235/castin24a.html},
}

@inproceedings{caron2021emerging,
  title={Emerging Properties in Self-Supervised Vision Transformers},
  author={Caron, Mathilde and Touvron, Hugo and Misra, Ishan and J\'egou, Herv\'e  and Mairal, Julien and Bojanowski, Piotr and Joulin, Armand},
  booktitle={Proceedings of the International Conference on Computer Vision (ICCV)},
  year={2021}
}

@inproceedings{dosovitskiy2021vit,
title={An Image is Worth 16x16 Words: Transformers for Image Recognition at Scale},
author={Alexey Dosovitskiy and Lucas Beyer and Alexander Kolesnikov and Dirk Weissenborn and Xiaohua Zhai and Thomas Unterthiner and Mostafa Dehghani and Matthias Minderer and Georg Heigold and Sylvain Gelly and Jakob Uszkoreit and Neil Houlsby},
booktitle={International Conference on Learning Representations},
year={2021},
url={https://openreview.net/forum?id=YicbFdNTTy}
}

@inproceedings{nie2023patchtst,
title={A Time Series is Worth 64 Words:  Long-term Forecasting with Transformers},
author={Yuqi Nie and Nam H Nguyen and Phanwadee Sinthong and Jayant Kalagnanam},
booktitle={The Eleventh International Conference on Learning Representations },
year={2023},
url={https://openreview.net/forum?id=Jbdc0vTOcol}
}

@misc{deepseekai2025deepseekr1,
      title={DeepSeek-R1: Incentivizing Reasoning Capability in LLMs via Reinforcement Learning}, 
      author={Daya Guo and Dejian Yang and Haowei Zhang and Junxiao Song and Ruoyu Zhang and Runxin Xu and Qihao Zhu and Shirong Ma and Peiyi Wang and Xiao Bi and Xiaokang Zhang and Xingkai Yu and Yu Wu and Z. F. Wu and Zhibin Gou et al.},
      year={2025},
      eprint={2501.12948},
      archivePrefix={arXiv},
      primaryClass={cs.CL},
      url={https://arxiv.org/abs/2501.12948}, 
}

@inproceedings{
conmy2023towards,
title={Towards Automated Circuit Discovery for Mechanistic Interpretability},
author={Arthur Conmy and Augustine N. Mavor-Parker and Aengus Lynch and Stefan Heimersheim and Adri{\`a} Garriga-Alonso},
booktitle={Thirty-seventh Conference on Neural Information Processing Systems},
year={2023},
url={https://openreview.net/forum?id=89ia77nZ8u}
}

@inproceedings{
takase2025spike,
title={Spike No More: Stabilizing the Pre-training of Large Language Models},
author={Sho Takase and Shun Kiyono and Sosuke Kobayashi and Jun Suzuki},
booktitle={Second Conference on Language Modeling},
year={2025},
url={https://openreview.net/forum?id=52YBEzcI0l}
}
\bibliographystyle{template/icml2026}

\clearpage
\appendix
\onecolumn
\onecolumn
{\Large Appendix}

\paragraph{Roadmap.} We conduct additional experiments in \cref{app:add_exp}. In \cref{app:theory}, we discuss in detail the theoretical insights on the learning dynamics and provide the proofs in \cref{app:proofs}. 

\addtocontents{toc}{\protect\setcounter{tocdepth}{2}}

\renewcommand*\contentsname{\Large Table of Contents}

\tableofcontents
\clearpage

\section{Experimental results}
\label{app:add_exp}
In this section, we present additional results related to the study of the circuits implemented by the transformer. Notably, we investigate the concrete realizations to solve the task, the sparsity of the model's activation along training, and the transferability of circuits. 

\paragraph{Implementation details.} In our experiments, we trained our networks using \( 2^{11} \) data points, which were sampled uniformly with replacement from the \( p^N \) possible sentences.
We utilized the Adam optimizer with parameters \( \beta_1 = 0.9 \) and \( \beta_2 = 0.999 \) \citep{kingma2017adam}, and we initialized the network weights using the default schemes provided by PyTorch \citep{paszke2019pytorch}. Experiments were conducted on high-performance GPUs such as V100 and can be reproduced on a single device. The code will be open-sourced upon publication.

\subsection{Connectivity of neurons}
\label{app:sparsity}
Sparsity is a phenomenon of interest in many fields such as signal processing, neuroscience, and machine learning~\citep{mairal2009sparse, chen1998atomic, barth2012neocortex}. Recent studies focused on the sparsity in deep neural network activations. In particular, \citet{li2023lazy} showed that trained transformers have sparse activations and concluded that it was caused by the training dynamics rather than by a compact representation of the training data, as commonly thought in computer vision and NLP. \citet{mirzadeh2024relu} observed a similar phenomenon and showed how to leverage sparsity to reduce the inference cost of large language models. Inspired by this line of work, we study the activation sparsity of our model from a performance viewpoint. It should be noted that those works study deep transformers and identified that the sparsity increases with the depth, while we only consider a one-layer transformer. In our setting, the activation sparsity can help us better understand how neurons connect to implement \emph{clustering heads}.

\paragraph{Successful models have dense activations.} Following the framework from~\citet{li2023lazy}, we recall that the activation sparsity corresponds to the percentage of non-zero entries of the feed-forward activation map. Without loss of generality, the feed-forward block is an MLP with weights $W_1, W_2$ and a non-linear activation $\sigma$ that outputs for any input $x$ a vector $z = W_1 \sigma(W_2 x)$. Formally, the activation sparsity is the percentage of non-zero entries in $\sigma(W_2 x)$ and takes values in $[0,1]$. In the classical setting with ReLU activation~\citep{fukushima1969relu}, this is equivalent to computing the percentage of non-negative neurons before the activation. However, some activations do not have non-negative outputs. This is the case of the SiLU~\citep{elwing2018silu}, used in Llama models~\citep{touvron2023llama}, and of the GeLU~\citet{hendrycks2023gaussianerrorlinearunits} used in Falcon~\citep{almazrouei2023falcon}, PaLM~\citep{chowdhery2024palm}, and in our transformer implementation. Instead of replacing such activations with a ReLU~\citep{mirzadeh2024relu, li2023lazy}, we compute a smoothed sparsity. This is akin to using the $\ell_1$-norm, respectively the nuclear norm, instead of the $\ell_0$ quasi-norm, respectively the rank~\citep{ilbert2024samformer, gribonval2003sparse}. as the percentage of entries with an absolute value lower than $\varepsilon > 0$. A sparsity of $1$ means that all entries are $\varepsilon$-close to $0$ (i.e., sparse activations), and a sparsity of $0$ means that all entries are at least $\varepsilon$-away from $0$ (i.e., dense activations). 

\paragraph{Evolution of the sparsity with performance.} To better understand the impact of sparsity, we train $20$ independent models and display in \cref{fig:sparsity} their sparsity after training for $\varepsilon \in [10^{-5}, 10^2]$ (the range is chosen such that the sparsity reaches its extremal values $0$ and $1$). Given the task's difficulty, achieving an accuracy above $0.9$ is a success; otherwise, it is a failure. On the left, we plot successful models in \textcolor{blue}{blue} and failed ones in \textcolor{red}{red}.  We observe a striking separation between successful and failed training. In the permissible range $[10^{-3}, 10]$, successful models tend to have less sparse activation than failed ones. To further study this phenomenon, we plot in the right subplot of \cref{fig:sparsity} the evolution of the sparsity with $\varepsilon$, and here, the color indicates the models' test accuracy (the lighter the color, the better the model). We can see that the sparsity decreases as the performance increases. This explains the sharp transition between failure and success in the left subplot. These experiments seem to indicate that, contrary to images and textual data~\citep{li2023lazy, mirzadeh2024relu}, the sparse modular addition problem needs the involvement of many neurons during inference, and hence requires non-sparse activations.

\subsection{Transferability of circuits}
\label{app:transfer}
Among the lessons learned from training very large models is the importance of careful data engineering \citep{touvron2023llamaopenefficientfoundation,dubey2024llama3herdmodels}.
\citet{cabannes2024iteration} has highlighted that certain sources of data may facilitate the learning of invariances, while \citet{abbe2023provable} discusses how data curation enables models to escape saddle points more quickly.
These insights are consistent with the observations we made regarding our problem.
In the sparse modular addition problem, the number of unique sequences is equal to $p^N$, which increases rapidly with both $p$ and $N$ and makes the problem quite hard to learn.
In particular, when setting $N = 12$ and limiting the training set to $2048$ data points, training for 1000 epochs does not result in any learning for $p>4$.

\begin{figure}[!t]
    \centering
    \begin{subfigure}[t]{0.45\textwidth}
        \centering
        \includegraphics[width=\linewidth]{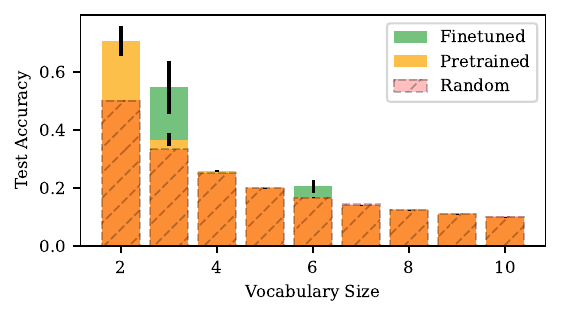}
        \vspace{-1.5em}
        \caption{Mean}
    \end{subfigure}
    \begin{subfigure}[t]{0.45\textwidth}
        \centering
        \includegraphics[width=\linewidth]{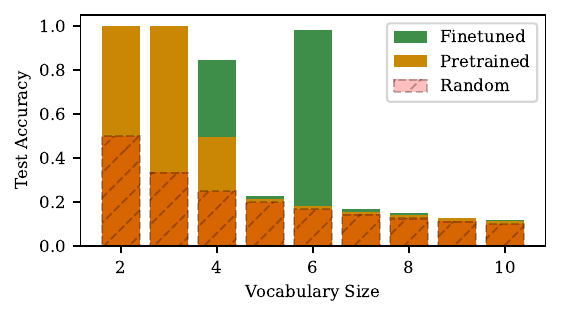}
        \vspace{-1.5em}
        \caption{Maximum}
    \end{subfigure}
    \vspace{-.75em}
    \caption{Accuracies obtained from pretraining only with $p \in [2, 10]$ and finetuning with $p \in [3, 10]$ starting from $n=2$ for various vocabulary sizes. \textbf{Left:} Averaged accuracy; \textbf{Right:} Maximum accuracy. Finetuned models display better performance than pretrained-only models.}
    \label{fig:curse-size}
\vspace{-1em}
\end{figure}

\paragraph{Impact on the performance.} As previously seen in \cref{fig:circuits_corrected}, when training with $p=2$, we often find circuits that capture both permutation and suffix invariants.
These invariants generalize for any $p\in\N$ when $k$ and $N$ are fixed.
Consequently, initializing models with these invariants makes learning the sparse modular addition problem much easier.
This observation was made after conducting the following experiments: we first trained a model with sequences in $\bF_p$ for $p=2$ over 1000 epochs, before switching the dataset to sequences in $\bF_p$ for $p=3$ for another 1000 epochs.
We found that this procedure significantly facilitates the learning of the sparse modular addition problem for $p=3$, which we summarize in \cref{fig:curse-size} where each bar plot is obtained by over $250$ runs.
Remarkably, we found that the only models that achieved 100\% test accuracy were those that captured both the token and permutation invariances after the first 1000 epochs. 
Specifically, these were the models that created six sequence embedding clusters, as shown in Figures \ref{fig:circuit_ideal} and \ref{fig:circuits_corrected}, rather than those depicted in Figures \ref{fig:circuits_invariant} or \ref{fig:circuits_fuzzy}.

\begin{figure}[b!]
    \centering
    \begin{subfigure}[t]{.5\textwidth}
    \centering
    \includegraphics[width=\linewidth]{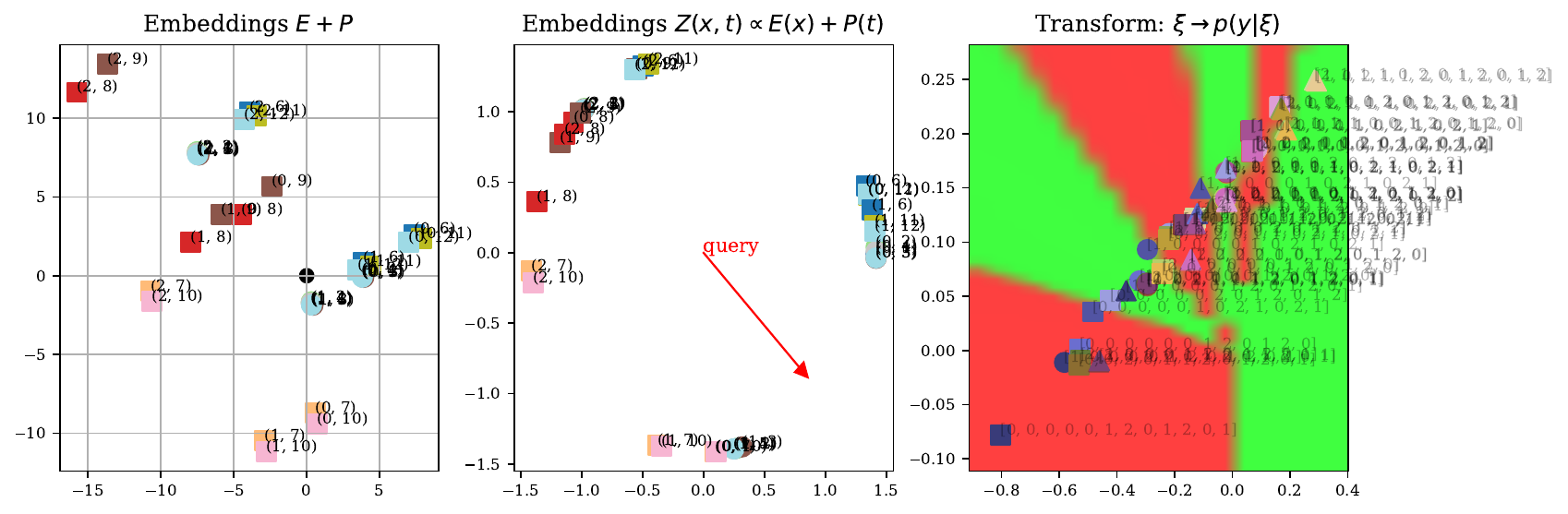}
    \caption{Initialization with an additional token embedding for $p=3$.}
    \label{fig:circuit-finetune-start}
    \end{subfigure}~%
    \begin{subfigure}[t]{.5\textwidth}
    \centering
    \includegraphics[width=\linewidth]{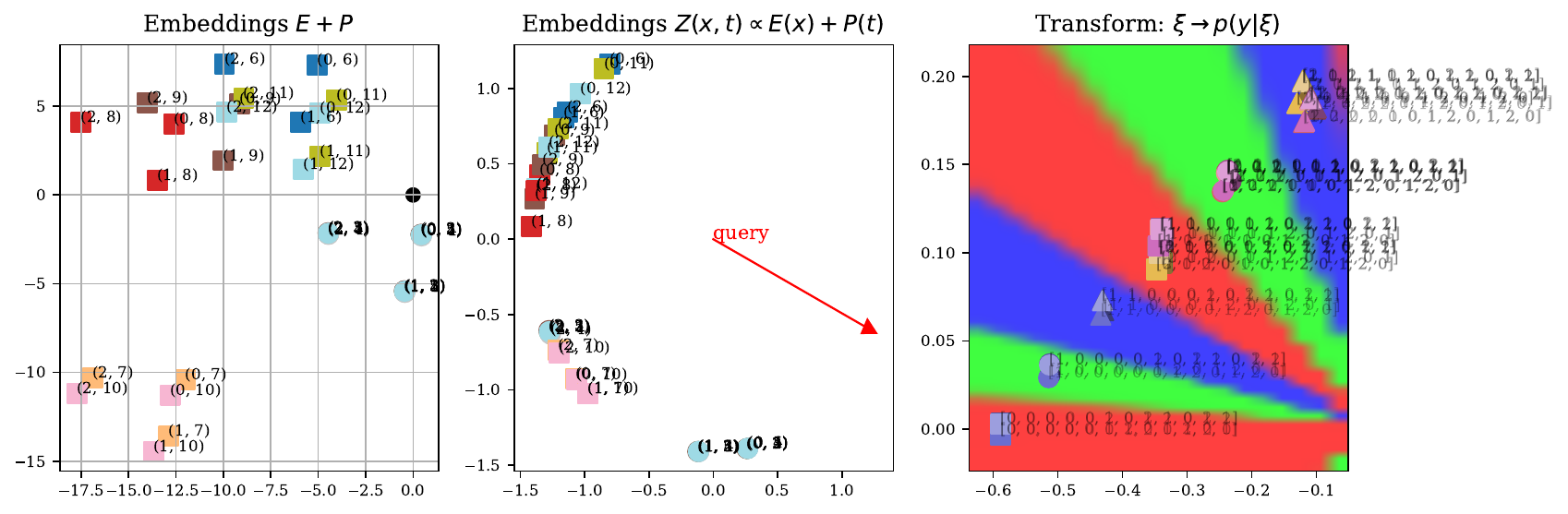}
    \caption{Circuit learned after $1000$ epochs of finetuning with $p=3$.}
    \label{fig:circuit-finetune}
    \end{subfigure}
    \caption{\emph{Transfer of clustering head}. The obtained circuit is akin to \cref{fig:circuits_corrected} where the value corrects the faulty attention.}
\end{figure}

\paragraph{Impact on the learned circuits.} \cref{fig:circuit-finetune} shows the final circuit found in one of our finetuning experiments.
The training was initialized with the circuit in \cref{fig:circuits_corrected}, after adding a token embedding to encode for $x=2$, resulting in \cref{fig:circuit-finetune-start}.
The final embeddings are not that far from the initial ones, with the transformer having learned to mainly pay attention to the non-spurious tokens that are not equal to 2.
It also pays some attention to 0 and 1 in positions $t=7$ and $t=10$.
However, this faulty attention is corrected by the value matrix.
Once again, the final configuration seems somewhat close to the initial one, as shown by the attention pattern reported in \cref{fig:initialization}.
This is consistent with the observations made in the previous subsections.

\paragraph{Benefits of transfer.} Overall, our transfer experiment highlights the transferability of circuits and the usefulness of a curriculum in facilitating the learning of challenging tasks by inducing effective circuits through tasks that are easier to solve.
Although our experiments are performed in two stages, we hypothesize that the same type of mechanism can explain the importance of data curation. We also note that curriculum learning has shown its benefits in other settings, such as semi-supervised learning and self-training~\citep{odonnat2024diversity, cascante2021curriculum}.

\subsection{Impact of the hyperparameters}
\label{app:hyper_params}
\begin{figure}[!h]
    \centering
    \begin{subfigure}[t]{0.49\textwidth}
        \centering
    \includegraphics[width=.9\linewidth]{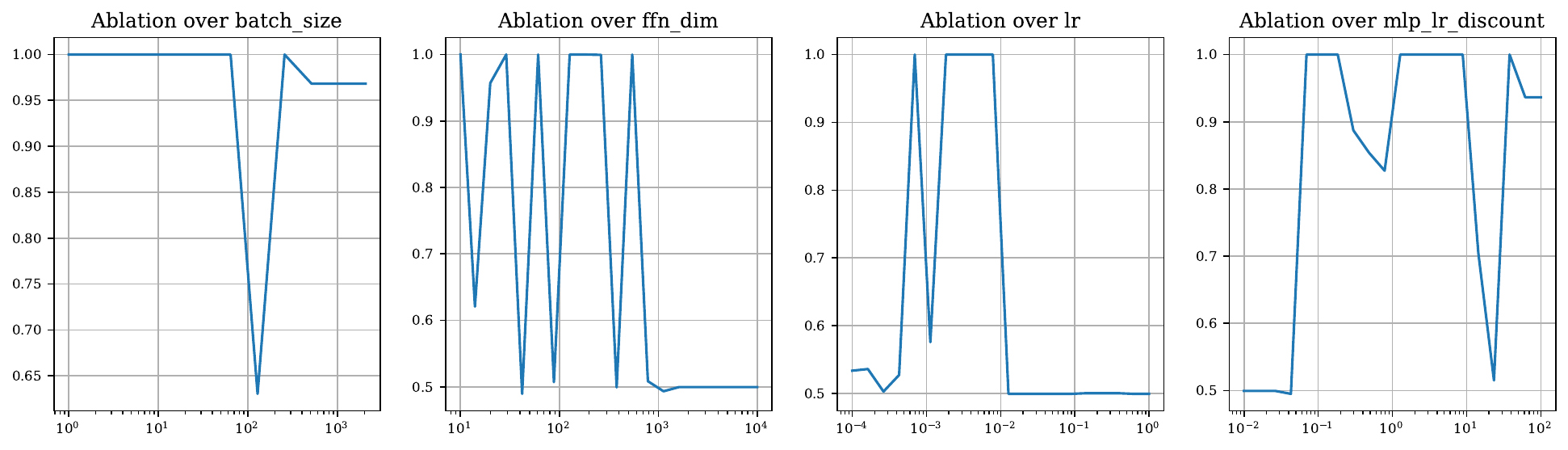}
    \caption{Single run with 1000 epochs.}
    \label{fig:ablation-seed}
    \end{subfigure}%
    ~ 
    \begin{subfigure}[t]{0.49\textwidth}
        \centering
    \includegraphics[width=.9\linewidth]{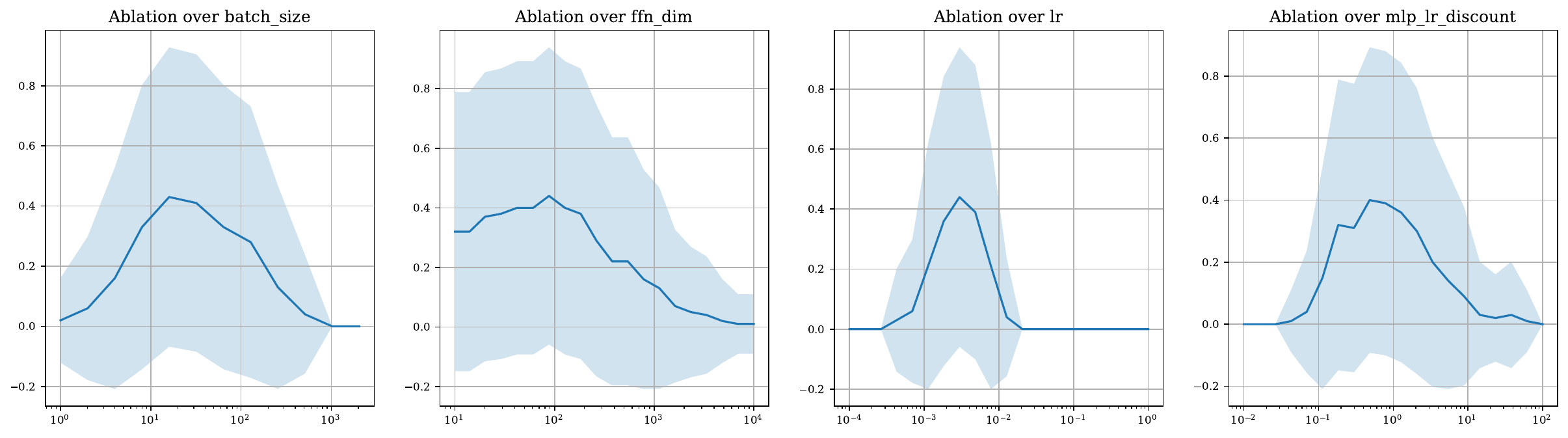}
    \caption{
    Averaged over $100$ runs with standard deviations.}
    \label{fig:ablation}
    \end{subfigure}
    \caption{{\em Ablation studies} regarding test accuracy as a function of batch size, hidden neurons, learning rates, and MLP learning rates discount factor for a single run with 1000 epochs.
    We ensured consistency in initial weights and batch designs when changing hyperparameters. We observe (mostly in \cref{fig:ablation}) some regularity in the effect of hyperparameters on the resulting test accuracy.}
\end{figure}

In this section, we conduct ablation studies on the training hyperparameters. In \cref{fig:ablation-seed}, we observe the high variability of the test accuracy for a fixed seed after $1000$ epochs when varying the batch size, hidden dimension, learning rates, and MLP learning rates. \cref{fig:ablation} is similar with more regularity as it is averaged over $100$ runs with different seeds. This study shows that the exit of saddle points is highly dependent on the hyperparameters.

\subsection{Towards more practical setting}
\label{app:emb_dim}
Relying on our visual sandbox, we studied several phenomena on the training dynamics of neural networks from the learning of the different parts of the network in \cref{sec:training_dyn} to the loss spikes occurring during the optimization in \cref{sec:loss_spikes} through the efficiency of transfer learning in \cref{app:transfer}. We also analyzed the connection between saddle points, loss spikes, and gradient norms in Sections~\ref{sec:training_dyn} and~\ref{sec:loss_spikes} and the impact of the activation sparsity on the models' performance in \cref{app:sparsity}. The low-dimensional embeddings allowed us to pair loss profiles with the visualization of each layer to decipher the training behaviors. However, we note that the studied behaviors and the insights we obtain can be analyzed independently of the embedding dimension $d$. 
\begin{figure}[!h]
    \centering
    \includegraphics[width=\linewidth]{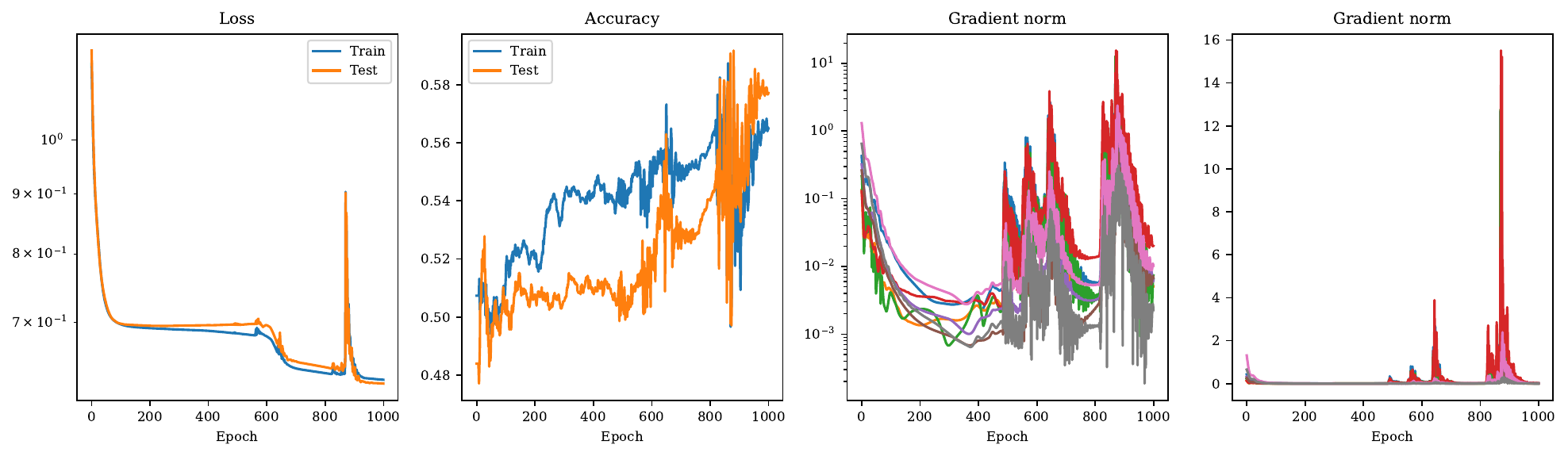}
    \includegraphics[width=\linewidth]{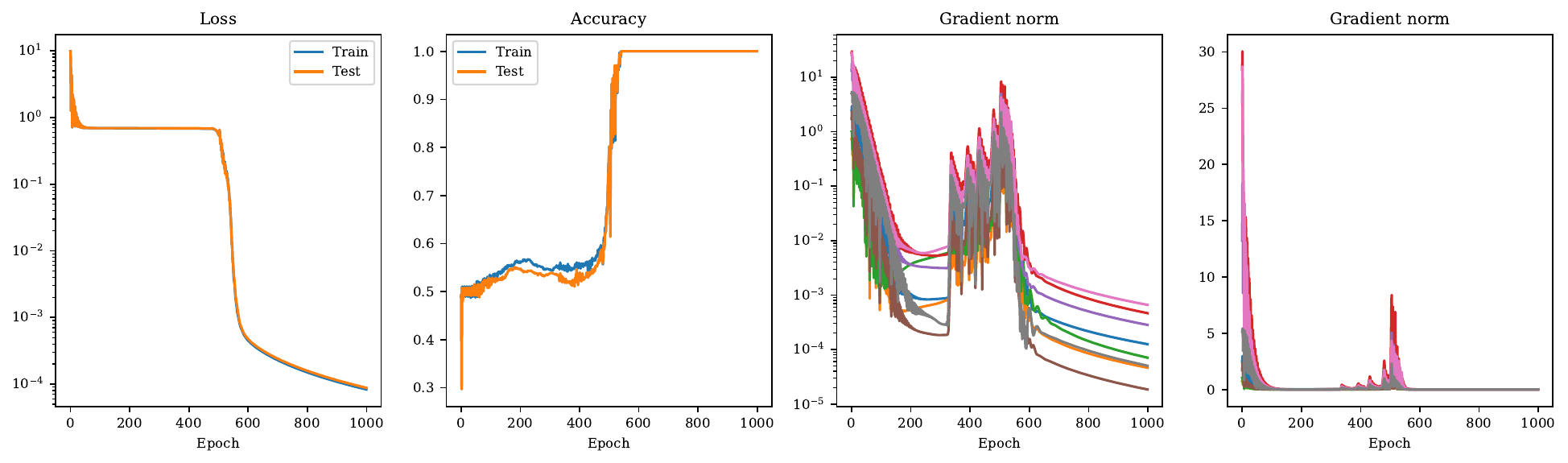}
    \caption{\emph{Connection to saddle points}. This study is akin to \cref{fig:grad-norms-full-batch} with $d=3$ in the \textbf{top} plot and $d=64$ in the \textbf{bottom} one. \textbf{From left to right:} Evolution of train and test losses, the corresponding accuracies, the evolution of gradient norms for each layer in log-scale, and the similar evolution in linear scale in full-batch. Akin to \cref{fig:grad-norms-full-batch}, we see that the learning phases studied in \cref{sec:training_dyn}  appear in tandem with high gradient norms. This can be seen in the last subfigure, where the peaks correspond to the loss drops and spikes.}
    \label{fig:grad-norms-full-batch-d-3-64}
\end{figure}
\paragraph{Motivation.} Taking an embedding size $d=2$ was mainly motivated by visualization purposes, and we saw that such a setting was already challenging enough to observe training phenomena of interest in more practical settings. For the sake of self-consistency, we extend our experiments with higher-dimensional embeddings and show that our insights remain valid in more practical settings with $d>2$. 

\paragraph{Two-phase learning and saddle points}
\label{app:loss_prof_saddle}
In this section, we extend the experiments of \cref{sec:training_dyn} for $d>2$. In \cref{fig:grad-norms-full-batch-d-3-64}, we consider $d=3$ (top) and $d=64$ (bottom). We obtain similar conclusions to those in the case $d=2$ in \cref{fig:grad-norms-full-batch} where the loss drops occur in tandem with high gradient norms for each layer, hinting at the exit of saddle points. 

\begin{figure}[!h]
    \centering
    \includegraphics[width=\linewidth]{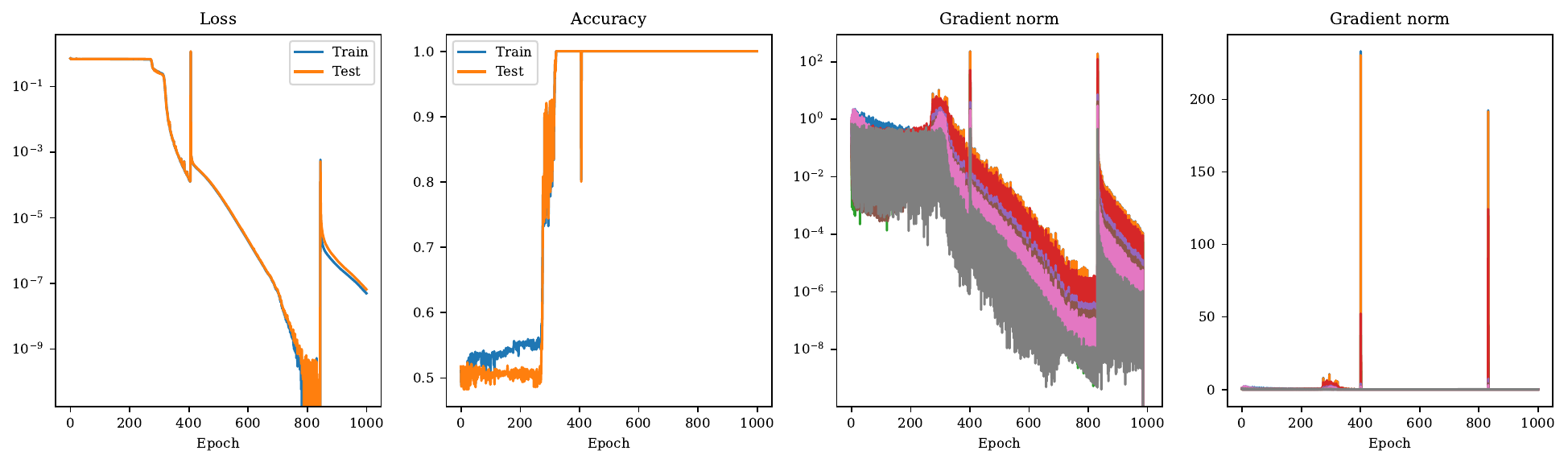}
    \includegraphics[width=\linewidth]{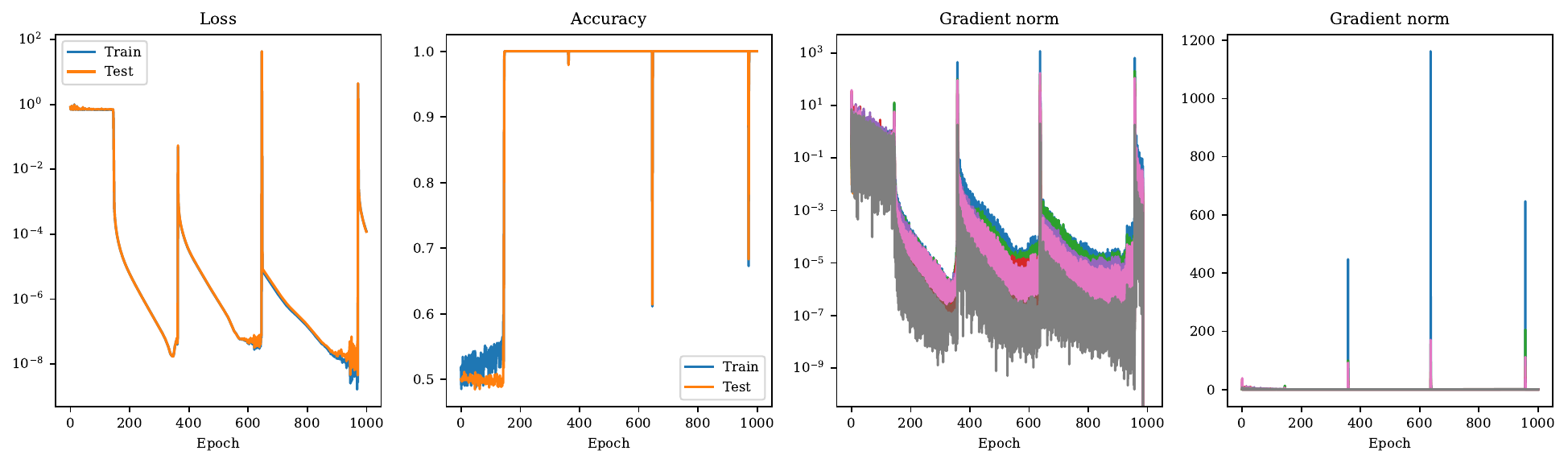}
    \caption{\emph{Loss spikes and gradient norms}. This study is akin to \cref{fig:grad-norms} in mini-batch with $d=3$ in the \textbf{top} plot and $d=64$ in the \textbf{bottom} one. Again, we observe that the loss spikes and gradient norm spikes appear in tandem.}
    \label{fig:grad-norms-d-3-64}
\end{figure}

\paragraph{Loss spikes and gradient norms}
\label{app:loss_spikes_grad_norms}
In this section, we extend the experiments of \cref{sec:loss_spikes} for $d>2$. In \cref{fig:grad-norms-d-3-64}, we consider $d=3$ (top) and $d=64$ (bottom). We observe a similar behavior as in the case $d=2$ in \cref{fig:grad-norms} where the loss spikes and gradient norm spikes occur at the same time. This again highlights the connection between gradient norms and loss spikes. 

\begin{figure}[!h]
    \centering
    \includegraphics[width=\linewidth]{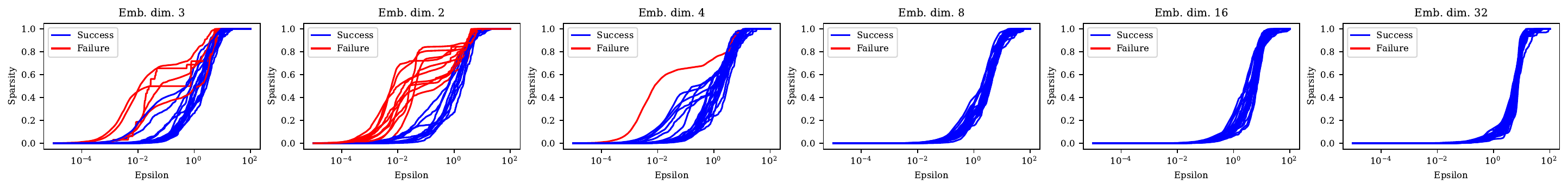}
    \caption{\emph{Connection between performance and sparsity}. We display the evolution of the activation sparsity of $20$ trained models with $\varepsilon \in [10^{-5}, 10^2]$ for $d \in \{2, 3, 4, 8, 16, 32\}$. Successful models (i.e., with test accuracy above $0.9$) in \textcolor{blue}{blue} have less sparse activation than failed models in \textcolor{red}{red}. 
    }
    \label{fig:sparsity_emb_dim}
\end{figure}

\paragraph{Activation sparsity and performance}
\label{app:sparsity_performance}
Finally, we extend the experiments of Appendix~\ref{app:sparsity} with an embedding size $d \in \{2, 3, 4, 8, 16, 32\}$ in \cref{fig:sparsity_emb_dim}. We first note that the higher the embedding size, the more the model succeeds at the task. Especially, as of $d=8$, all the models are successful, i.e., they all achieve an accuracy higher than $0.9$ (as defined in \cref{app:sparsity}).  This is expected given that imposing low-dimensional embeddings limits the expressiveness and the generalization power of our model. It should be noted that this was one of the many challenges of our study: obtaining a generalizable neural network with embeddings in $R^2$ for a mathematical reasoning task such as the sparse modular addition problem. We obtain similar conclusions than in \cref{app:sparsity} with successful models having denser activations.

\section{Theoretical results}
\label{app:theory}
In this section, we detail our theoretical insights into the learning dynamics. It should be noted that all our derivations are conducted with an embedding dimension of $d$ and our transformer block contains the key components of its larger counterparts (see \cref{sec:setting}).

\paragraph{Notations.} In this section, we recall the notations used in our theoretical results and proofs. For any integer $n>0$, we denote $\{1, \cdots, n\}$ as $[n]$. The $i$-th row of a matrix $A \in \R^{n \times m}$ is denoted by $A_i$, its $j$-th
column is denoted by $A_{\cdot,j}$ and its transpose is denoted by by $A^\top$. By abuse of notation, we denote by $\lVert A \rVert$ the entry-wise $\ell_2$-norm of the matrix $A$, which amounts to computing the Euclidean norm of $A$ flattened. Note that we have $\lVert A \rVert = \lVert A \rVert_\mathrm{F} = \sqrt{\sum_{i=1}^{\min\{n,m\}} \sigma_i(A)^2}$, where $\lVert \cdot \rVert_\mathrm{F}$ is the Frobenius norm of $A$ and the $\sigma_i(A)$ are its singular values.The operator norm (also called spectral norm) of $A$ writes $\lVert A \rVert_{\mathrm{op}} = \sigma_{\mathrm{max}}(A)$, with $\sigma_{\mathrm{max}}(A)$ the largest singular value of $A$. The max norm of $A$ is denoted by $\lVert A \rVert_{\infty} = \max_{ij} \lvert A_{ij} \rvert$. The identity
matrix of size $n$ is denoted by $I_n \in \R^{n \times n}$. The indicator function is denoted by $\mathbbm{1}$ and verifies $\mathbbm{1}\{a=b\}$ with value $1$ if $a=b$ and $0$ otherwise. The transpose of a vector $x$ is denoted by $x^\top$, and its Euclidean norm is denoted by $\lVert x \rVert = \sqrt{x^\top x}$. The total variation distance between two probability distributions $P_1, P_2$ is denoted by $\TV{P_1}{P_2}$. Let $d$ be the embedding dimension, $h$ be the feedforward hidden dimension, and $N$ be the input sequence length. For ease of notation, we define the following quantities for $\xi \in \R^d, W \in \R^{h \times d}, z \in \R^{d \times N}, q \in \R^d$ and the GeLU function $\sigma \colon \R \to \R$:
\begin{equation}
\label{eq:bar_xi}
    \bar{\xi} = \frac{\xi}{\lVert \xi \rVert} \in \R^d,
\end{equation}

\begin{equation}
\label{eq:sigma_z}
    A_z = \mathrm{diag}\mleft(\operatorname{softmax}(\frac{z^\top q}{\sqrt{d}}) \mright) \in \R^{N \times N},
\end{equation}

\begin{equation}
\label{eq:phi_xi}
    \Sigma_{\bar{\xi}} = \mathrm{diag}(\sigma'(W \bar{\xi})) \in \R^{h \times h}.
\end{equation}

\begin{equation}
    \label{eq:def_delta}
    \Delta_z = \frac{z}{\sqrt{d}} \mleft[ I_N - \operatorname{softmax}(\frac{z^\top q}{\sqrt{d}})\mathbbm{1}_N^\top\mright].
\end{equation}

\subsection{Gradients derivations}
\label{app:grad_derivations}
In this section, we derive the gradient of the loss with respect to the neural network's weights. To avoid cumbersome derivations that do not fundamentally alter the analysis, and since we are primarily interested in the dynamics at work in the attention and feed-forward layers, we assume that the embedding matrices $ E$ and $ P$ are randomly initialized and do not participate in the optimization. It also concerns the unembedding matrix $W_U$ thanks to weight tying. This is a common assumption in prior works studying the training dynamics of transformers~\citep{cabannes2024scaling, bietti2023birth, cabannes2024learning, nichani2024factualrecall}. To derive the loss gradients, we first recall the quantities involved and reformulate the loss.

\begin{enumerate}
    \item \textbf{Input}: $x = [x_1, \ldots, x_N] \in (\mathbb{F}_p)^L$,
    \item \textbf{Embedding}: $z_t := Z(x_t, t) := \frac{E(x_t) + P(t)}{\norm{E(x_t) + P(t)}} \in \R^d$ with $P, E$ learnable,
    \item \textbf{Attention}: $\xi := (V z) \operatorname{softmax}(\frac{z^\top q}{\sqrt{d}}) \in \R^d$ with $V \in \R^{d \times d}, q \in \R^d$ randomly fixed following~\citep{bietti2023birth},
    \item \textbf{Feed-forward}: $\psi := \xi + \sum_{i \in [h]} u_i \cdot \sigma\mleft(\frac{w_i^\top \xi}{\|\xi\|}\mright) = \xi + U \sigma\mleft(\frac{W\xi}{\|\xi\|}\mright) \in \mathbb{R}^d$ where $U \in \mathbb{R}^{d \times h}$ has columns $(u_i)_{i=1}^h$, $W \in \mathbb{R}^{h \times d}$ has rows $(w_i^\top)_{i=1}^h$ and $U, W$ are learnable,
    \item \textbf{Unembedding}: $\zeta = (E(v)^\top \psi)_{v \in \mathbb{F}_p} = W_U \psi \in \mathbb{R}^n$ where $W_U \in \R^{p \times d}$ has rows $E(v)^T$.
\end{enumerate}

\paragraph{Gradient of the training loss.} The vector $\theta$ encompasses all the model's parameters. More specifically, $\theta$ can be seen as the (flattened) concatenation of $q \in \R^d, V \in \R^{d \times d}, W \in \R^{h \times d}, U \in \R^{d \times h}$, i.e., as a vector of $R^m$ with $m = d (1 + d + 2h)$. We recall that the cross-entropy loss can be reformulated as
\[
\mathcal{L}(\theta) := \mathbb{E}_{x,y}\mleft[\ell(\theta; x, y\mright],
\]
with the expectation taken over training samples $(x,y)$, and
where we have
\begin{equation}
\label{eq:loss_reformulation}
\ell(\theta; x, y) =- \sum_{i=1}^p \mathbbm{1}\{y=i\} \log(\frac{\exp(\zeta_i)}{\sum_{n=1}^p \exp(\zeta_n)}) = \log(\sum_{n=1}^p \exp(\zeta_n)) - \sum_{i=1}^p \mathbbm{1}\{y=i\} \cdot \zeta_i.
\end{equation}
We can now proceed with computing the gradient of the loss function. The following lemma, whose proof is deferred to \cref{app:proof_grad_loss}, gives the closed-form expression of the gradient.
\begin{boxlem}[Full statement of~\cref{lem:closed_form_grad_informal}]
\label{lem:grad_loss}
Let $\Delta_z= \frac{z}{\sqrt{d}} \mleft[ I_N - \operatorname{softmax}(\frac{z^\top q}{\sqrt{d}})\mathbbm{1}_N^\top\mright] \in \R^{d \times N}$, $A_z = \mathrm{diag}\mleft(\operatorname{softmax}(\frac{z^\top q}{\sqrt{d}}) \mright) \in \R^{N \times N}$, $\Sigma_{\bar{\xi}} = \mathrm{diag}(\sigma'(W \bar{\xi})) \in \R^{h \times h}$, $M = \frac{1}{\lVert \xi \rVert} \mleft( I_d - \bar{\xi} \bar{\xi}^\top\mright) \in \R^{d \times d}$ and $Q = U \Sigma_{\bar{\xi}} W M \in \R^{d \times d}$. Assume that the embedding weights $E, P$ are not trainable and let $[\cdot]$ indicate vector concatenation where matrices are flattened into vectors whenever needed. The gradient of the loss function is a vector $\nabla_\theta \mathcal{L}(\theta) \in \R^m$ with $m = d(1 + d + 2h)$ that writes
    \begin{equation*}
    \nabla \mathcal{L}(\theta) = \sum_{j=1}^p \mathbb{E}_x \mleft[ (\hat{p}_{\theta}(y=j|x) - p(y=j|x))\cdot \nabla_\theta \zeta_j \mright],
    \end{equation*}
where 
\begin{equation*}
\begin{cases}
\nabla_\theta \zeta_j &=\quad [\nabla_q \zeta_j, \nabla_V \zeta_j, \nabla_W \zeta_j, \nabla_U \zeta_j] \in \R^m, \\ 
& \\
\nabla_q \zeta_j &=\quad \Delta_zA_z (Vz)^\top \mleft( I_d + Q^\top\mright) E(j) \in \R^d,\\
& \\
\nabla_V \zeta_j &=\quad \mleft( I_d + Q^\top \mright) E(j) \mleft[z \operatorname{softmax}(\frac{z^\top q}{\sqrt{d}})\mright]^\top \in \R^{d \times d},\\
& \\
\nabla_W \zeta_j &=\quad \Sigma_{\bar{\xi}} U^\top E(j) \bar{\xi}^\top \in \R^{h \times d},\\
& \\
\nabla_U \zeta_j &=\quad E(j)\sigma(W\bar{\xi})^\top \in \R^{d \times h}.
\end{cases}
\end{equation*}
\end{boxlem}
 We first notice the importance of the token embeddings since each gradient depends (linearly) on the $E(j)$. We can then see that the attention gradients are both controlled by $ \mleft( I_d + Q^\top \mright) E(j)$. We note that $M = \frac{1}{\lVert \xi \rVert} \mleft( I_d - \bar{\xi} \bar{\xi}^\top\mright)$ projects any vector into $\bar{\xi}$’s orthogonal complementary space, which is included in the unit-sphere of $R^d$ denoted by $\mathbb{S}^d = \{ x \in \R^d \text{ s.t. } \lVert x \rVert = 1\}$. Similar orthogonal projections appear in the derivations of~\citet{tian2023scan} and are used to mimic normalization layers in~\citet{geshkovski2023clusters}. With $M$ symmetric, $Q^\top = M (U \Sigma_{\bar{\xi}} W)^\top$ simply perturbs the vector $E(j)$ before projecting it in the sphere, and $\mleft( I_d + Q^\top \mright) E(j)$ can be seen as a perturbed residual connection. Finally, we observe that the gradients of the feed-forward weights have a linear dependency on $U, W$, which seems to allow implementing the decision boundaries once the sequence embeddings are clustered (see \cref{sec:training_dyn}).  Given that the gradient formulae in \cref{lem:grad_loss} are not straightforward to analyze due to the high dimension and the cross-dependency of the terms, we do not attempt to study the gradient updates, noting that a precise analysis is an important theoretical question for future work. 
\begin{rmk}[Faithful to practice]
The derivations to obtain the gradients in closed form are somewhat involved due to the complexity of the network since we do not theoretically simplify the transformer architecture or the data contrary to most prior works that either model associative memories~\citep{cabannes2024learning, cabannes2024scaling, bietti2023birth, nichani2024factualrecall}, consider orthogonal token embeddings~\citep{bietti2023birth, tian2023scan}, linearize the attention~\citep{nichani2024factualrecall, zhang2024linear, mahankali2024one, ahn2023transformers}, discard residual connections~\citep{tian2023scan, geshkovski2023clusters, jelassi2022vision} or the feed-forward block~\citep{geshkovski2023clusters, edelman2024evolution, mahankali2024one, ahn2023transformers, tian2023scan, jelassi2022vision}. Moreover, we preserve the classification setting with the cross-entropy loss used in practice instead of considering the regression case with a square loss~\citep{chistikov2023learning, pesme2023saddletosaddle}. While this ensures our analysis stays close to the practice, it also comes with technical challenges.
\end{rmk}

\subsection{Stationary points characterization}
\label{app:stationary_points}
The neural networks considered in this work are learned via gradient descent, which converges to one of its fixed points~\citep{nesterov2014convex}. Such points are \emph{stationary points} (or \emph{critical points}) of the loss function~\citep{chi2019nonconvex}, formally defined below.
\begin{boxdef}[Stationary points]
\label{def:stationary_points}
    The stationary points of $\mathcal{L}$ are the solutions of the Euler equation 
    \begin{equation}
    \label{eq:stationary_points}
    \nabla\mathcal{L}(\theta) = 0.
    \end{equation}
    A stationary point $\bar{\theta}$ can be either a local minimum, a local maximum, or a saddle point of $\mathcal{L}$. 
\end{boxdef}
In deep learning, since most objectives are non-convex and in high dimensions, minima\footnote{We omit the mention of maxima since we are mostly interested in minimization problems.} are rarely global, and the gradient descent likely converges to a local minimum or a saddle point. In particular, saddle points are prevalent in deep neural networks' loss landscape~\citep{dauphin2014identifyingattackingsaddlepoint}, which can hinder the optimization. 
\begin{rmk}[Beyond stationary points]
We note that distinguishing the critical points between local minimum, local maximum, or saddle points involves computing the Hessian of the training loss~\citep{chi2019nonconvex}, which is beyond the scope of this work. 
\end{rmk}
As a first step towards better understanding the training dynamics, we show in the following theorem that \emph{clustering heads} can be learned by gradient descent. The proof is deferred to \cref{app:proof_stationary_points}. 
\begin{boxprop}[Full statement of~\cref{prop:learning_clustering_heads_informal}]
\label{prop:stationary_points}
Let $\Delta_z= \frac{z}{\sqrt{d}} \mleft[ I_N - \operatorname{softmax}(\frac{z^\top q}{\sqrt{d}})\mathbbm{1}_N^\top\mright] \in \R^{d \times N}$, $A_z = \mathrm{diag}\mleft(\operatorname{softmax}(\frac{z^\top q}{\sqrt{d}}) \mright) \in \R^{N \times N}$, $\Sigma_{\bar{\xi}} = \mathrm{diag}(\sigma'(W \bar{\xi})) \in \R^{h \times h}$, $M = \frac{1}{\lVert \xi \rVert} \mleft( I_d - \bar{\xi} \bar{\xi}^\top\mright) \in \R^{d \times d}$, $Q = U \Sigma_{\bar{\xi}} W M \in \R^{d \times d}$, $\bar{Q} = \bar{U} \Sigma_{\bar{\xi}} \bar{W} M \in \R^{d \times d}$ and $\mathscr{C}_x = \sum_{j=1}^p (\hat{p}_{\theta}(y=j|x) - p(y=j|x)) E(j)$. The stationary points $\theta$ of $\mathcal{L}$ verify
    \begin{equation}
    \label{eq:stationary_conditions}
    \begin{split}
        &\mathbb{E}_x\mleft[\Delta_zA_z (Vz)^\top \mleft( I_d + Q^\top\mright) \mathscr{C}_x\mright] = 0, \\
        &\mathbb{E}_x\mleft[\mleft( I_d + Q^\top \mright) \mathscr{C}_x \mleft[z \operatorname{softmax}(\frac{z^\top q}{\sqrt{d}})\mright]^\top\mright] =0,\\
        &\mathbb{E}_x\mleft[ \Sigma_{\bar{\xi}} U^\top \mathscr{C}_x \bar{\xi}^\top \mright] =0,\\
        &\mathbb{E}_x\mleft[ \mathscr{C}_x\sigma(W\bar{\xi})^\top\mright]=0.
        \end{split}
    \end{equation}
Moreover, \cref{eq:stationary_conditions} holds for transformers that implement clustering heads. This implies that such circuits can be learned by gradient descent. 
\end{boxprop}
Proposition~\ref{prop:stationary_points} allows us to access the final configuration of the model in the form of a system of equations. It involves the main quantities of the problem: the trainable weights $q, V, W, U$, the embedding tokens $E(j)$, and the training data $(x, y)$. A particularly interesting observation is the linear dependency of each equation on the term $\mathscr{C}_x$. This term captures the model's prediction discrepancy with the ground truth through the $\hat{p}_{\theta}(y=j|x) - p(y=j|x)$ and plays an important role in the proof. In particular, once the stationary points are characterized, showing that clustering heads can be learned by gradient descent is rather straightforward using the $\mathscr{C}_x$. In addition, it shows that the natural way we proposed to solve the sparse modular addition problem also qualifies as a fixed-point of gradient descent. 
\begin{rmk}[Other pathways]
While \cref{prop:stationary_points} shows that clustering heads are viable fixed points of the gradient descent, it does not provide conditions to ensure it is reached, nor qualify them as local minima, global minima, or saddle points. This interesting research direction is left for future work. We note that our experimental investigation suggests that clustering heads solve the sparse modular addition problem, but there may be other pathways to do so, too. In addition, as discussed in \cref{sec:clustering_heads_implementation}, the pathways learned in practice can differ from the idealized circuit, although the variations still capture most of the invariants of the problem.
\end{rmk}

\subsection{Bounding the gradients}
\label{app:grad_bounds}
In \cref{sec:training_dyn}, we connect some of our findings with the evolution of gradient norms along the training. Leveraging the closed-form expressions given in \cref{lem:grad_loss}, we can study upper-bound gradient norms to elucidate their dependency on the parameters and data. The following proposition, whose proof is deferred to \cref{app:proof_grad_bounds}, makes those upper bounds explicit.
\begin{boxprop}[Full statement of \cref{prop:informal_gradient_bounds}]
    \label{prop:gradients_bound}
     Assume the token embedding is bounded, i.e., there exists $B > 0$ such that $\lVert E \rVert_{\infty} \leq B$. We introduce $\mathscr{E} = \mathbb{E}_x\mleft[\TV{\hat{p}_{\theta}(\cdot|x)}{p(\cdot|x)}\mright]$ that captures the ability of the model to solve the sparse modular addition problem task. We have
     \[
     \lVert \nabla \mathcal{L}(\theta)\rVert = \mathcal{O}\mleft(\tilde{B} \cdot \sqrt{\mathscr{E}}\mright),
     \]
where 
\[
\tilde{B} = B \sqrt{\lVert V \rVert_\mathrm{op} \mleft[ 1 + \lVert U \rVert_\mathrm{op}\lVert W \rVert_\mathrm{op}^2\mright]^2 + \mleft[1 +  \lVert U \rVert_\mathrm{op}\lVert W \rVert_\mathrm{op} \mright]^2 +  \lVert U \rVert_\mathrm{op}^2 + \lVert W \rVert_\mathrm{op}^2}.
\]
\end{boxprop}
This result is of interest for two main reasons. The first one is that it can be used to understand the loss profiles identified in \cref{sec:training_dyn}. Notably, we see that the term $\mathscr{E}$ captures the ability of the model to correctly solve the task. In the early iterations, the model behaves poorly, which imposes weak control on the gradient norm. This allows for large gradient norms and hence for bigger updates to the weights. The loss can vary rapidly, as can be seen by the steep descent of \cref{fig:dynamics}. Along with the training, the model improves, as can be seen by the increase in accuracy (both train and test). This leads to a smaller $\mathscr{E}$ and strengthens the control on the norm. In parallel, the gradient updates are of smaller magnitude, as well as the loss variations. This goes along a less steep descent in the loss as observed in \cref{fig:dynamics}. The second benefit of \cref{prop:gradients_bound} is that one could derive sufficient conditions on the term related to the weight matrices and on the term related to classification to achieve an approximate critical point~\citep{chi2019nonconvex}, that is parameters $\theta$ such that the gradient verifies $\lVert\nabla \mathcal{L}(\theta)\rVert \leq \varepsilon$ for some $\varepsilon > 0$. Since finding local approximate minima can be faster (and more beneficial) than finding critical points~\citep{agarwal2017approximate}, practitioners could derive optimization algorithms that impose an appropriate bound on the two terms of \cref{prop:gradients_bound} instead of using gradient descent. These interesting considerations are left for future work. 

\section{Proofs}
\label{app:proofs}
In this section, we provide detailed proof of our theoretical results.
\subsection{Proof of \cref{lem:closed_form_grad_informal}}
\label{app:proof_grad_loss}
We detail below the proof of \cref{lem:grad_loss}.
\begin{proof}
The proof consists of three main steps: 
\begin{enumerate}
\item Expressing the gradient of the loss with respect to the gradients of the attention embeddings,
\item Deriving the gradients for each trainable weight,
\item Putting everything together.
\end{enumerate}
We first formulate the gradient of the loss with respect to the gradient of the attention embeddings. The next lemma provides the formula.
\begin{boxlem}[Gradient compact formulation]
\label{lem:grad_compact_formulation}
    Denoting the model's trainable parameters by $\theta \in \R^m$ and recalling that the attention embedding $\xi \in \R^d$ can be seen as a function of $\theta$, we have
    \begin{equation}
        \label{eq:grad_compact_formulation}
    \nabla_\theta \mathcal{L}(\theta) = \sum_{j=1}^p \mathbb{E}_x \mleft[ (\hat{p}_{\theta}(y=j|x) - p(y=j|x))\cdot \nabla_\theta \zeta_j \mright]
    \end{equation}
\end{boxlem}
\begin{proof}
Noting that the logits can be written as $\zeta = g(\theta; x, y)$ for some $g$, the loss\footnote{We keep denoting the loss by $\ell$ by abuse of notation.} can then be written as $\ell(\theta; x, y) = \ell(\zeta) = \ell (g(\theta; x, y))$. Using the chain rule, with $D$ denoting the Jacobian operator, we have
\[
\begin{split}
\nabla_{\theta} \mathcal{L}(\theta) &= \mathbb{E}_{x,y}\mleft[\nabla_{\theta} \ell(\theta)\mright] \\
&= \mathbb{E}_{x,y}\mleft[\nabla_{\theta} \ell (g(\theta))\mright] \\
&= \mathbb{E}_{x,y}\mleft[(D_\theta g(\theta))^\top (D_{g(\theta)} \ell(g(\theta)))^\top \mright]\\
&= \mathbb{E}_{x,y}\mleft[\nabla_\theta g(\theta)\nabla_{\zeta} \ell (\zeta)\mright].
\end{split}
\]
By definition, we have $\nabla_\theta g(\theta) = (\nabla_\theta \zeta_1, \ldots, \nabla_\theta \zeta_p)$ and $\nabla_{\zeta}\ell (\zeta) = (\frac{\partial \ell}{\partial \zeta_j} (\zeta)) \in \mathbb{R}^{p}$. Hence, the matrix-vector product writes 
\[
\begin{split}
(\nabla_\theta g(\theta))\nabla_{\zeta} \ell (\zeta) &= \sum_{j=1}^p \frac{\partial \ell}{\partial \zeta_j} (\zeta) \cdot [\nabla_\theta g(\theta)]_{\cdot, j}  \\
&= \sum_{j=1}^p \frac{\partial \ell}{\partial \zeta_j} (\zeta) \cdot \nabla_\theta \zeta_j.
\end{split}
\]
Using Eq.~\eqref{eq:loss_reformulation}, we have
\[
\begin{split}
\frac{\partial \ell}{\partial \zeta_j} (\zeta) &= \frac{\exp(\zeta_j)}{\sum_{n=1}^p \exp(\zeta_n)} - \mathbbm{1}\{y=j\}\\
&= \operatorname{softmax}(\zeta)_j - \mathbbm{1}\{y=j\} \\
&= \hat{p}_{\theta}(y=j|x) - \mathbbm{1}\{y=j\},
\end{split} 
\]
Putting everything together, we obtain
\[
\begin{split}
\nabla_\theta \mathcal{L}(\theta) &= \mathbb{E}_{x,y}\mleft[ \sum_{j=1}^p (\hat{p}_{\theta}(y=j|x) - \mathbbm{1}\{y=j\}) \cdot \nabla_\theta \zeta_j\mright] \\
&=  \sum_{j=1}^p \mathbb{E}_{x,y}\mleft[(\hat{p}_{\theta}(y=j|x) - \mathbbm{1}\{y=j\}) \cdot \nabla_\theta \zeta_j\mright] \\
&=  \sum_{j=1}^p \mathbb{E}_x \mleft[ \mathbb{E}_y \mleft[ (\hat{p}_{\theta}(y=j|x) - \mathbbm{1}\{y=j\}) \cdot \nabla_\theta \zeta_j | x\mright]\mright],
\end{split}
\]
where the first equality comes from the linearity of the expectation and the second stems from the property of the conditional expectation. Using the fact that $\hat{p}_{\theta}(y=j|x)$ and $\nabla_\theta \zeta_j$ do not depend on the random variable $y$ leads us to
\[
\begin{split}
&\mathbb{E}_y \mleft[ (\hat{p}_{\theta}(y=j|x) - \mathbbm{1}\{y=j\}) \cdot \nabla_\theta \zeta_j | x\mright] \\
&\qquad = (\hat{p}_{\theta}(y=j|x) - \mathbb{E}_y \mleft[\mathbbm{1}\{y=j\} | x\mright])\cdot \nabla_\theta \zeta_j \\
&\qquad = (\hat{p}_{\theta}(y=j|x) - p(y=j|x))\cdot \nabla_\theta \zeta_j,
\end{split}
\]
where the last equality comes from the property of the conditional expectation and the fact that $x,y\sim p(\cdot)$. Indeed, we have
\[
\mathbb{E}_y \mleft[\mathbbm{1}\{y=j\} | x\mright] = \sum_{l=1}^p  \mathbbm{1}\{l=j\} p(y=l|x) = p(y=j|x).
\]
In summary, the gradient of the loss writes
\[
\nabla_\theta \mathcal{L}(\theta) = \sum_{j=1}^p \mathbb{E}_x \mleft[ (\hat{p}_{\theta}(y=j|x) - p(y=j|x))\cdot \nabla_\theta \zeta_j \mright].
\]
\end{proof}
\cref{lem:grad_compact_formulation} implies that deriving the $\nabla_\theta \zeta_j = \mleft(\frac{\partial \zeta_j}{\partial \theta_n}\mright)_{n=1}^m$ allows us to obtain the final gradient formula. The following lemmas provide for a fixed $j$ the gradients of $\xi_j$ with respect to the different parts of the neural networks, namely the feed-forward weights $U$ and $W$ and the attention weights $q$ and $V$. 

\subsubsection{Attention derivations}
In this section, we derive the gradients with respect to the attention weights.

\paragraph{Attention query.} The following lemma provides the formula of $\nabla_q \xi_j$.
\begin{boxlem}[Query]
    \label{lem:grad_xi_query}
Let $\Delta_z= \frac{z}{\sqrt{d}} \mleft[ I_N - \operatorname{softmax}(\frac{z^\top q}{\sqrt{d}})\mathbbm{1}_N^\top\mright] \in \R^{d \times N}$ following Eq.~\eqref{eq:def_delta}, $A_z = \mathrm{diag}\mleft(\operatorname{softmax}(\frac{z^\top q}{\sqrt{d}}) \mright) \in \R^{N \times N}$ following Eq.~\eqref{eq:sigma_z}, $\Sigma_{\bar{\xi}} = \mathrm{diag}(\sigma'(W \bar{\xi})) \in \R^{h \times h}$ following Eq.~\eqref{eq:phi_xi}, $M = \frac{1}{\lVert \xi \rVert} \mleft( I_d - \bar{\xi} \bar{\xi}^\top\mright) \in \R^{d \times d}$ following Eq.~\eqref{eq:def_notations} and $Q = U \Sigma_{\bar{\xi}} W M \in \R^{d \times d}$. The entries of $\nabla_\theta \zeta_j$ associated to the value $V$ are the entries of $\nabla_q \zeta_j \in \R^d$ where we have
    \begin{equation*}
    \nabla_q \zeta_j = \Delta_zA_z \mleft[ \mleft( I_d + Q\mright) Vz \mright]^\top E(j).
    \end{equation*}
\end{boxlem}
\begin{proof}
We want to compute the $\frac{\partial \zeta_j}{\partial \theta_n}$ for the $\theta_n$ that are entries of $q$. We first note that 
\begin{equation}
\label{eq:zeta_j}
    \frac{\partial \zeta_j}{\partial \theta_n} = \sum_{l=1}^p (W_U)_{jl} \frac{\partial \psi_l}{\partial \theta_n},
\end{equation}
where we have thanks to weight tying, $(W_U)_{jl} = E(j)_l$ is fixed. Moreover, we have
\begin{equation}
\label{eq:psi_l}
    \psi_l = \xi_l + \sum_{i=1}^h (u_i)_l \sigma(w_i^\top \bar{\xi}),
\end{equation}
where we recall that $\bar{\xi} = \frac{\xi}{\lvert \xi \rVert}$ following Eq.~\eqref{eq:bar_xi}. Considering $n$ such that $\theta_n = q_a$ for some $a \in [d]$, we have 
\begin{equation}
\label{eq:psi_q_a}
\frac{\partial \psi_l}{\partial \theta_n} = \frac{\partial \psi_l}{\partial q_a} = \frac{\partial \xi_l}{\partial q_a} + \frac{\partial}{\partial q_a}(\sum_{i=1}^h (u_i)_l \sigma(w_i^\top \bar{\xi})) = \underbrace{\frac{\partial f_1(q_a)}{\partial q_a}}_{\mathrm{LHS}} + \underbrace{\sum_{i=1}^h (u_i)_l \frac{\partial f_2^i(q_a)}{\partial q_a}}_{\mathrm{RHS}},
\end{equation}
with 
\[
\begin{cases}
f_1 \colon& \R \to \R, \quad q_a \mapsto \xi_l \\
f_2^i \colon& \R \to \R, \quad q_a \mapsto (\sigma \circ f_3^i \circ f_4^i \circ f_5^i)(q_a) = \sigma(w_i^\top \bar{\xi})\\
    f_3^i \colon& \R^d \to \R, \quad \bar{\xi} \mapsto w_i^\top \bar{\xi} \\
    f_4^i \colon& \R^d \to \R^d, \quad \xi \mapsto \bar{\xi} = \frac{\xi}{\lVert \xi \rVert} \\
    f_5^i \colon& \R \to \R^d, \quad q_a \mapsto \xi. \\
\end{cases}
\]
We proceed in two steps and first compute the LHS of Eq.~\eqref{eq:psi_q_a}. 

\underline{\bfseries LHS of Eq.~\eqref{eq:psi_q_a}:} we first note that
\[
\xi_l = \sum_{t=1}^N (Vz)_{lt} \operatorname{softmax}(\frac{z^\top q}{\sqrt{d}})_t.
\]
This leads to
\begin{equation}
\label{eq:psi_l_q_a}
\frac{\partial f_1(q_a)}{\partial q_a} = \frac{\partial \xi_l}{\partial q_a} = \sum_{t=1}^N (Vz)_{lt} \frac{\partial s_t}{\partial q_a},
\end{equation}
where 
\begin{equation}
\label{eq:def_s_t}
s_t = \operatorname{softmax}(\frac{z^\top q}{\sqrt{d}})_t = \frac{\exp{(\frac{1}{\sqrt{d}}\sum_{o=1}^d z_{ot}q_o)}}{\sum_{t'=1}^N \exp{(\frac{1}{\sqrt{d}}\sum_{o=1}^d z_{ot'}q_o)}}.
\end{equation}
This can be rewritten as
\[
s_t = \frac{\exp{(\frac{1}{\sqrt{d}}z_{at}q_a + \frac{1}{\sqrt{d}}\sum_{o\neq a} z_{ot}q_o)}}{\sum_{t'=1}^N \exp{(\frac{1}{\sqrt{d}}z_{at'}q_a + 
 \frac{1}{\sqrt{d}}\sum_{o\neq a} z_{ot'}q_o)}} = \frac{g(q_a)}{\tilde{g}(q_a)},
\]
where
\[
\begin{cases}
&g \colon x \to \exp{(\frac{1}{\sqrt{d}}z_{at}x + \frac{1}{\sqrt{d}}\sum_{o\neq a} z_{ot}q_o)} \\
& \tilde{g} \colon x \to \sum_{t'=1}^N \exp{(\frac{1}{\sqrt{d}}z_{at'}x + \frac{1}{\sqrt{d}}\sum_{o\neq a} z_{ot'}q_o)}
\end{cases}.
\]
This leads to
\[
\frac{\partial s_t}{\partial q_a} = \frac{g'(q_a) \tilde{g}(q_a) - \tilde{g}'(q_a) g(q_a)}{[\tilde{g}(q_a)]^2}.
\]
By simple application of the chain rule, we have
\begin{align*}
g'(q_a) &= \frac{1}{\sqrt{d}}z_{at} \exp{(\frac{1}{\sqrt{d}}\sum_{o=1}^d z_{ot}q_o)} \\
&= \frac{1}{\sqrt{d}}z_{at} \exp{(\frac{1}{\sqrt{d}}z_{\cdot, t}^\top q)}.
\end{align*}
Similarly, we have
\[
\tilde{g}'(q_a) = \sum_{t'=1}^N\frac{1}{\sqrt{d}}z_{at'} \exp{(\frac{1}{\sqrt{d}}z_{\cdot, t'}^\top q)}.
\]
It follows that
\begin{align*}
g'(q_a) \tilde{g}(q_a) &= \frac{1}{\sqrt{d}}z_{at} \exp{(\frac{1}{\sqrt{d}} z_{\cdot, t}^\top q)} \sum_{t'=1}^N \exp{(\frac{1}{\sqrt{d}} z_{\cdot, t'}^\top q)} \\
&= \frac{1}{\sqrt{d}}z_{at} \sum_{t'=1}^N \exp{(\frac{1}{\sqrt{d}}\sum_{o=1}^d (z_{\cdot, t} + z_{\cdot, t'})^\top q)}
\end{align*}
and similarly
\begin{align*}
g(q_a)\tilde{g}'(q_a) &= \exp{(\frac{1}{\sqrt{d}} z_{\cdot, t}^\top q)} \sum_{t'=1}^N\frac{1}{\sqrt{d}}z_{at'} \exp{(\frac{1}{\sqrt{d}} z_{\cdot, t'}^\top q)} \\
&= \frac{1}{\sqrt{d}} \sum_{t'=1}^N z_{at'} \exp{(\frac{1}{\sqrt{d}}\sum_{o=1}^d (z_{\cdot, t} + z_{\cdot, t'})^\top q)}.
\end{align*}
Putting everything together, we obtain
\begin{align*}
\frac{\partial s_t}{\partial q_a} &= \frac{1}{\sqrt{d}} \frac{\sum_{t'=1}^N (z_{at} - z_{at'}) \exp{(\frac{1}{\sqrt{d}} (z_{\cdot, t} + z_{\cdot, t'})^\top q)}}{[\sum_{r=1}^N \exp{(\frac{1}{\sqrt{d}}\sum_{o=1}^d z_{or}q_o)}]^2} \\
&= \frac{1}{\sqrt{d}} \sum_{t'=1}^N (z_{at} - z_{at'}) \frac{\exp{(\frac{1}{\sqrt{d}}z_{\cdot, t}^\top q)}}{\sum_{r=1}^N \exp{(\frac{1}{\sqrt{d}}\sum_{o=1}^d z_{or}q_o)}} \cdot \frac{\exp{(\frac{1}{\sqrt{d}}z_{\cdot, t'}^\top q)}}{\sum_{r=1}^N \exp{(\frac{1}{\sqrt{d}}\sum_{o=1}^d z_{or}q_o)}} \\
&= \frac{1}{\sqrt{d}} \sum_{t'=1}^N (z_{at} - z_{at'}) \frac{\exp{(\frac{1}{\sqrt{d}}\sum_{o=1}^d z_{ot} q_o)}}{\sum_{r=1}^N \exp{(\frac{1}{\sqrt{d}}\sum_{o=1}^d z_{or}q_o)}} \cdot \frac{\exp{(\frac{1}{\sqrt{d}}\sum_{o=1}^d z_{ot'} q_o)}}{\sum_{r=1}^N \exp{(\frac{1}{\sqrt{d}}\sum_{o=1}^d z_{or}q_o)}} \\
&= \frac{1}{\sqrt{d}} \sum_{t'=1}^N (z_{at} - z_{at'}) \operatorname{softmax}(\frac{z^\top q}{\sqrt{d}})_t \operatorname{softmax}(\frac{z^\top q}{\sqrt{d}})_{t'} \\
&= \frac{1}{\sqrt{d}} \sum_{t'=1}^N (z_{at} - z_{at'}) s_t s_{t'}.
\end{align*}

Using Eq.~\eqref{eq:psi_l_q_a} leads to
\begin{equation}
\label{eq:f_1_q_a}
\begin{split}
\frac{\partial f_1(q_a)}{\partial q_a} = \frac{\partial \xi_l}{\partial q_a} &= \sum_{t=1}^N (Vz)_{lt} \frac{1}{\sqrt{d}} \sum_{t'=1}^N (z_{at} - z_{at'}) s_t s_{t'} \\
&= \frac{1}{\sqrt{d}} \sum_{tt'} (Vz)_{lt} z_{at}s_ts_{t'} - \frac{1}{\sqrt{d}} \sum_{tt'} (Vz)_{lt} z_{at'}s_ts_{t'} \\
&= \frac{1}{\sqrt{d}} \sum_{t=1}^N (Vz)_{lt} z_{at}s_t \underbrace{\sum_{t'=1}^Ns_{t'}}_{=1} - \frac{1}{\sqrt{d}} \sum_{t=1}^N (Vz)_{lt} s_t \underbrace{\sum_{t=1}^N z_{at'}s_{t'}}_{=\mu(z_a)} \\
&= \frac{1}{\sqrt{d}} \sum_{t=1}^N (z_{at} -\mu(z_a) (Vz)_{lt} s_t \\
&= \sum_{t=1}^N \Delta_{at} (Vz)_{lt}s_t,
\end{split}
\end{equation}
where $\Delta_{at} = \frac{z_{at}- \mu(z_a)}{\sqrt{d}}$ corresponds to the deviation from the mean embedding at dimension $a$, assuming the distribution probability follows the softmax $(s_t)_{t=1}^N$. We note that the $\Delta_{at}$ are the entries of 
\[
\Delta_z= \frac{z}{\sqrt{d}} \mleft[ I_N - \operatorname{softmax}(\frac{z^\top q}{\sqrt{d}})\mathbbm{1}_N^\top\mright] \in \R^{d \times N}
\]
defined in Eq.~\eqref{eq:def_delta}.
Hence, $\frac{\partial f_1(q_a)}{\partial q_a}$ can be understood as the $l$-element of a sentence embedding where the deviation from the mean embedding has rescaled the value matrix. We now proceed to the RHS of Eq.~\eqref{eq:psi_q_a}.

\underline{\bfseries RHS of Eq.~\eqref{eq:psi_q_a}:} a simple application of the chain rule gives
\[
\frac{\partial f_2^i(q_a)}{\partial q_a} = \underbrace{\nabla f_5^i(q_a)}_{\in \R^{1 \times d}} \underbrace{\nabla f_4^i(\xi)}_{\in \R^{d \times d}} \underbrace{\nabla f_3^i(\bar{\xi})}_{\in \R^d} \underbrace{\sigma'(w_i^\top \bar{\xi})}_{\in \R},
\]
where we recall that $\bar{\xi} = \xi / \lVert \xi \rVert$. We have
\[
\nabla f_5^i(q_a) = \frac{\partial \xi}{\partial q_a} = \mleft[\frac{\partial \xi_1}{\partial q_a}, \ldots, \frac{\partial \xi_d}{\partial q_a}\mright],
\]
and 
\[
\nabla f_3^i(\bar{\xi}) = w_i.
\]
For $f_4^i$, we note that $f_4^i(\xi) = (h_1 \circ h_2)(\xi) \cdot \xi$, with $h_1 \colon \R \to \R, t \mapsto \frac{1}{t}$ and $h_2 \colon \R^d \to \R, x \mapsto \lVert x \rVert$. It leads to
\begin{align*}
\nabla f_4^i(\xi) &=  \frac{\partial (h_1 \circ h2)(\xi)}{\partial \xi} \cdot \xi^\top + (h_1 \circ h_2)(\xi) \cdot \frac{\partial \xi}{\partial \xi} \\
&=   \frac{\partial (h_1 \circ h2)(\xi)}{\partial \xi} \cdot \xi^\top + \frac{1}{\lVert \xi \rVert} I_d.
\end{align*}

Since we have $h_2(\xi) = \lVert \xi \rVert = \sqrt{\xi^\top \xi}$, we know that 
\[
 \frac{\partial h2(\xi)}{\partial \xi} = \frac{1}{2 \sqrt{\xi^\top \xi}} \cdot 2\xi = \frac{\xi}{\lVert \xi \rVert}.
\]
It leads to 
\[
\frac{\partial (h_1 \circ h2)(\xi)}{\partial \xi} = -\frac{1}{\lVert \xi \rVert^2} \cdot \frac{\xi}{\lVert \xi \rVert} = -\frac{\xi}{\lVert \xi \rVert^3}.
\]
Putting everything together, we obtain
\[
\nabla f_4^i(\xi) = -\frac{\xi}{\lVert \xi \rVert^3} \xi^\top + \frac{1}{\lVert \xi \rVert} I_d = \frac{1}{\lVert \xi \rVert} \mleft( I_d - \bar{\xi} \bar{\xi}^\top\mright).
\]
Going back to $\frac{\partial f_2^i(q_a)}{\partial q_a}$, we have
\[
\frac{\partial f_2^i(q_a)}{\partial q_a} = \mleft[\frac{\partial \xi_1}{\partial q_a}, \ldots, \frac{\partial \xi_d}{\partial q_a}\mright] \frac{1}{\lVert \xi \rVert} \mleft( I_d - \bar{\xi} \bar{\xi}^\top\mright) w_i \sigma'(w_i^\top \bar{\xi}) = \lambda M w_i \alpha_i,
\]
where we introduce 
\begin{equation}
    \label{eq:def_notations}
    \lambda = \mleft[\frac{\partial \xi_1}{\partial q_a}, \ldots, \frac{\partial \xi_d}{\partial q_a}\mright] \in \R^{1 \times d}, \quad M = \frac{1}{\lVert \xi \rVert} \mleft( I_d - \bar{\xi} \bar{\xi}^\top\mright) \in \R^{d \times d}, \quad \alpha_i = \sigma'(w_i^\top \bar{\xi}),
\end{equation} 
to ease notations. Developing the matrix multiplication, we obtain
\begin{align*}
\lambda M w_i \alpha_i &= \sum_{r=1}^d \frac{\partial \xi_r}{\partial q_a} \mleft[ Mw_i\mright]_r \alpha_i \\
&= \alpha_i \sum_{r=1}^d \sum_{t=1}^N \Delta_{at} (Vz)_{rt}s_t \mleft[ Mw_i\mright]_r \tag{using Eq.~\eqref{eq:f_1_q_a}} \\
&= \alpha_i \sum_{t=1}^N \Delta_{at} s_t \sum_{r=1}^d (Vz)_{rt} \sum_{o=1}^d M_{ro}(w_i)_o \\
&= \alpha_i \sum_{t=1}^N \Delta_{at} s_t \sum_{o=1}^d (w_i)_o \sum_{r=1}^d (Vz)_{rt}  M_{ro} \\
&= \alpha_i \sum_{t=1}^N \Delta_{at} s_t \sum_{o=1}^d (w_i)_o \sum_{r=1}^d  M_{ro} (Vz)_{rt} \\
&= \alpha_i \sum_{t=1}^N \Delta_{at} s_t \sum_{o=1}^d (w_i)_o \sum_{r=1}^d  M^\top_{or} (Vz)_{rt} \\
&= \alpha_i \sum_{t=1}^N \Delta_{at} s_t \sum_{o=1}^d (w_i)_o \mleft[M^\top Vz\mright]_{ot} \\
&= \alpha_i \sum_{t=1}^N \Delta_{at} s_t \sum_{o=1}^d (w_i)_o \tilde{M}_{ot},
\end{align*}
where we introduce 
\begin{equation}
    \label{eq:def_M_tilde}
    \tilde{M} = M^\top Vz \in \R^{d \times N}.
\end{equation}
Now multiplying by $(u_i)_l$ and summing over $i \in [h]$, we obtain
\begin{align*}
\sum_{i=1}^h (u_i)_l \frac{\partial f_2^i(q_a)}{\partial q_a} &= \sum_{i=1}^h (u_i)_l \alpha_i \sum_{t=1}^N \Delta_{at} s_t \sum_{o=1}^d (w_i)_o \tilde{M}_{ot} \\
&= \sum_{t=1}^N \Delta_{at} s_t \sum_{o=1}^d \tilde{M}_{ot} \sum_{i=1}^h \alpha_i (u_i)_l (w_i)_o.
\end{align*}
We can now put everything together to rewrite Eq.~\eqref{eq:psi_q_a}.

\underline{\bfseries LHS + RHS of Eq.~\eqref{eq:psi_q_a}:} We have
\[
\frac{\partial \psi_l }{\partial q_a} = \frac{\partial \xi_l}{\partial q_a} + \sum_{t=1}^N \Delta_{at} s_t \sum_{o=1}^d \tilde{M}_{ot} \sum_{i=1}^h \alpha_i (u_i)_l (w_i)_o.
\]
and Eq.~\eqref{eq:zeta_j} leads to
\begin{equation}
\label{eq:zeta_q_a_comp}
\frac{\partial \zeta_j }{\partial q_a} = \underbrace{\sum_{l=1}^p (W_U)_{jl} \frac{\partial \xi_l}{\partial q_a}}_{=\mathrm{LHS}} + \underbrace{\sum_{l=1}^p (W_U)_{jl} \sum_{t=1}^N \Delta_{at} s_t \sum_{o=1}^d \tilde{M}_{ot} \sum_{i=1}^h \alpha_i (u_i)_l (w_i)_o}_{=\mathrm{RHS}}.
\end{equation}
The LHS can be rewritten as
\begin{align*}
\mathrm{LHS} & = \sum_{l=1}^p (W_U)_{jl} \sum_{t=1}^N \Delta_{at} (Vz)_{lt}s_t \\
&= \sum_{t=1}^N \Delta_{at} s_t \sum_{l=1}^p (W_U)_{jl} (Vz)_{lt} \\
&= \sum_{t=1}^N \Delta_{at} s_t \sum_{l=1}^p E(j)_l (Vz)_{lt} \\
&= \sum_{t=1}^N \Delta_{at} s_t E(j)^\top (Vz)_{\cdot, t} \\
&= \mleft[\sum_{t=1}^N \Delta_{at} s_t [(Vz)_{\cdot, t}] ^\top\mright]E(j).
\end{align*}
The RHS can be rewritten as
\begin{align*}
    \mathrm{RHS} &= \sum_{l=1}^p (W_U)_{jl} \sum_{t=1}^N \Delta_{at} s_t \sum_{o=1}^d \tilde{M}_{ot} \sum_{i=1}^h \alpha_i (u_i)_l (w_i)_o\\
    &= \sum_{t=1}^N \Delta_{at} s_t \sum_{l=1}^p (W_U)_{jl}  \sum_{i=1}^h \alpha_i (u_i)_l \sum_{o=1}^d \tilde{M}_{ot}(w_i)_o \\
    &= \sum_{t=1}^N \Delta_{at} s_t \sum_{l=1}^p (W_U)_{jl}  \sum_{i=1}^h \alpha_i (u_i)_l \sum_{o=1}^d (w_i)_o \tilde{M}_{ot}\\
    &= \sum_{t=1}^N \Delta_{at} s_t \sum_{l=1}^p (W_U)_{jl}  \sum_{i=1}^h \alpha_i (u_i)_l \sum_{o=1}^d W_{io} \tilde{M}_{ot}\\
    &= \sum_{t=1}^N \Delta_{at} s_t \sum_{l=1}^p (W_U)_{jl}  \sum_{i=1}^h \alpha_i (u_i)_l (W \tilde{M})_{it}.
\end{align*}
We now introduce the matrix 
\begin{equation}
    \label{eq:def_P}
    \tilde{P} = U \begin{pmatrix}
    \alpha_1 & 0 & \cdots \\
    \cdots & \cdots & \cdots \\
    \cdots & 0 & \alpha_h 
\end{pmatrix} W\tilde{M} \in \R^{d \times N}.
\end{equation}
Its entry in row $l$ and column $t$ writes
\begin{align*}
\tilde{P}_{lt} &= \mleft[U \begin{pmatrix}
    \alpha_1 & 0 & \cdots \\
    \cdots & \cdots & \cdots \\
    \cdots & 0 & \alpha_h 
\end{pmatrix} W \tilde{M}\mright]_{lt} \\
&= \sum_{i=1}^h \mleft[ U  \begin{pmatrix}
    \alpha_1 & 0 & \cdots \\
    \cdots & \cdots & \cdots \\
    \cdots & 0 & \alpha_h 
\end{pmatrix} \mright]_{li} (W\tilde{M})_{it} \\
&= \sum_{i=1}^h \alpha_i U_{li} (W\tilde{M})_{it} \\
&= \sum_{i=1}^h \alpha_i (u_i)_l (W\tilde{M})_{it} \tag{as $U$ has columns $u_i$}.
\end{align*}
Plugging this term into the computation of the RHS leads to
\begin{align*}
\mathrm{RHS} &= \sum_{t=1}^N \Delta_{at} s_t \sum_{l=1}^p (W_U)_{jl} \tilde{P}_{lt} \\
&= \sum_{t=1}^N \Delta_{at} s_t \sum_{l=1}^p \tilde{P}_{lt} (W_U)_{jl} \\
&= \sum_{t=1}^N \Delta_{at} s_t \sum_{l=1}^p \tilde{P}_{lt} E(j)_l \tag{as $W_U$ has rows $E(j)^\top$} \\
&= \sum_{t=1}^N \Delta_{at} s_t \sum_{l=1}^p \tilde{P}^\top_{tl} E(j)_l \\
&= \sum_{t=1}^N \Delta_{at} s_t \tilde{P}^\top_{t} E(j).
\end{align*} 
Going back to Eq.~\eqref{eq:zeta_q_a_comp}, we have
\begin{align*}
    \frac{\partial \zeta_j }{\partial q_a} &= \mleft[\sum_{t=1}^N \Delta_{at} s_t [(Vz)_{\cdot, t}] ^\top\mright]E(j) + \sum_{t=1}^N \Delta_{at} s_t \tilde{P}^\top_{t} E(j) \\
    &= \sum_{t=1}^N \Delta_{at} s_t \mleft[ (Vz)_{\cdot, t} + \tilde{P}^\top_{t} \mright] E(j) \\
    &= \sum_{t=1}^N \Delta_{at} s_t \mleft[ (Vz)_{\cdot, t} + \tilde{P}_{\cdot, t} \mright] E(j) \\
    &= \sum_{t=1}^N \Delta_{at} s_t \mleft[Vz + \tilde{P}\mright]_{\cdot, t} E(j) \\
    &= \sum_{t=1}^N \Delta_{at} s_t \mleft[Vz + \tilde{P}\mright]^\top_t E(j) \\
    &= \sum_{t=1}^N \Delta_{at} \mleft[ 
    \begin{pmatrix}
    s_1 & 0 & \cdots \\
    \cdots & \cdots & \cdots \\
    \cdots & 0 & s_N 
\end{pmatrix} \mleft[Vz + \tilde{P}\mright]^\top \mright]_t E(j) \\
&= \Delta_a \begin{pmatrix}
    s_1 & 0 & \cdots \\
    \cdots & \cdots & \cdots \\
    \cdots & 0 & s_N 
\end{pmatrix} \mleft[Vz + \tilde{P}\mright]^\top E(j),
\end{align*}
where $\Delta_a$ is the $a$-th column of $\Delta$ defined in Eq.~\eqref{eq:def_delta}. Using Eq.~\eqref{eq:sigma_z} and Eq.~\eqref{eq:def_s_t}, we notice that
\[
A_z = \mathrm{diag}(\operatorname{softmax}(\frac{z^\top q}{\sqrt{d}})) =  \mathrm{diag}((s_t)_{t=1}^N) = 
\begin{pmatrix}
    s_1 & 0 & \cdots \\
    \cdots & \cdots & \cdots \\
    \cdots & 0 & s_N 
\end{pmatrix} \in \R^{N \times N}.
\]
Using Eq.~\eqref{eq:phi_xi} and Eq.~\eqref{eq:def_notations}, we notice that
\[
\Sigma_{\bar{\xi}} = \mathrm{diag}(\sigma'(W\bar{\xi}) = \mathrm{diag}((\sigma'(w_i^\top\bar{\xi})_{i=1}^h) = \mathrm{diag}((\alpha_i)_{i=1}^h) = \begin{pmatrix}
    \alpha_1 & 0 & \cdots \\
    \cdots & \cdots & \cdots \\
    \cdots & 0 & \alpha_h 
\end{pmatrix} \in \R^{N \times N}.
\]

Using Eq.~\eqref{eq:phi_xi} and Eq.~\eqref{eq:def_P}, we thus have
\[
\tilde{P} = U \begin{pmatrix}
    \alpha_1 & 0 & \cdots \\
    \cdots & \cdots & \cdots \\
    \cdots & 0 & \alpha_h 
\end{pmatrix} W\tilde{M} = U \Sigma_{\bar{\xi}} M \tilde{W}
\]
and we obtain 
\begin{align*}
Vz + \tilde{P} &= Vz + U \Sigma_{\bar{\xi}} M \tilde{W} \\
&= Vz + U \Sigma_{\bar{\xi}} W M^\top Vz \tag{using Eq.~\eqref{eq:def_M_tilde}} \\
&= Vz + U \Sigma_{\bar{\xi}} W M Vz \tag{$M$ symmetric from Eq.~\eqref{eq:def_notations}} \\
&= \mleft( I_d + U \Sigma_{\bar{\xi}} W M\mright) Vz.
\end{align*}

It follows that
\[
\frac{\partial \zeta_j }{\partial q_a} = \Delta_a A_z \mleft[ \mleft( I_d + U \Sigma_{\bar{\xi}} W M\mright) Vz \mright]^\top E(j).
\]

It leads to $\nabla_q \zeta_j = \mleft( \frac{\partial \zeta_j }{\partial q_a}\mright)_a \in \R^d$ that writes
\[
\nabla_q \zeta_j = \Delta_zA_z \mleft[ \mleft( I_d + U \Sigma_{\bar{\xi}} W M\mright) Vz \mright]^\top E(j).
\]
The corresponding entries of $\nabla_\theta \zeta_j$ are the entries of $\nabla_q \zeta_j$.
\end{proof}

\paragraph{Attention value.} The following lemma provides the formula of $\nabla_V \xi_j$.
\begin{boxlem}[Value]
    \label{lem:grad_xi_value}
    Let $\Sigma_{\bar{\xi}} = \mathrm{diag}(\sigma'(W \bar{\xi})) \in \R^{h \times h}$ following Eq.~\eqref{eq:phi_xi}, $M = \frac{1}{\lVert \xi \rVert} \mleft( I_d - \bar{\xi} \bar{\xi}^\top\mright) \in \R^{d \times d}$ following Eq.~\eqref{eq:def_notations} and $Q = U \Sigma_{\bar{\xi}} W M \in \R^{d \times d}$. The entries of $\nabla_\theta \zeta_j$ associated to the value $V$ are the entries of $\nabla_U \zeta_j \in \R^{d \times d}$ flattened in $\R^{d^2}$ where we have
    \[
    \nabla_V \zeta_j = \mleft( I_d + Q^\top \mright) E(j) z \operatorname{softmax}(\frac{z^\top q}{\sqrt{d}}).
    \]
\end{boxlem}
\begin{proof}
We want to compute the $\frac{\partial \zeta_j}{\partial \theta_n}$ for the $\theta_n$ that are entries of $V$. Using Eq.~\eqref{eq:zeta_j} we first need to derive $\frac{\partial \psi_l}{\partial \theta_n}$. As $\theta$ encompasses all the learnable weights, we will derive it depending on the value of $n$. Considering $n$ such that $\theta_n = V_{bc}$ for some $b, c \in [d]$ and using Eq.~\eqref{eq:psi_l}, we have 
\begin{equation}
\label{eq:psi_V_bc}
\frac{\partial \psi_l}{\partial \theta_n} = \frac{\partial \psi_l}{\partial V_{bc}} = \frac{\partial \xi_l}{\partial V_{bc}} + \frac{\partial}{\partial V_{bc}}(\sum_{i=1}^h (u_i)_l \sigma(w_i^\top \bar{\xi})) = \underbrace{\frac{\partial \beta_1(V_{bc})}{\partial V_{bc}}}_{\mathrm{LHS}} + \underbrace{\sum_{i=1}^h (u_i)_l \frac{\partial \beta_2^i(V_{bc})}{\partial V_{bc}}}_{\mathrm{RHS}},
\end{equation}
with 
\[
\begin{cases}
\beta_1 \colon& \R \to \R, \quad V_{bc} \mapsto \xi_l \\
\beta_2^i \colon& \R \to \R, \quad V_{bc} \mapsto (\sigma \circ \beta_3^i \circ \beta_4^i \circ \beta_5^i)(V_{bc}) = \sigma(w_i^\top \bar{\xi})\\
    \beta_3^i \colon& \R^d \to \R, \quad \bar{\xi} \mapsto w_i^\top \bar{\xi} \\
    \beta_4^i \colon& \R^d \to \R^d, \quad \xi \mapsto \bar{\xi} = \frac{\xi}{\lVert \xi \rVert} \\
    \beta_5^i \colon& \R \to \R^d, \quad V_{bc} \mapsto \xi. \\
\end{cases}
\]
We proceed in two steps and first compute the LHS of Eq.~\eqref{eq:psi_V_bc}. 

\underline{LHS of Eq.~\eqref{eq:psi_V_bc}:} we first note that
\[
\xi_l = \sum_{t=1}^N (Vz)_{lt} s_t = \sum_{t=1}^N \sum_{c=1}^d V_{lc} z_{ct} s_t = \sum_{c=1}^d V_{lc} \sum_{t=1}^N z_{ct} s_t,
\]
where $s_t = \operatorname{softmax}(\frac{z^\top q}{\sqrt{d}})_t$. We notice that $\xi_l$ depends only on the $(V_{lc})$, hence the gradients with respect to $V_{bc}$ for some $c \in [d]$ and $b \neq l$ equals zero. It follows that

\begin{equation}
\label{eq:psi_l_V_bc}
\frac{\partial \beta_1(V_{bc})}{\partial V_{bc}} = \frac{\partial \xi_l}{\partial V_{bc}} = 
\begin{cases}
    \sum_{t=1}^N z_{ct} s_t & \text{ if } b =l \\
    0 &\text{ otherwise }
\end{cases}.
\end{equation}

\underline{\bfseries RHS of Eq.~\eqref{eq:psi_V_bc}:} a simple application of the chain rule gives
\[
\frac{\partial \beta_2^i(V_{bc})}{\partial V_{bc}} = \underbrace{\nabla \beta_5^i(V_{bc})}_{\in \R^{1 \times d}} \underbrace{\nabla \beta_4^i(\xi)}_{\in \R^{d \times d}} \underbrace{\nabla \beta_3^i(\bar{\xi})}_{\in \R^d} \underbrace{\sigma'(w_i^\top \bar{\xi})}_{\in \R},
\]
where we recall that $\bar{\xi} = \xi / \lVert \xi \rVert$. We have, using similar arguments as the computation of $\frac{\partial \xi_l}{\partial V_{bc}}$ in Eq.~\eqref{eq:psi_l_V_bc} that 
\[
\nabla \beta_5^i(V_{bc}) = \frac{\partial \xi}{\partial V_{bc}} = \mleft[\frac{\partial \xi_1}{\partial V_{bc}}, \ldots, \frac{\partial \xi_d}{\partial V_{bc}}\mright] = \mleft[ 0, \ldots, 0, \sum_{t=1}^N z_{ct} s_t, 0, \ldots, 0\mright],
\]
where the only non-zero entry is in position $b$ of the row-vector $\nabla \beta_5^i(V_{bc}) \in \R^{1 \times d}$. We notice that the functions $\beta_3^i$ and $\beta_4^i$ of Eq.~\eqref{eq:psi_V_bc} corresponds to the functions $f_3^i$ and $f_4^i$ of Eq.~\eqref{eq:psi_q_a}. Hence, the gradients are the same and we have
\[
\nabla \beta_3^i(\bar{\xi}) = \nabla f_3^i(\bar{\xi}) = w_i
\]
and 
\[
\nabla \beta_4^i(\bar{\xi}) = \nabla f_4^i(\bar{\xi}) = \frac{1}{\lVert \xi \rVert} \mleft( I_d - \bar{\xi}\bar{\xi}^\top\mright).
\]

Going back to $\frac{\partial \beta_2^i(V_{bc})}{\partial V_{bc}}$, we have
\[
\frac{\partial \beta_2^i(V_{bc})}{\partial V_{bc}} = \mleft[ 0, \ldots, 0, \sum_{t=1}^N z_{ct} s_t, 0, \ldots, 0\mright] \frac{1}{\lVert \xi \rVert} \mleft( I_d - \bar{\xi} \bar{\xi}^\top\mright) w_i \sigma'(w_i^\top \bar{\xi}) = \mu M w_i \alpha_i,
\]
where we introduce
\begin{equation}
    \label{eq:def_notations_bis}
    \mu = \mleft[ 0, \ldots, 0, \sum_{t=1}^N z_{ct} s_t, 0, \ldots, 0\mright] \in \R^{1 \times d}, \quad M = \frac{1}{\lVert \xi \rVert} \mleft( I_d - \bar{\xi} \bar{\xi}^\top\mright) \in \R^{d \times d}, \quad \alpha_i = \sigma'(w_i^\top \bar{\xi}),
\end{equation} 
to ease notations, with $M, \alpha_i$ first defined in Eq.~\eqref{eq:def_notations}. Developing the matrix multiplication, we obtain 
\[
\mu M w_i \alpha_i = \alpha_i \sum_{t=1}^N z_{ct} s_t \mleft[Mw_i\mright]_b = \alpha_i \sum_{t=1}^N z_{ct} s_t \sum_{o=1}^d M_{bo} (w_i)_o.
\]
It follows that
\[
\sum_{i=1}^h (u_i)_l \frac{\partial \beta_2^i(V_{bc})}{\partial V_{bc}} = \sum_{i=1}^h (u_i)_l \alpha_i \sum_{t=1}^N z_{ct} s_t \sum_{o=1}^d M_{bo} (w_i)_o =  \sum_{t=1}^N z_{ct} s_t \sum_{i=1}^h \alpha_i (u_i)_l \sum_{o=1}^d M_{bo} (w_i)_o.
\]
We can now put everything together to rewrite Eq.~\eqref{eq:psi_V_bc}.

\underline{LHS + RHS of Eq.~\eqref{eq:psi_V_bc}:} We have
\[
\frac{\partial \psi_l }{\partial V_{bc}} = \frac{\partial \xi_l}{\partial V_{bc} } + \sum_{t=1}^N z_{ct} s_t \sum_{i=1}^h \alpha_i (u_i)_l \sum_{o=1}^d M_{bo} (w_i)_o
\]
and Eq.~\eqref{eq:zeta_j} leads to
\begin{equation}
\label{eq:zeta_V_bc_comp}
\frac{\partial \zeta_j }{\partial V_{bc}} = \underbrace{\sum_{l=1}^p (W_U)_{jl} \frac{\partial \xi_l}{\partial V_{bc}}}_{=\mathrm{LHS}} + \underbrace{\sum_{l=1}^p (W_U)_{jl}\sum_{t=1}^N z_{ct} s_t \sum_{i=1}^h \alpha_i (u_i)_l \sum_{o=1}^d M_{bo} (w_i)_o}_{=\mathrm{RHS}}.
\end{equation}
Using Eq.~\eqref{eq:psi_l_V_bc}, the LHS can be rewritten as
\[
\mathrm{LHS} = (W_U)_{jb} \sum_{t=1}^N z_{ct}s_t = E(j)_b \sum_{t=1}^N z_{ct}s_t.
\]
The RHS can be rewritten as
\begin{align*}
    \mathrm{RHS} &= \sum_{l=1}^p (W_U)_{jl}\sum_{t=1}^N z_{ct} s_t \sum_{i=1}^h \alpha_i (u_i)_l \sum_{o=1}^d M_{bo} (w_i)_o \\
    &= \sum_{t=1}^N z_{ct} s_t \sum_{l=1}^p (W_U)_{jl} \mleft[\sum_{i=1}^h \alpha_i (u_i)_l \sum_{o=1}^d M_{bo} (w_i)_o \mright] \\
    &= \sum_{t=1}^N z_{ct} s_t \sum_{l=1}^p (W_U)_{jl} \mleft[\sum_{i=1}^h \alpha_i (u_i)_l \sum_{o=1}^dW_{io}  M_{bo} \mright] \tag{as $W$ has rows $w_i^\top$}\\
    &= \sum_{t=1}^N z_{ct} s_t \sum_{l=1}^p (W_U)_{jl} \mleft[\sum_{i=1}^h \alpha_i (u_i)_l (WM)_{ib} \mright].
\end{align*}
We now introduce the matrix 
\begin{equation}
    \label{eq:def_P_bis}
    Q = U \begin{pmatrix}
    \alpha_1 & 0 & \cdots \\
    \cdots & \cdots & \cdots \\
    \cdots & 0 & \alpha_h 
\end{pmatrix} W M \in \R^{d \times d}.
\end{equation}
Its entry in row $l$ and column $b$ writes 
\begin{align*}
Q_{lt} &= \mleft[U \begin{pmatrix}
    \alpha_1 & 0 & \cdots \\
    \cdots & \cdots & \cdots \\
    \cdots & 0 & \alpha_h 
\end{pmatrix} W M\mright]_{lb} \\
&= \sum_{i=1}^h \mleft[ U  \begin{pmatrix}
    \alpha_1 & 0 & \cdots \\
    \cdots & \cdots & \cdots \\
    \cdots & 0 & \alpha_h 
\end{pmatrix} \mright]_{li} (WM)_{ib} \\
&= \sum_{i=1}^h \alpha_i U_{li} (WM)_{ib} \\
&= \sum_{i=1}^h \alpha_i (u_i)_l (WM)_{ib} \tag{as $W$ has rows $w_i$}^\top.
\end{align*}
Plugging this term in the computation of the RHS leads to
\begin{align*}
\mathrm{RHS} &= \sum_{t=1}^N z_{ct} s_t \sum_{l=1}^p (W_U)_{jl} Q_{lb} \\
&= \sum_{t=1}^N z_{ct} s_t \sum_{l=1}^p E(j)_l Q_{lb} \\
&= \sum_{t=1}^N z_{ct} s_t \sum_{l=1}^p Q^\top_{bl} E(j)_l \\
&= \sum_{t=1}^N z_{ct} s_t \mleft[ Q^\top E(j)\mright]_b\\
\end{align*}
Going back to Eq.~\eqref{eq:zeta_V_bc_comp}, we have
\begin{align*}
    \frac{\partial \zeta_j }{\partial V_{bc}} &= E(j)_b \sum_{t=1}^N z_{ct}s_t + \sum_{t=1}^N z_{ct} s_t \mleft[ Q^\top E(j)\mright]_b \\
    &= \mleft[ E(j) + Q^\top E(j)\mright]_b \sum_{t=1}^N z_{ct} s_t \\
    &= \mleft[\mleft( I_d + Q^\top \mright) E(j)\mright]_b \sum_{t=1}^N z_{ct} s_t.
\end{align*}
We introduce the vector $z \operatorname{softmax}(\frac{z^\top q}{\sqrt{d}})\in \R^d$ whose entries writes
\[
\mleft[ z \operatorname{softmax}(\frac{z^\top q}{\sqrt{d}}) \mright]_c = \sum_{t=1}^N z_{ct}\operatorname{softmax}(\frac{z^\top q}{\sqrt{d}})_t = \sum_{t=1}^N z_{ct} s_t,
\]
where we used the fact that $s_t = \operatorname{softmax}(\frac{z^\top q}{\sqrt{d}})_t$ following Eq.~\eqref{eq:def_s_t}. It follows that
\[
\frac{\partial \zeta_j }{\partial V_{bc}} = \mleft[\mleft( I_d + Q^\top \mright) E(j)\mright]_b\mleft[ z \operatorname{softmax}(\frac{z^\top q}{\sqrt{d}}) \mright]_c.
\]
Using Eq.~\eqref{eq:phi_xi} and Eq~\eqref{eq:def_P_bis}, we have 
\[
Q = U \begin{pmatrix}
    \alpha_1 & 0 & \cdots \\
    \cdots & \cdots & \cdots \\
    \cdots & 0 & \alpha_h 
\end{pmatrix} W M =  U \Sigma_{\bar{\xi}} WM.
\]
Hence, $\nabla_V \zeta_j = \mleft( \frac{\partial \zeta_j }{\partial V_{bc}}\mright)_{b, c} \in \R^{d \times d}$ is a matrix equal to
\[
\nabla_V \zeta_j = \mleft( I_d + \mleft[ U \Sigma_{\bar{\xi}} WM \mright]^\top \mright) E(j) \mleft[z \operatorname{softmax}(\frac{z^\top q}{\sqrt{d}})\mright]^\top.
\]
The corresponding entries of $\nabla_\theta \zeta_j$ amounts to the entries of $\nabla_V \zeta_j$ flattened in $\R^{d^2}$. This concludes the proof.
\end{proof}

\paragraph{Feed-forward receptors.} The following lemma provides the formula of $\nabla_W \xi_j$.
\begin{boxlem}[Receptors]
    \label{lem:grad_xi_receptors}
Let $\Sigma_{\bar{\xi}} = \mathrm{diag}(\sigma'(W \bar{\xi})) \in \R^{h \times h}$ following Eq.~\eqref{eq:phi_xi}. The entries of $\nabla_\theta \zeta_j$ associated to the assemblers $(w_i)_{i=1}^h$ are the entries of $\nabla_W \zeta_j \in \R^{h times d}$ flattened in $\R^{dh}$ where we have
    \begin{equation*}
    \nabla_W \zeta_j = \Sigma_{\bar{\xi}} U^\top E(j) \bar{\xi}^\top.
    \end{equation*}
\end{boxlem}
\begin{proof}
We want to compute the $\frac{\partial \zeta_j}{\partial \theta_n}$ for the $\theta_n$ that are entries of $W$. Using Eq.~\eqref{eq:zeta_j} we first need to derive $\frac{\partial \psi_l}{\partial \theta_n}$. As $\theta$ encompasses all the learnable weights, we will derive it depending on the value of $n$. Here, $\psi_l$ depends on all the $(w_i)_e$. Considering $n$ such that $\theta_n = (w_i)_e$ for some $i \in [h]$ and some $e \in [d]$ and using Eq.~\eqref{eq:psi_l}, we have
\[
\frac{\partial \psi_l}{\partial \theta_n} = \frac{\partial \psi_l}{\partial (w_i)_e} = (u_i)_l \frac{\partial \sigma(w_i^\top \bar{\xi})}{\partial (w_i)_e}.
\]
Noting that $w_i^\top \xi = \sum_{f=1}^d (w_i)_f \xi_f$ and using the chain rule, we have
\begin{align*}
\frac{\partial \sigma(w_i^\top \xi / \lVert \xi \rVert)}{\partial (w_i)_e} &= \frac{\partial (w_i^\top \bar{\xi})}{\partial (w_i)_e} \sigma'(w_i^\top \bar{\xi}) \\
&= \frac{\partial(\sum_{f=1}^d (w_i)_f \bar{\xi}_f)}{\partial (w_i)_e}\sigma'(w_i^\top \bar{\xi}) \\
&= \frac{\partial((w_i)_e \bar{\xi}_e)}{\partial (w_i)_e}\sigma'(w_i^\top \bar{\xi}) \\
&= \bar{\xi}_e\sigma'(w_i^\top \bar{\xi}).
\end{align*}
We recall that $\sigma$ is the GeLU activation defined as $\sigma \colon x \to x \phi(x)$ with $\phi$ the cumulative distribution function of the standard normal distribution. By the fundamental theorem of analysis, it leads to
\[
\sigma'(x) = \phi(x) + x\phi'(x) = \phi(x) + xf(x),
\] 
with $f \colon u \to \frac{1}{\sqrt{2\pi}}\exp{(-\frac{u^2}{2})}$ the probability density function of the standard normal distribution. Putting everything together, we have
\[
\frac{\partial \psi_l}{\partial (w_i)_e} = (u_i)_l \bar{\xi}_e \sigma'(w_i^\top \bar{\xi}).
\]

Using Eq.~\eqref{eq:zeta_j}, we have
\begin{align*}
\frac{\partial \zeta_j}{\partial (w_i)_e} &= \sum_{l=1}^p (W_U)_{jl} (u_i)_l \bar{\xi}_e \sigma'(w_i^\top \bar{\xi}) \\
&= \sum_{l=1}^p E(j)_l (u_i)_l \bar{\xi}_e \sigma'(w_i^\top \bar{\xi}) \\
&= E(j)^\top u_i \bar{\xi}_e \sigma'(w_i^\top \xi / \lVert \xi \rVert) \\
&= \sigma'(w_i^\top \bar{\xi}) u_i^\top E(j) \bar{\xi}_e.
\end{align*}
To ease the notations, we denote $\alpha_i = \sigma'(w_i^\top \bar{\xi})$. Recalling that $\sigma$ and thus $\sigma'$ is entry-wise, it leads to $\nabla_W \zeta_j = (\frac{\partial \zeta_j}{\partial (w_i)_e})_{i, e} \in \R^{h \times d}$ that writes
\begin{align*}
\nabla_W \zeta_j &=
\begin{pmatrix}
\alpha_1 u_1^\top E(j)\\
\cdots \\
\alpha_h u_h^\top E(j)\\
\end{pmatrix} \bar{\xi}^\top\\
&= \begin{pmatrix}
\alpha_1 u_1^\top\\
\cdots \\
\alpha_h u_h^\top
\end{pmatrix} E(j) \bar{\xi}^\top \\
&= \begin{pmatrix}
\alpha_1 & 0 & \cdots \\
\cdots & \cdots & \cdots  \\
\cdots & 0 & \alpha_h
\end{pmatrix} U^\top E(j) \bar{\xi}^\top \\
&= \begin{pmatrix}
\sigma'(w_1^\top \bar{\xi}) & 0 & \cdots \\
\cdots & \cdots & \cdots  \\
\cdots & 0 & \sigma'(w_h^\top \bar{\xi})
\end{pmatrix} U^\top E(j) \bar{\xi}^\top \\
&= \mathrm{diag}(\sigma'(W\bar{\xi})) U^\top E(j) \bar{\xi}^\top,
\end{align*}
where $\mathrm{diag}(\sigma'(W\bar{\xi})) \in \R^{h \times h}$  is diagonal with entries $(\sigma'(w_i^\top \bar{\xi}))_{i=1}^h$. Using Eq.~\eqref{eq:phi_xi}, it finally leads to
\[
\nabla_W \zeta_j = \Sigma_{\bar{\xi}} U^\top E(j) \bar{\xi}^\top,
\]
and the corresponding entries of $\nabla_\theta \zeta_j$ amounts to the entries of $\nabla_W \zeta_j$ flattened in $\R^{hd}$. This concludes the proof.
\end{proof}

\paragraph{Feed-forward assemblers.} The following lemma provides the formula of $\nabla_U \xi_j$.
\begin{boxlem}[Assemblers]
    \label{lem:grad_xi_assemblers}
The entries of $\nabla_\theta \zeta_j$ associated to the assemblers $(u_i)_{i=1}^h$ are the entries of $\nabla_U \zeta_j \in \R^{d \times h}$ flattened in $\R^{dh}$ where we have
    \begin{equation*}
    \nabla_U \zeta_j = E(j)\sigma(W\bar{\xi})^\top \in  R^{d \times h}.
    \end{equation*}
\end{boxlem}
\begin{proof}
We want to compute the $\frac{\partial \zeta_j}{\partial \theta_n}$ for the $\theta_n$ that are entries of $U$. Using Eq.~\eqref{eq:zeta_j} we first need to derive $\frac{\partial \psi_l}{\partial \theta_n}$. As $\theta$ encompasses all the learnable weights, we will derive it depending on the value of $n$. As $\psi_l$ only depends on the $l$-elements of the $(u_i)$, we only consider indices $n$ such that $\theta_n = (u_i)_l$ for some $i \in [h]$ as the other values of $n$ will lead to a null gradient. Using Eq.~\eqref{eq:psi_l}, we have
\[
\frac{\partial \psi_l}{\partial \theta_n} = \frac{\partial \psi_l}{\partial (u_i)_l} = \frac{\partial (u_i)_l}{\partial (u_i)_l} \sigma(w_i^\top \bar{\xi}) = \sigma(w_i^\top \bar{\xi}).
\]
Using Eq.~\eqref{eq:zeta_j}, we have
\[
\frac{\partial \zeta_j}{\partial (u_i)_l} = (W_U)_{jl} \sigma(w_i^\top \bar{\xi}) = E(j)_l \sigma(w_i^\top \bar{\xi}).
\]
Hence, recalling that $\sigma$ is applied element-wise, $\nabla_U \zeta_j = (\frac{\partial \zeta_j}{\partial (u_i)_{l}})_{l, i} \in \R^{d \times h}$ is a matrix equal to
\[
\nabla_U \zeta_j = E(j)\sigma(W\bar{\xi})^\top,
\]
and the corresponding entries of $\nabla_\theta \zeta_j$ amounts to the entries of $\nabla_U \zeta_j$ flattened in $\R^{dh}$. This concludes the proof.
\end{proof}
We can now put everything together. Using \cref{lem:grad_compact_formulation} in combination with the gradientsderived in \cref{lem:grad_xi_query}, \cref{lem:grad_xi_value}, \cref{lem:grad_xi_receptors} and \cref{lem:grad_xi_assemblers} concludes the proof of \cref{lem:grad_loss}.
\end{proof}

\subsection{Proof of \cref{prop:learning_clustering_heads_informal}}
\label{app:proof_stationary_points}
We detail below the proof of \cref{prop:stationary_points}.
\begin{proof}
We assume the gradient descent continues until convergence at a critical point and discard other practical considerations, such as early stopping. Following~\citet{chi2019nonconvex} and using \cref{def:stationary_points}, we know that the parameters $\theta$ of the transformer verify
\begin{equation}
\label{eq:euler_equation}
    \nabla_{\theta} \mathcal{L}(\theta) = 0
\end{equation}
From \cref{lem:grad_loss}, we know that $\nabla_{\theta} \mathcal{L}(\theta) \in \R^m$ writes 
\begin{equation}
\label{eq:grad_decomposition}
    \nabla_{\theta} \mathcal{L}(\theta) = [\nabla_{q} \mathcal{L}(\theta), \nabla_{V} \mathcal{L}(\theta), \nabla_{W} \mathcal{L}(\theta), \nabla_{U} \mathcal{L}(\theta)],
\end{equation}
using $[\cdot]$ as the vector concatenation and flattening matrices whenever needed. We now reformulate the gradients of each weight group. We have
\begin{align*}
    \nabla_{q} \mathcal{L}(\theta) &=\sum_{j=1}^p \mathbb{E}_x \mleft[ (\hat{p}_{\theta}(y=j|x) - p(y=j|x))\cdot \nabla_q \zeta_j \mright] \\
    &= \sum_{j=1}^p \mathbb{E}_x \mleft[ (\hat{p}_{\theta}(y=j|x) - p(y=j|x))\cdot \mleft(\Delta_zA_z (Vz)^\top \mleft( I_d + Q^\top\mright) E(j)\mright) \mright] \\
    &= \mathbb{E}_x \mleft[ \sum_{j=1}^p 
 (\hat{p}_{\theta}(y=j|x) - p(y=j|x))\cdot \mleft(\Delta_zA_z (Vz)^\top \mleft( I_d + Q^\top\mright) E(j)\mright) \mright] \tag{by linearity of the expectation} \\
    &= \mathbb{E}_x \mleft[\Delta_zA_z (Vz)^\top \mleft( I_d + Q^\top\mright)  \cdot  \sum_{j=1}^p 
 (\hat{p}_{\theta}(y=j|x) - p(y=j|x))E(j) \mright] \tag{$\hat{p}_{\theta}(y=j|x) - p(y=j|x) \in \R$} \\
 &= \mathbb{E}_x\mleft[\Delta_zA_z (Vz)^\top \mleft( I_d + Q^\top\mright) \mathscr{C}_x\mright]. \tag{by definition of $\mathscr{C}_x$}
\end{align*}

The same arguments lead to
\begin{align*}
    \nabla_{V} \mathcal{L}(\theta) &=\sum_{j=1}^p \mathbb{E}_x \mleft[ (\hat{p}_{\theta}(y=j|x) - p(y=j|x))\cdot \nabla_V \zeta_j \mright]  \\
    &= \sum_{j=1}^p \mathbb{E}_x \mleft[ (\hat{p}_{\theta}(y=j|x) - p(y=j|x))\cdot \mleft(\mleft( I_d + Q^\top \mright) E(j) \mleft[z \operatorname{softmax}(\frac{z^\top q}{\sqrt{d}})\mright]^\top\mright) \mright] \\
    &= \mathbb{E}_x \mleft[ \sum_{j=1}^p 
 (\hat{p}_{\theta}(y=j|x) - p(y=j|x))\cdot \mleft(\mleft( I_d + Q^\top \mright) E(j) \mleft[z \operatorname{softmax}(\frac{z^\top q}{\sqrt{d}})\mright]^\top\mright) \mright]   \\
    &= \mathbb{E}_x \mleft[\mleft( I_d + Q^\top\mright)  \cdot \mleft(\sum_{j=1}^p 
 (\hat{p}_{\theta}(y=j|x) - p(y=j|x))E(j)\mright) \cdot \mleft[z \operatorname{softmax}(\frac{z^\top q}{\sqrt{d}})\mright]^\top\mright] \\
 &= \mathbb{E}_x \mleft[\mleft( I_d + Q^\top\mright)  \mathscr{C}_x \mleft[z \operatorname{softmax}(\frac{z^\top q}{\sqrt{d}})\mright]^\top\mright],
\end{align*}
then to
\begin{align*}
    \nabla_{W} \mathcal{L}(\theta)&=\sum_{j=1}^p \mathbb{E}_x \mleft[ (\hat{p}_{\theta}(y=j|x) - p(y=j|x))\cdot \nabla_W \zeta_j \mright]  \\
    &= \sum_{j=1}^p \mathbb{E}_x \mleft[ (\hat{p}_{\theta}(y=j|x) - p(y=j|x))\cdot \mleft(\Sigma_{\bar{\xi}} U^\top E(j) \bar{\xi}^\top\mright) \mright]  \\
    &=\mathbb{E}_x \mleft[ \sum_{j=1}^p 
 (\hat{p}_{\theta}(y=j|x) - p(y=j|x))\cdot \mleft(\Sigma_{\bar{\xi}} U^\top E(j) \bar{\xi}^\top\mright) \mright]   \\
    &=\mathbb{E}_x \mleft[\Sigma_{\bar{\xi}} U^\top  \cdot \mleft(\sum_{j=1}^p 
 (\hat{p}_{\theta}(y=j|x) - p(y=j|x))E(j)\mright) \cdot \bar{\xi}^\top\mright] \\
 &= \mathbb{E}_x \mleft[\Sigma_{\bar{\xi}} U^\top \mathscr{C}_x \bar{\xi}^\top\mright] ,
\end{align*}
and finally to
\begin{align*}
    \nabla_{U} \mathcal{L}(\theta)&=\sum_{j=1}^p \mathbb{E}_x \mleft[ (\hat{p}_{\theta}(y=j|x) - p(y=j|x))\cdot \nabla_U \zeta_j \mright]  \\
    &= \sum_{j=1}^p \mathbb{E}_x \mleft[ (\hat{p}_{\theta}(y=j|x) - p(y=j|x))\cdot \mleft(E(j)\sigma(W\bar{\xi})^\top\mright) \mright] \\
    &= \mathbb{E}_x \mleft[ \sum_{j=1}^p 
 (\hat{p}_{\theta}(y=j|x) - p(y=j|x))\cdot \mleft(E(j)\sigma(W\bar{\xi})^\top\mright) \mright] \\
    &= \mathbb{E}_x \mleft[\mleft(\sum_{j=1}^p 
 (\hat{p}_{\theta}(y=j|x) - p(y=j|x))E(j)\mright) \cdot \sigma(W\bar{\xi})^\top\mright]  \\
 &= \mathbb{E}_x \mleft[\mathscr{C}_x \sigma(W\bar{\xi})^\top\mright].
\end{align*}

The equality to $0$ of the vector in $\R^m$ is equivalent to each of its components being equal to $0$, and in particular, this is the case of the gradients with respect to $q, V, W$, and $U$. Combining \cref{eq:euler_equation} with \cref{lem:grad_loss} leads to the following stationarity conditions
\begin{equation*}
\begin{cases}
    &\mathbb{E}_x\mleft[\Delta_zA_z (Vz)^\top \mleft( I_d + Q^\top\mright) \mathscr{C}_x\mright] = 0 \\
    &\mathbb{E}_x\mleft[\mleft( I_d + Q^\top \mright) \mathscr{C}_x \mleft[z \operatorname{softmax}(\frac{z^\top q}{\sqrt{d}})\mright]^\top\mright] =0\\
    &\mathbb{E}_x\mleft[ \Sigma_{\bar{\xi}} U^\top \mathscr{C}_x \bar{\xi}^\top \mright] =0\\
    &\mathbb{E}_x\mleft[ \mathscr{C}_x\sigma(W\bar{\xi})^\top\mright]=0.
    \end{cases}
\end{equation*}
Once those conditions are established, the rest of the proof is straightforward. Indeed, it suffices to recall that the clustering heads described in \cref{sec:solving_task} capture all the invariants of the problem. This leads to a perfect separation of the clusters that can be seen in \cref{fig:circuit_ideal}. In particular, the transformer correctly classifies all the sequences, which amounts to the $\hat{p}_{\theta}(y=j|x)$ and the $p(y=j|x)$ being aligned. By definition, this implies $\mathscr{C}_x = 0$ for all $x$, and noticing the linear dependency of the gradients with $\mathscr{C}_x$ ensures that the stationary conditions previously stated are verified. Hence, a transformer that implements a circuit head has parameters $\theta$ for which $\nabla \mathcal{L}(\theta) = 0$ holds and that can thus be reached via gradient descent. This implies circuit heads can emerge during the training of the transformer. This concludes the proof. 
\end{proof}

\subsection{Proof of \cref{prop:informal_gradient_bounds}}
\label{app:proof_grad_bounds}
We detail below the proof of \cref{prop:informal_gradient_bounds} whose formal statement is given in \cref{prop:gradients_bound}.
\begin{proof}
We first decompose the gradient of the loss using \cref{eq:grad_decomposition}. Gathering the terms by group leads to
\begin{equation}
\label{eq:norm_decomp}
\lVert \nabla_{\theta}\mathcal{L}(\theta) \rVert^2 = \sum_{n=1}^m \mleft|\frac{\partial \mathcal{L}(\theta)}{\partial \theta_n}\mright|^2 = \lVert \nabla_{q}\mathcal{L}(\theta) \rVert^2 + \lVert \nabla_{V}\mathcal{L}(\theta) \rVert^2 + \lVert \nabla_{W}\mathcal{L}(\theta) \rVert^2 + \lVert \nabla_{U}\mathcal{L}(\theta) \rVert^2,
\end{equation}
where the norm is the Euclidean one with matrices flattened as vectors. Similarly to the proof of \cref{prop:stationary_points}, we have
\[
\begin{cases}
    &\nabla_{q}\mathcal{L}(\theta)=\mathbb{E}_x\mleft[\Delta_zA_z (Vz)^\top \mleft( I_d + Q^\top\mright) \mathscr{C}_x\mright] \\
    &\nabla_{V}\mathcal{L}(\theta)=\mathbb{E}_x\mleft[\mleft( I_d + Q^\top \mright) \mathscr{C}_x \mleft[z \operatorname{softmax}(\frac{z^\top q}{\sqrt{d}})\mright]^\top\mright] \\
    &\nabla_{W}\mathcal{L}(\theta)=\mathbb{E}_x\mleft[ \Sigma_{\bar{\xi}} U^\top \mathscr{C}_x \bar{\xi}^\top \mright] \\
    &\nabla_{U}\mathcal{L}(\theta)=\mathbb{E}_x\mleft[ \mathscr{C}_x\sigma(W\bar{\xi})^\top\mright].
    \end{cases}
\]
We now proceed with upper-bounding each term of \cref{eq:norm_decomp}. For clarity of the proofs, we first recall the following basic relations. 
\begin{boxlem}[Properties]
\label{lem:properties}
Let $a \in \R^m, b \in \R^n, A \in \R^{n \times m}$ and $B \in \R^{m\times l}$. We have
\begin{enumerate}[leftmargin=*]
    \item $\lVert ab^\top \rVert = \lVert a \rVert \cdot \lVert b \rVert$.
    \item $\lVert Ab \rVert \leq \lVert A \rVert_{\mathrm{op}} \lVert b \rVert$.
    \item The operator norm (resp. the Frobenius norm) is sub-multiplicative, i.e., $\lVert AB \rVert_{\mathrm{op}} \leq \lVert A \rVert_{\mathrm{op}} \lVert B \rVert_{\mathrm{op}}$.
    \item If $A$ has elements bounded by $C > 0$, then $\lVert A \rVert_{\mathrm{op}} \leq C \sqrt{nm}$.
    \item If $P$ be an orthogonal projection, then, we have $\lVert P \rVert_\mathrm{op} = 1$.
    \item The singular values of $A$ and $A^\top$ coincides. In particular, $\lVert A \rVert_{\mathrm{op}} = \lVert A^\top \rVert_{\mathrm{op}}$.
\end{enumerate}
\end{boxlem}
\begin{proof}
The first point follows from
\[
\lVert ab^\top \rVert^2 = \sum_{i,j} (ab^\top)_{ij}^2 = \sum_{i,j} (a_ib_j)^2 = \mleft(\sum_{i}a_i^2\mright) \cdot \mleft(\sum_{j}b_j^2\mright) = \lVert a \rVert^2 \cdot \lVert b \rVert^2,
\]
where taking the square root gives the desired equality. 

The second point comes from the definition of the operator norm
\[
\lVert A \rVert_{\mathrm{op}} = \sup \{ \frac{\lVert Ax \rVert}{\lVert x \rVert} : x \in \R^m, x \neq 0\} = \sup \{ \lVert Ax \rVert: x \in \R^m, \lVert x \rVert =1\}.
\]
It implies that for any $x \in \R^m$ non-zero, by definition of the sup, we have 
\[
\lVert Ax \rVert = \frac{\lVert Ax \rVert}{\lVert x \rVert} \cdot \lVert x \rVert \leq \lVert A \rVert_{\mathrm{op}} \lVert x \rVert.
\]
The inequality holds (with equality) for $x=0$. 

The third point is simply a property of matrix norms. 

The fourth point comes from the definition of operator norm. Let $x \in \R^m$ with unit norm. We have 
\begin{equation}
\label{eq:ineq_op_intermediary}
\lVert Ax \rVert^2 = \sum_{i=1}^n \mleft(\sum_{j=1}^m A_{ij} x_j \mright)^2 \leq \sum_{i=1}^n \mleft(\sum_{j=1}^m A_{ij}^2\mright) \mleft( \sum_{j=1}^m x_j^2\mright) \leq nm \cdot C^2,
\end{equation}
using the fact that $\lVert x \rVert^2 = \sum_{j=1}^m x_j^2 =1$ and the bound on the entries of $A$ for the last inequality. Taking the root square and the sup on the left-hand term gives the desired formula.

For the fifth point, we recall that an orthogonal projection is symmetric. Hence, we have $P = P^2 = p^\top P$, which also ensures $P$ to be semi-definite positive, i.e., with nonnegative eigenvalues. Let $\lambda$ be such an eigenvalue and $v$ an eigenvector. We must have $Pv = \lambda v$ but applying $P$ on this relation also leads to
\[
Pv =P^2v = P(Pv) = P(\lambda v) = \lambda^2 v.
\]
Taking the norm forces $\lambda = \lambda^2$ which implies $\lambda \in \{0, 1\}$. As the singular values of $P$ are its eigenvalues (by the symmetry of $P$), it implies that $\lVert P \rVert_\mathrm{op} \leq 1$. Moreover, writing the singular value decomposition of $P$ leads to $P = \sum_i \sigma_i u_i v_i^\top$. Let $u_k$ be in the range of $P$ (it suffices to show that $Pv_k = \sigma_k u_k$ and since $P \neq 0$, there is at least one $\sigma_k$ non-zero which provides at least one $u_k$ in the range of $P$). Using $P^2 = P$ implies that 
\[
Pu_k = u_k \iff \sum_i \sigma_i u_i v_i^\top u_k = u_k \iff \sum_i (\sigma_i v_i^\top u_k) u_i = u_k.
\]
Since the $(u_i)$ forms an orthogonal basis, it forces $(\sigma_k v_k^\top u_k) = 1$. A simple application of Cauchy-Schwartz (recalling that the $u_i, v_i$ have unit norm) leads to $v_k^\top u_k \leq 1$, and finally it implies that $\sigma_k \geq 1$. This forces by definition of the operator norm $\lVert P \rVert_\mathrm{op} \geq 1$. Putting everything together gives $\lVert P \rVert_\mathrm{op} = 1$. 

For the last point, assume $A$ has a singular value decomposition that writes
\[
A = \sum_{i=1}^{\min\{n, m\}} \sigma_i u_i v_i^\top,
\]
where the $u_i \in \R^n, v_i \in \R^m$ are orthogonal basis. Hence, we have 
\[
A^\top = \sum_{i=1}^{\min\{n, m\}} \sigma_i v_i u_i^\top,
\]
which shows that the singular values are the same. Since the operator norm computes the largest singular value, we have the last equality.
\end{proof}
In the rest, we upper-bound all the terms by extensively using the monotonicity of the expectation.
\paragraph{Query term.} We have
\begin{align*}
    \lVert \nabla_{q}\mathcal{L}(\theta) \rVert^2 &= \lVert\mathbb{E}_x\mleft[\Delta_zA_z (Vz)^\top \mleft( I_d + Q^\top\mright) \mathscr{C}_x\mright] (\theta) \rVert^2 \\
    &\leq \mathbb{E}_x\mleft[ \lVert\Delta_zA_z (Vz)^\top \mleft( I_d + Q^\top\mright) \mathscr{C}_x \rVert^2 \mright]  \tag{Jensen inequality since $\lVert \cdot \rVert^2$ is convex} \\
    &\leq \mathbb{E}_x\mleft[ \lVert\Delta_zA_z (Vz)^\top \mleft( I_d + Q^\top\mright)\rVert_{\mathrm{op}}^2 \lVert\mathscr{C}_x \rVert^2 \mright]  \tag{Point $(2)$ of \cref{lem:properties}} \\
\end{align*}
Using the fact that operator norms are sub-multiplicative, we have
\[
\lVert\Delta_zA_z (Vz)^\top \mleft( I_d + Q^\top\mright)\rVert_{\mathrm{op}} \leq \lVert\Delta_z\rVert_{\mathrm{op}} \lVert A_z \rVert_{\mathrm{op}} \lVert(Vz)^\top\rVert_{\mathrm{op}} \lVert\mleft( I_d + Q^\top\mright)\rVert_{\mathrm{op}}.
\]
Recalling from \cref{eq:def_delta} that
\[
\Delta_z = \frac{z}{\sqrt{d}} \mleft[ I_N - \operatorname{softmax}(\frac{z^\top q}{\sqrt{d}})\mathbbm{1}_N^\top\mright],
\]
and using the fact that the embeddings $z \in \R^{d \times M}$ have columns in $\mathbb{S}^d$ (thanks to the RMS-norm), we have that $I_N - \operatorname{softmax}(\frac{z^\top q}{\sqrt{d}})\mathbbm{1}_N^\top$ has entries bounded by $2$. Indeed, given that 
\[
\forall t \in [N], \lVert z_{\cdot, t} \rVert^2 = \sum_{i=1}^d z_{dt}^2 = 1,
\]
we know that the $z_{dt}$ are bounded by one. As the softmax outputs a probability vector, the $z_{dt}$ are reweighted by elements in $[0,1]$. This implies that the entries of $\operatorname{softmax}(\frac{z^\top q}{\sqrt{d}})\mathbbm{1}_N^\top$ (a matrix with rows equal to $\operatorname{softmax}(\frac{z^\top q}{\sqrt{d}})$) are bounded by 1. Taking the minus sign into account leads to $I_N - \operatorname{softmax}(\frac{z^\top q}{\sqrt{d}})\mathbbm{1}_N^\top$ with entries bounded by $2$ and using the point $(3)$ and $(4)$ of \cref{lem:properties}, we have
\begin{align*}
\lVert \Delta_z \rVert_\mathrm{op} &\leq \frac{1}{\sqrt{d}} \lVert z \rVert_\mathrm{op} \lVert I_N - \operatorname{softmax}(\frac{z^\top q}{\sqrt{d}})\mathbbm{1}_N^\top \rVert_\mathrm{op} \\
&\leq \frac{1}{\sqrt{d}} \lVert z^\top \rVert_\mathrm{op} \cdot (2N) \tag{last point of \cref{lem:properties}}
\end{align*}
Moreover, using a proof similar to the point $(4)$ of \cref{lem:properties} (in particular involving \cref{eq:ineq_op_intermediary}), one can show that for any $A \in \R^{n \times m}$,
\[
\lVert A \rVert_\mathrm{op} \leq \sqrt{m} \max_{i\in [m]} \lVert A_{i} \rVert.
\]
Applying the previous inequality to $z^\top \in \R^{N \times d}$ that has rows $z_{\cdot, t}$ with unit norm leads to
\begin{equation}
\label{eq:op_z}
\lVert z^\top \rVert_\mathrm{op} \leq \sqrt{d}.
\end{equation}
We thus obtain $\lVert \Delta_z \rVert_\mathrm{op} \leq 2N$. Recalling from \eqref{eq:sigma_z} that
\[
A_z = \mathrm{diag}\mleft(\operatorname{softmax}(\frac{z^\top q}{\sqrt{d}}) \mright)
\]
is diagonal with entries that form a probability vector, we know that 
\[
\lVert A_z \rVert_\mathrm{op} = \sigma_\mathrm{max}(A_z) = \lambda_\mathrm{max}(A_z) = \max_{t \in [N]} \{\operatorname{softmax}(\frac{z^\top q}{\sqrt{d}})_t \}\leq 1. 
\]
Then, using \cref{eq:op_z}, we notice that
\[
\lVert (Vz)^\top \rVert_\mathrm{op} = \lVert z^\top V^\top \rVert_\mathrm{op} \leq \lVert z^\top \rVert_\mathrm{op} \lVert V^\top \rVert_\mathrm{op} \leq \sqrt{d} \lVert V^\top \rVert_\mathrm{op} = \sqrt{d}\lVert V\rVert_\mathrm{op},
\]
where the last equality comes from the last point of \cref{lem:properties}.

We now proceed to $I + Q^\top$. Using the triangular inequality leads to $\lVert I + Q^\top \rVert_\mathrm{op} \leq \lVert I \rVert_\mathrm{op} + \lVert Q^\top \rVert_\mathrm{op}$. We know that $\lVert I \rVert_\mathrm{op} = 1$. Moreover, we have
\[
\lVert Q^\top \rVert_\mathrm{op} = \lVert M(U \Sigma_{\bar{\xi}} W)^\top \rVert_\mathrm{op} \leq \lVert M \rVert_\mathrm{op} \cdot \lVert (U \Sigma_{\bar{\xi}} W)^\top \rVert_\mathrm{op}.
\]
Using the fact that $M$ is an orthogonal projection ensures from the point $(6)$ of \cref{lem:properties} that $\lVert M \rVert_\mathrm{op} = 1$. Moreover, using the points $(3)$ and $(6)$ of \cref{lem:properties} leads to
\[
\lVert (U \Sigma_{\bar{\xi}} W)^\top \rVert_\mathrm{op}= \lVert U \Sigma_{\bar{\xi}} W \rVert_\mathrm{op} \leq \lVert U \rVert_\mathrm{op} \lVert \Sigma_{\bar{\xi}} \rVert_\mathrm{op} \lVert W \rVert_\mathrm{op} \lVert.
\]
Noting that $\Sigma_{\bar{\xi}}$ is diagonal with entries $\sigma'(w_i^\top\bar{\xi})$ (\cref{eq:phi_xi}) and recalling that $\sigma'(x) = \psi(x) + xf(x)$ with $\phi$ the cumulative distribution function of the normal distribution and $f$ its probability density function (in particular, $\phi$ has values in $[0,1]$ and $f$ in $[0, 1/\sqrt{2\pi}]$), we know that $\lvert \sigma'(x) \rvert \leq 1 + (1/\sqrt{2\pi}) \lvert x \rvert$. Since $\sigma'$ is applied entry-wise, using again the point (2) of \cref{lem:properties} leads to 
\[
\lVert \Sigma_{\bar{\xi}} \rVert_\mathrm{op} \leq 1 + \frac{1}{\sqrt{2\pi}}\lVert W\bar{\xi} \rVert \leq 1+\frac{1}{\sqrt{2\pi}} \lVert W\rVert_\mathrm{op}.
\] 
Putting everything together leads to
\[
\lVert\Delta_zA_z (Vz)^\top \mleft( I_d + Q^\top\mright)\rVert_{\mathrm{op}} \leq 2n \sqrt{d} \lVert V \rVert_\mathrm{op} (1 +\sqrt{\frac{2}{\pi}}\lVert W \rVert_\mathrm{op}^2\lVert U \rVert_\mathrm{op}).
\]
Since the weight matrices are randomly initialized and learned by gradient descent whose updates have the form $\mathbb{E}_x \mleft[\phi(\mathrm{parameters})\mright]$, they do not depend on the expectation with respect to the training data distribution. We can thus take those terms outside of the expectation which leads to
\begin{equation}
\label{eq:q_term}
\lVert\nabla_q \mathcal{L}(\theta) \rVert^2  \leq \mleft[2n \sqrt{d} \lVert V \rVert_\mathrm{op} (1 +\sqrt{\frac{2}{\pi}}\lVert W \rVert_\mathrm{op}^2\lVert U \rVert_\mathrm{op})\mright]^2 \cdot \mathbb{E}_x\mleft[ \lVert \mathscr{C}_x \rVert^2\mright].
\end{equation}
\paragraph{Value term.} We have
\begin{align*}
\lVert\nabla_{V}\mathcal{L}(\theta) \rVert^2&\leq \mathbb{E}_x\mleft[\lVert \underbrace{\mleft( I_d + Q^\top \mright) \mathscr{C}_x}_{\in \R^d} \underbrace{\mleft[z \operatorname{softmax}(\frac{z^\top q}{\sqrt{d}})\mright]^\top}_{\in \R^{1 \times d}} \rVert^2\mright] \tag{Jensen inequality}\\
&\leq \mathbb{E}_x\mleft[\lVert\mleft( I_d + Q^\top \mright) \mathscr{C}_x\rVert^2\left\lVert\mleft[z \operatorname{softmax}(\frac{z^\top q}{\sqrt{d}})\mright]^\top \right\rVert^2\mright] \tag{Point (1) of \cref{lem:properties}}
\end{align*}
Again, using point (2) of \cref{lem:properties} and the derivations done for the query term, we have
\[
\lVert\mleft( I_d + Q^\top \mright) \mathscr{C}_x\rVert \leq \lVert\mleft( I_d + Q^\top \mright)\rVert_{\mathrm{op}} \cdot \lVert\mathscr{C}_x\rVert \leq (1 +2\lVert W \rVert_\mathrm{op}\lVert U \rVert_\mathrm{op}) \cdot \lVert\mathscr{C}_x\rVert.
\]
Similarly, we have
\[
\left\lVert\mleft[z \operatorname{softmax}(\frac{z^\top q}{\sqrt{d}})\mright]^\top \right\rVert = \left\lVert z \operatorname{softmax}(\frac{z^\top q}{\sqrt{d}}) \right\rVert \leq \lVert z \rVert_\mathrm{op} \lVert \operatorname{softmax}(\frac{z^\top q}{\sqrt{d}}) \rVert.
\]
As shown in the proof of the query term, we have $\lVert z \rVert_\mathrm{op} = \lVert z^\top \rVert_\mathrm{op} \leq \sqrt{d}$ and $ p =\operatorname{softmax}(\frac{z^\top q}{\sqrt{d}})$ is a probability vector which implies
\[
\lVert p \rVert^2 = \sum_{i=1}^N p_i^2 \leq \sum_{i=1}^N p_i = 1,
\]
where the first inequality comes from $x \mapsto x^2 \leq x \mapsto x$ on $[0,1]$. It follows that
\[
\left\lVert\mleft[z \operatorname{softmax}(\frac{z^\top q}{\sqrt{d}})\mright]^\top \right\rVert \leq \sqrt{d}.
\]
In summary, with the same arguments as before, we can take the weight terms outside of the expectation and we obtain
\begin{equation}
    \label{eq:V_term}
    \lVert\nabla_{V}\mathcal{L}(\theta) \rVert^2 \leq d(1 +2\lVert W \rVert_\mathrm{op}\lVert U \rVert_\mathrm{op})^2 \cdot \mathbb{E}_x\mleft[\lVert\mathscr{C}_x\rVert^2\mright].
\end{equation}

\paragraph{Receptors terms.} With similar arguments, we have
\begin{align*}
\lVert \nabla_{W}\mathcal{L}(\theta) \rVert^2 &= \lVert \mathbb{E}_x\mleft[ \Sigma_{\bar{\xi}} U^\top \mathscr{C}_x \bar{\xi}^\top \mright] \rVert^2 \\
&\leq \mathbb{E}_x\mleft[ \lVert \underbrace{\Sigma_{\bar{\xi}}U^\top \mathscr{C}_x}_{\in \R^d}  \underbrace{\bar{\xi}^\top}_{\in \R^{1 \times d}} \rVert^2 \mright] \tag{Jensen} \\
&\leq \mathbb{E}_x\mleft[ \lVert \Sigma_{\bar{\xi}}U^\top \mathscr{C}_x\rVert^2  \lVert \bar{\xi}^\top \rVert^2 \mright] \tag{Point (1) of \cref{lem:properties}} 
\end{align*}
By definition, $\bar{\xi} = \xi / \lVert \xi \rVert$ has unit norm. Moreover, using the point (2) of \cref{lem:properties}, we know that
\[
\lVert \Sigma_{\bar{\xi}}U^\top \mathscr{C}_x\rVert \leq \lVert \Sigma_{\bar{\xi}}U^\top \rVert_\mathrm{op} \cdot \lVert \mathscr{C}_x\rVert \leq \lVert \Sigma_{\bar{\xi}}\rVert_\mathrm{op} \cdot \lVert U^\top \rVert_\mathrm{op} \cdot \lVert \mathscr{C}_x\rVert.
\]
where the last inequality comes from the sub-multiplicity of the operator norm. As we showed in the deviation of the query term, we have $\lVert \Sigma_{\bar{\xi}}\rVert_\mathrm{op} \leq 2$ and we know that $\lVert U^\top \rVert_\mathrm{op} = \lVert U\rVert_\mathrm{op}$. In summary, taking the weight terms outside of the expectation, we have
\begin{equation}
    \label{eq:W_terms}
    \lVert \nabla_{W}\mathcal{L}(\theta) \rVert^2 \leq 4 \lVert U \rVert_\mathrm{op}^2 \mathbb{E}_x\mleft[\lVert \mathscr{C}_x\rVert^2\mright].
\end{equation}
\paragraph{Assemblers terms.} The same arguments as before lead to 
\[
\lVert
\nabla_{U}\mathcal{L}(\theta) \rVert^2 =\lVert \mathbb{E}_x\mleft[ \mathscr{C}_x\sigma(W\bar{\xi})^\top\mright]\rVert^2  \leq \mathbb{E}_x\mleft[ \lVert\mathscr{C}_x\sigma(W\bar{\xi})^\top \rVert^2\mright] \leq \mathbb{E}_x\mleft[ \lVert\mathscr{C}_x\rVert^2 \cdot \lVert\sigma(W\bar{\xi})^\top \rVert^2\mright]
\]
where the last inequality comes from the first point of \cref{lem:properties}. Recalling that the GeLU function is defined as $x \leq x \phi(x)$ with $\phi$ the cumulative distribution function of the standard normal distribution (in particular, $\phi$ outputs values in [0, 1]), we know that for any real number $x$, $\lvert \sigma(x) \rvert \leq \lvert x \rvert$. Using again the point (2) of \cref{lem:properties} and that $\bar{\xi}$ has unit norm, since $\sigma$ is applied entry-wise, we have
\[
\lVert\sigma(W\bar{\xi})^\top \rVert = \lVert\sigma(W\bar{\xi})\rVert \leq \lVert W\bar{\xi} \rVert \leq \lVert W\rVert_\mathrm{op}.
\]
In summary, taking the weight terms outside of the expectation, we have 
\begin{equation}
    \label{eq:U_terms}
    \lVert
\nabla_{U}\mathcal{L}(\theta) \rVert^2 \leq \lVert W\rVert_\mathrm{op}^2 \mathbb{E}_x\mleft[\lVert \mathscr{C}_x\rVert^2\mright].
\end{equation}
\paragraph{Conclusion.} Putting \eqref{eq:q_term}, \eqref{eq:V_term}, \eqref{eq:W_terms} and \eqref{eq:U_terms} together and using \eqref{eq:norm_decomp} leads to
\begin{equation}
\label{eq:norm_loss_ub}
\begin{split}
&\lVert \nabla_\theta \mathcal{L}(\theta) \rVert^2 \\
& \leq \mleft[ \mleft[2n \sqrt{d} \lVert V \rVert_\mathrm{op} (1 +\sqrt{\frac{2}{\pi}}\lVert W \rVert_\mathrm{op}^2\lVert U \rVert_\mathrm{op})\mright]^2 + d(1 +2\lVert W \rVert_\mathrm{op}\lVert U \rVert_\mathrm{op})^2 + 4\lVert U \rVert_\mathrm{op}^2 + \lVert W \rVert_\mathrm{op}^2\mright] \cdot \mathbb{E}_x\mleft[\lVert \mathscr{C}_x\rVert^2\mright].
\end{split}
\end{equation}
Finally, denoting $\gamma_j = \hat{p}_{\theta}(y=j|x) - p(y=j|x)$ to ease notations, we notice that 
\begin{align*}
\lVert \mathscr{C}_x\rVert^2 &= \left \lVert \sum_{j=1}^p (\hat{p}_{\theta}(y=j|x) - p(y=j|x)) E(j) \right \rVert^2 \\
&= \left \lVert \sum_{j=1}^p \gamma_j E(j) \right \rVert^2 \\
&= \sum_{l=1}^d \mleft( \sum_{j=1}^p \gamma_j E(j)_l\mright)^2 \\
&\leq \sum_{l=1}^d \mleft( \sum_{j=1}^p \lvert \gamma_j \rvert \lvert E(j)_l \rvert\mright)^2 \\
&\leq \mleft(\sum_{l=1}^d \max_{j\in [p]} \lvert  E(j)_l \rvert^2 \mright)\mleft( \sum_{j=1}^p \lvert \gamma_j \rvert\mright)^2 \\
&\leq dB^2 \mleft( \sum_{j=1}^p \lvert \gamma_j \rvert\mright)^2 \tag{using $\lVert E \rVert_{\infty} \leq B$}.
\end{align*}
Recalling that the total variation distance verifies for any probability vectors $p, q$
\[
\TV{p}{q} = \frac{1}{2} \lVert p-q \rVert_1 = \frac{1}{2} \sum_{\omega \in \Omega} \lvert p(\omega) - q(\omega) \rvert,
\]
with $\lVert \cdot \rVert_1$ the $\ell_1$-norm. We notice that
\[
\sum_{j=1}^p \lvert \gamma_j \rvert = \sum_{j=1}^p \lvert \hat{p}_{\theta}(y=j|x) - p(y=j|x) \rvert = \lVert \hat{p}_{\theta}(\cdot|x) - p(\cdot|x) \rVert_1 = 2 \cdot \TV{\hat{p}_{\theta}(\cdot|x)}{p(\cdot|x)}.
\]
Recalling that the total variation distance takes values in $[0, 1]$ and that $x \mapsto x^2 \leq x \mapsto x$ on this interval, we know that $\TV{\hat{p}_{\theta}(\cdot|x)}{p(\cdot|x)} ^ 2 \leq \TV{\hat{p}_{\theta}(\cdot|x)}{p(\cdot|x)}$. It leads to
\[
\mathbb{E}_x\mleft[\lVert \mathscr{C}_x\rVert^2\mright] \leq 4dB^2 \mathbb{E}_x\mleft[\TV{\hat{p}_{\theta}(\cdot|x)}{p(\cdot|x)}\mright].
\]
Rearranging the terms in front of the expectation in \eqref{eq:norm_loss_ub} finally leads to
\[
\lVert \nabla_\theta \mathcal{L}(\theta) \rVert^2 \leq \hat{B} \cdot \mathbb{E}_x\mleft[\TV{\hat{p}_{\theta}(\cdot|x)}{p(\cdot|x)}\mright]
\]
where $\hat{B} = 4dB^2 \mleft[ \mleft[2n \sqrt{d} \lVert V \rVert_\mathrm{op} (1 +\sqrt{\frac{2}{\pi}}\lVert W \rVert_\mathrm{op}^2\lVert U \rVert_\mathrm{op})\mright]^2 + d(1 +2\lVert W \rVert_\mathrm{op}\lVert U \rVert_\mathrm{op})^2 + 4\lVert U \rVert_\mathrm{op}^2 + \lVert W \rVert_\mathrm{op}^2\mright]$. Noticing that $\hat{B} = \mathcal{O}\mleft(B^2 \mleft[\lVert V \rVert_\mathrm{op} \mleft[ 1 + \lVert U \rVert_\mathrm{op}\lVert W \rVert_\mathrm{op}^2\mright]^2 + \mleft[1 +  \lVert U \rVert_\mathrm{op}\lVert W \rVert_\mathrm{op} \mright]^2 +  \lVert U \rVert_\mathrm{op}^2 + \lVert W \rVert_\mathrm{op}^2\mright]\mright)$ and taking the square root leads to
\begin{align*}
&\lVert \nabla_\theta \mathcal{L}(\theta) \rVert \\
&= \mathcal{O}\mleft(B\sqrt{\lVert V \rVert_\mathrm{op} \mleft[ 1 + \lVert U \rVert_\mathrm{op}\lVert W \rVert_\mathrm{op}^2\mright]^2 + \mleft[1 +  \lVert U \rVert_\mathrm{op}\lVert W \rVert_\mathrm{op} \mright]^2 +  \lVert U \rVert_\mathrm{op}^2 + \lVert W \rVert_\mathrm{op}^2} \cdot \sqrt{\mathbb{E}_x\mleft[\TV{\hat{p}_{\theta}(\cdot|x)}{p(\cdot|x)}\mright]} \mright),
\end{align*}
which concludes the proof. 
\end{proof}

\end{document}